\documentclass[preprint,12pt]{elsarticle}

\usepackage{amssymb}
\usepackage{amsmath}
\usepackage{graphicx}
\usepackage{xcolor}
\definecolor{revisionred}{RGB}{0,0,0}
\newcommand{\rev}[1]{\textcolor{revisionred}{#1}}
\graphicspath{{sim_files/}{sim_files/revision_outputs/}}

\usepackage{lineno}

\begin{document}

\begin{frontmatter}



\title{Drivetrain simulation using variational autoencoders}

\author[label1,label2,label3]{Pallavi Sharma}
\author[label4]{Jorge-Humberto Urrea-Quintero}
\author[label1]{Bogdan Bogdan}
\author[label1]{Adrian-Dumitru Ciotec}
\author[label1]{Laura Vasilie}
\author[label4]{Henning Wessels}
\author[label1]{Matteo Skull}

\affiliation[label1]{%
  organization={Porsche Engineering},
  addressline={Etzelstraße 1},
  city={Bietigheim-Bissingen},
  postcode={74321},
  country={Germany}
}

\affiliation[label2]{%
  organization={Technische Universität Braunschweig, Graduate Program Computational Sciences in Engineering},
  city={Braunschweig},
  country={Germany}
}

\affiliation[label3]{%
  organization={Fraunhofer Institute for Energy Economics and Energy System Technology},
  addressline={Joseph-Beuys-Straße 8},
  city={Kassel},
  postcode={34117},
  country={Germany}
}

\affiliation[label4]{%
  organization={Technische Universität Braunschweig, Institute of Applied Mechanics, Division Data-Driven Modeling of Mechanical Systems},
  addressline={Pockelsstraße 3},
  city={Braunschweig},
  postcode={38106},
  country={Germany}
}



\begin{abstract}
\rev{Simulation-based drivetrain jerk assessment is central for ride comfort analysis in automotive engineering. Traditional simulation approaches are rooted in physical principles and require exhaustive system parameterization. For the design of new platforms, in particular battery electric vehicles, experimental data for system parameterization may be extremely limited or very costly to obtain. This is especially true for early design stages. The main contribution of this paper is to demonstrate that Variational Autoencoder (VAE) architectures can serve as effective generative models for drivetrain jerk simulation. The results support that VAEs can yield physically plausible results even when measured data are scarce, conditioning is required, and full physical re-parameterization is not available. The approach presented reduces the dependence on costly real-world experiments and extensive manual modeling.
}
\end{abstract}




\end{frontmatter}



\newcommand{\HW}{\textcolor{blue}}
\newcommand{\JPF}{\textcolor{green}}
\newcommand{\CW}{\textcolor{orange}}

\newcommand{\lrap}[1]{``#1''}

\newcommand{\Caption}[1]{ \begin{singlespace} \em{\caption{#1}} \end{singlespace} }

\long\def\symbolfootnote[#1]#2{\begingroup \def\thefootnote{\fnsymbol{footnote}}\footnote[#1]{#2} \endgroup} 

\renewcommand{\vec}[1]{ \ensuremath{ \mathbf{ #1 } } }
\newcommand{\ten}[1]{ \ensuremath {\mathbf{#1} } }

\newcommand{\gvec}[1]{ \ensuremath{ \boldsymbol{ #1 } } }
\newcommand{\gten}[1]{ \ensuremath {\boldsymbol{#1} } }


\newcommand{\mn}{_{\mbox{\tiny{N}}}}
\newcommand{\mt}{_{\mbox{\tiny{T}}}}

\newcommand{\dint}{\mbox{d}}
\newcommand{\de}{\,\mbox{det}\,}
\newcommand{\gra}{\,\mbox{grad}\,}
\newcommand{\Gra}{\,\mbox{Grad}\,}
\newcommand{\tra}{\,\mbox{tr}\,}
\newcommand{\Div}{\mbox{Div}\,}
\renewcommand{\div}{\mbox{div}\,}
\newcommand{\sign}{\mbox{sign}}

\newcommand{\ncdot}{\hspace{-0.14cm}\cdot}

\newcommand{\hata}{\ensuremath{ \widehat{a} }}
\newcommand{\hatb}{\ensuremath{ \widehat{b} }}
\newcommand{\hatc}{\ensuremath{ \widehat{c} }}
\newcommand{\hatd}{\ensuremath{ \widehat{d} }}
\newcommand{\hate}{\ensuremath{ \widehat{e} }}
\newcommand{\hatf}{\ensuremath{ \widehat{f} }}
\newcommand{\hatg}{\ensuremath{ \widehat{g} }}
\newcommand{\hath}{\ensuremath{ \widehat{h} }}
\newcommand{\hati}{\ensuremath{ \widehat{i} }}
\newcommand{\hatj}{\ensuremath{ \widehat{j} }}
\newcommand{\hatk}{\ensuremath{ \widehat{k} }}
\newcommand{\hatl}{\ensuremath{ \widehat{l} }}
\newcommand{\hatm}{\ensuremath{ \widehat{m} }}
\newcommand{\hatn}{\ensuremath{ \widehat{n} }}
\newcommand{\hato}{\ensuremath{ \widehat{o} }}
\newcommand{\hatp}{\ensuremath{ \widehat{p} }}
\newcommand{\hatq}{\ensuremath{ \widehat{q} }}
\newcommand{\hatr}{\ensuremath{ \widehat{r} }}
\newcommand{\hats}{\ensuremath{ \widehat{s} }}
\newcommand{\hatt}{\ensuremath{ \widehat{t} }}
\newcommand{\hatu}{\ensuremath{ \widehat{u} }}
\newcommand{\hatv}{\ensuremath{ \widehat{v} }}
\newcommand{\hatw}{\ensuremath{ \widehat{w} }}
\newcommand{\hatx}{\ensuremath{ \widehat{x} }}
\newcommand{\haty}{\ensuremath{ \widehat{y} }}
\newcommand{\hatz}{\ensuremath{ \widehat{z} }}

\newcommand{\hatA}{\ensuremath{ \widehat{A} }}
\newcommand{\hatB}{\ensuremath{ \widehat{B} }}
\newcommand{\hatC}{\ensuremath{ \widehat{C} }}
\newcommand{\hatD}{\ensuremath{ \widehat{D} }}
\newcommand{\hatE}{\ensuremath{ \widehat{E} }}
\newcommand{\hatF}{\ensuremath{ \widehat{F} }}
\newcommand{\hatG}{\ensuremath{ \widehat{G} }}
\newcommand{\hatH}{\ensuremath{ \widehat{H} }}
\newcommand{\hatI}{\ensuremath{ \widehat{I} }}
\newcommand{\hatJ}{\ensuremath{ \widehat{J} }}
\newcommand{\hatK}{\ensuremath{ \widehat{K} }}
\newcommand{\hatL}{\ensuremath{ \widehat{L} }}
\newcommand{\hatM}{\ensuremath{ \widehat{M} }}
\newcommand{\hatN}{\ensuremath{ \widehat{N} }}
\newcommand{\hatO}{\ensuremath{ \widehat{O} }}
\newcommand{\hatP}{\ensuremath{ \widehat{P} }}
\newcommand{\hatQ}{\ensuremath{ \widehat{Q} }}
\newcommand{\hatR}{\ensuremath{ \widehat{R} }}
\newcommand{\hatS}{\ensuremath{ \widehat{S} }}
\newcommand{\hatT}{\ensuremath{ \widehat{T} }}
\newcommand{\hatU}{\ensuremath{ \widehat{U} }}
\newcommand{\hatV}{\ensuremath{ \widehat{V} }}
\newcommand{\hatW}{\ensuremath{ \widehat{W} }}
\newcommand{\hatX}{\ensuremath{ \widehat{X} }}
\newcommand{\hatY}{\ensuremath{ \widehat{Y} }}
\newcommand{\hatZ}{\ensuremath{ \widehat{Z }}}

\newcommand{\tila}{\ensuremath{ \widetilde{a} }}
\newcommand{\tilb}{\ensuremath{ \widetilde{b} }}
\newcommand{\tilc}{\ensuremath{ \widetilde{c} }}
\newcommand{\tild}{\ensuremath{ \widetilde{d} }}
\newcommand{\tile}{\ensuremath{ \widetilde{e} }}
\newcommand{\tilf}{\ensuremath{ \widetilde{f} }}
\newcommand{\tilg}{\ensuremath{ \widetilde{g} }}
\newcommand{\tilh}{\ensuremath{ \widetilde{h} }}
\newcommand{\tili}{\ensuremath{ \widetilde{i} }}
\newcommand{\tilj}{\ensuremath{ \widetilde{j} }}
\newcommand{\tilk}{\ensuremath{ \widetilde{k} }}
\newcommand{\till}{\ensuremath{ \widetilde{l} }}
\newcommand{\tilm}{\ensuremath{ \widetilde{m} }}
\newcommand{\tiln}{\ensuremath{ \widetilde{n} }}
\newcommand{\tilo}{\ensuremath{ \widetilde{o} }}
\newcommand{\tilp}{\ensuremath{ \widetilde{p} }}
\newcommand{\tilq}{\ensuremath{ \widetilde{q} }}
\newcommand{\tilr}{\ensuremath{ \widetilde{r} }}
\newcommand{\tils}{\ensuremath{ \widetilde{s} }}
\newcommand{\tilt}{\ensuremath{ \widetilde{t} }}
\newcommand{\tilu}{\ensuremath{ \widetilde{u} }}
\newcommand{\tilv}{\ensuremath{ \widetilde{v} }}
\newcommand{\tilw}{\ensuremath{ \widetilde{w} }}
\newcommand{\tilx}{\ensuremath{ \widetilde{x} }}
\newcommand{\tily}{\ensuremath{ \widetilde{y} }}
\newcommand{\tilz}{\ensuremath{ \widetilde{z} }}

\newcommand{\tilA}{\ensuremath{ \widetilde{A} }}
\newcommand{\tilB}{\ensuremath{ \widetilde{B} }}
\newcommand{\tilC}{\ensuremath{ \widetilde{C} }}
\newcommand{\tilD}{\ensuremath{ \widetilde{D} }}
\newcommand{\tilE}{\ensuremath{ \widetilde{E} }}
\newcommand{\tilF}{\ensuremath{ \widetilde{F} }}
\newcommand{\tilG}{\ensuremath{ \widetilde{G} }}
\newcommand{\tilH}{\ensuremath{ \widetilde{H} }}
\newcommand{\tilI}{\ensuremath{ \widetilde{I} }}
\newcommand{\tilJ}{\ensuremath{ \widetilde{J} }}
\newcommand{\tilK}{\ensuremath{ \widetilde{K} }}
\newcommand{\tilL}{\ensuremath{ \widetilde{L} }}
\newcommand{\tilM}{\ensuremath{ \widetilde{M} }}
\newcommand{\tilN}{\ensuremath{ \widetilde{N} }}
\newcommand{\tilO}{\ensuremath{ \widetilde{O} }}
\newcommand{\tilP}{\ensuremath{ \widetilde{P} }}
\newcommand{\tilQ}{\ensuremath{ \widetilde{Q} }}
\newcommand{\tilR}{\ensuremath{ \widetilde{R} }}
\newcommand{\tilS}{\ensuremath{ \widetilde{S} }}
\newcommand{\tilT}{\ensuremath{ \widetilde{T} }}
\newcommand{\tilU}{\ensuremath{ \widetilde{U} }}
\newcommand{\tilV}{\ensuremath{ \widetilde{V} }}
\newcommand{\tilW}{\ensuremath{ \widetilde{W} }}
\newcommand{\tilX}{\ensuremath{ \widetilde{X} }}
\newcommand{\tilY}{\ensuremath{ \widetilde{Y} }}
\newcommand{\tilZ}{\ensuremath{ \widetilde{Z }}}

\newcommand{\calA}{\ensuremath{ \mathcal{A} }}
\newcommand{\calB}{\ensuremath{ \mathcal{B} }}
\newcommand{\calC}{\ensuremath{ \mathcal{C} }}
\newcommand{\calD}{\ensuremath{ \mathcal{D} }}
\newcommand{\calE}{\ensuremath{ \mathcal{E} }}
\newcommand{\calF}{\ensuremath{ \mathcal{F} }}
\newcommand{\calG}{\ensuremath{ \mathcal{G} }}
\newcommand{\calH}{\ensuremath{ \mathcal{H} }}
\newcommand{\calI}{\ensuremath{ \mathcal{I} }}
\newcommand{\calJ}{\ensuremath{ \mathcal{J} }}
\newcommand{\calK}{\ensuremath{ \mathcal{K} }}
\newcommand{\calL}{\ensuremath{ \mathcal{L} }}
\newcommand{\calM}{\ensuremath{ \mathcal{M} }}
\newcommand{\calN}{\ensuremath{ \mathcal{N} }}
\newcommand{\calO}{\ensuremath{ \mathcal{O} }}
\newcommand{\calP}{\ensuremath{ \mathcal{P} }}
\newcommand{\calQ}{\ensuremath{ \mathcal{Q} }}
\newcommand{\calR}{\ensuremath{ \mathcal{R} }}
\newcommand{\calS}{\ensuremath{ \mathcal{S} }}
\newcommand{\calT}{\ensuremath{ \mathcal{T} }}
\newcommand{\calU}{\ensuremath{ \mathcal{U} }}
\newcommand{\calV}{\ensuremath{ \mathcal{V} }}
\newcommand{\calW}{\ensuremath{ \mathcal{W} }}
\newcommand{\calX}{\ensuremath{ \mathcal{X} }}
\newcommand{\calY}{\ensuremath{ \mathcal{Y} }}
\newcommand{\calZ}{\ensuremath{ \mathcal{Z} }}

\newcommand{\bbA}{\ensuremath{ \mathbb{A} }}
\newcommand{\bbB}{\ensuremath{ \mathbb{B} }}
\newcommand{\bbC}{\ensuremath{ \mathbb{C} }}
\newcommand{\bbD}{\ensuremath{ \mathbb{D} }}
\newcommand{\bbE}{\ensuremath{ \mathbb{E} }}
\newcommand{\bbF}{\ensuremath{ \mathbb{F} }}
\newcommand{\bbG}{\ensuremath{ \mathbb{G} }}
\newcommand{\bbH}{\ensuremath{ \mathbb{H} }}
\newcommand{\bbI}{\ensuremath{ \mathbb{I} }}
\newcommand{\bbJ}{\ensuremath{ \mathbb{J} }}
\newcommand{\bbK}{\ensuremath{ \mathbb{K} }}
\newcommand{\bbL}{\ensuremath{ \mathbb{L} }}
\newcommand{\bbM}{\ensuremath{ \mathbb{M} }}
\newcommand{\bbN}{\ensuremath{ \mathbb{N} }}
\newcommand{\bbO}{\ensuremath{ \mathbb{O} }}
\newcommand{\bbP}{\ensuremath{ \mathbb{P} }}
\newcommand{\bbQ}{\ensuremath{ \mathbb{Q} }}
\newcommand{\bbR}{\ensuremath{ \mathbb{R} }}
\newcommand{\bbS}{\ensuremath{ \mathbb{S} }}
\newcommand{\bbT}{\ensuremath{ \mathbb{T} }}
\newcommand{\bbU}{\ensuremath{ \mathbb{U} }}
\newcommand{\bbV}{\ensuremath{ \mathbb{V} }}
\newcommand{\bbW}{\ensuremath{ \mathbb{W} }}
\newcommand{\bbX}{\ensuremath{ \mathbb{X} }}
\newcommand{\bbY}{\ensuremath{ \mathbb{Y} }}
\newcommand{\bbZ}{\ensuremath{ \mathbb{Z} }}

\newcommand{\tenA}{\ensuremath{ \ten{A} }}
\newcommand{\tenB}{\ensuremath{ \ten{B} }}
\newcommand{\tenC}{\ensuremath{ \ten{C} }}
\newcommand{\tenD}{\ensuremath{ \ten{D} }}
\newcommand{\tenE}{\ensuremath{ \ten{E} }}
\newcommand{\tenF}{\ensuremath{ \ten{F} }}
\newcommand{\tenG}{\ensuremath{ \ten{G} }}
\newcommand{\tenH}{\ensuremath{ \ten{H} }}
\newcommand{\tenI}{\ensuremath{ \ten{I} }}
\newcommand{\tenJ}{\ensuremath{ \ten{J} }}
\newcommand{\tenK}{\ensuremath{ \ten{K} }}
\newcommand{\tenL}{\ensuremath{ \ten{L} }}
\newcommand{\tenM}{\ensuremath{ \ten{M} }}
\newcommand{\tenN}{\ensuremath{ \ten{N} }}
\newcommand{\tenO}{\ensuremath{ \ten{O} }}
\newcommand{\tenP}{\ensuremath{ \ten{P} }}
\newcommand{\tenQ}{\ensuremath{ \ten{Q} }}
\newcommand{\tenR}{\ensuremath{ \ten{R} }}
\newcommand{\tenS}{\ensuremath{ \ten{S} }}
\newcommand{\tenT}{\ensuremath{ \ten{T} }}
\newcommand{\tenU}{\ensuremath{ \ten{U} }}
\newcommand{\tenV}{\ensuremath{ \ten{V} }}
\newcommand{\tenW}{\ensuremath{ \ten{W} }}
\newcommand{\tenX}{\ensuremath{ \ten{X} }}
\newcommand{\tenY}{\ensuremath{ \ten{Y} }}
\newcommand{\tenZ}{\ensuremath{ \ten{Z} }}

\newcommand{\tena}{\ensuremath{ \ten{a} }}
\newcommand{\tenb}{\ensuremath{ \ten{b} }}
\newcommand{\tenc}{\ensuremath{ \ten{c} }}
\newcommand{\tend}{\ensuremath{ \ten{d} }}
\newcommand{\tene}{\ensuremath{ \ten{e} }}
\newcommand{\tenf}{\ensuremath{ \ten{f} }}
\newcommand{\teng}{\ensuremath{ \ten{g} }}
\newcommand{\tenh}{\ensuremath{ \ten{h} }}
\newcommand{\teni}{\ensuremath{ \ten{i} }}
\newcommand{\tenj}{\ensuremath{ \ten{j} }}
\newcommand{\tenk}{\ensuremath{ \ten{k} }}
\newcommand{\tenl}{\ensuremath{ \ten{l} }}
\newcommand{\tenm}{\ensuremath{ \ten{m} }}
\newcommand{\tenn}{\ensuremath{ \ten{n} }}
\newcommand{\teno}{\ensuremath{ \ten{o} }}
\newcommand{\tenp}{\ensuremath{ \ten{p} }}
\newcommand{\tenq}{\ensuremath{ \ten{q} }}
\newcommand{\tenr}{\ensuremath{ \ten{r} }}
\newcommand{\tens}{\ensuremath{ \ten{s} }}
\newcommand{\tent}{\ensuremath{ \ten{t} }}
\newcommand{\tenu}{\ensuremath{ \ten{u} }}
\newcommand{\tenv}{\ensuremath{ \ten{v} }}
\newcommand{\tenw}{\ensuremath{ \ten{w} }}
\newcommand{\tenx}{\ensuremath{ \ten{x} }}
\newcommand{\teny}{\ensuremath{ \ten{y} }}
\newcommand{\tenz}{\ensuremath{ \ten{z} }}

\newcommand{\ttena}{\dot{\tena}}
\newcommand{\ttenb}{\dot{\tenb}}
\newcommand{\ttenc}{\dot{\tenc}}
\newcommand{\ttend}{\dot{\tend}}
\newcommand{\ttene}{\dot{\tene}}
\newcommand{\ttenf}{\dot{\tenf}}

\newcommand{\ttenE}{\dot{\tenE}}
\newcommand{\ttenF}{\dot{\tenF}}
\newcommand{\ttenS}{\dot{\tenS}}
\newcommand{\ttenT}{\dot{\tenT}}
\newcommand{\ttenU}{\dot{\tenU}}
\newcommand{\ttenV}{\dot{\tenV}}
\newcommand{\ttenW}{\dot{\tenW}}
\newcommand{\ttenX}{\dot{\tenX}}
\newcommand{\ttenY}{\dot{\tenY}}
\newcommand{\ttenZ}{\dot{\tenZ}}

\newcommand{\ltenA}{\stackrel{\triangle}{\tenA}}
\newcommand{\ltenD}{\stackrel{\triangle}{\tenD}}
\newcommand{\ltenM}{\stackrel{\triangle}{\tenM}}

\newcommand{\btena}{\bar{\tena}}
\newcommand{\btenb}{\bar{\tenb}}
\newcommand{\btenc}{\bar{\tenc}}
\newcommand{\btend}{\bar{\tend}}
\newcommand{\btene}{\bar{\tene}}
\newcommand{\btenf}{\bar{\tenf}}
\newcommand{\bteng}{\bar{\teng}}

\newcommand{\btenA}{\bar{\tenA}}
\newcommand{\btenB}{\bar{\tenB}}
\newcommand{\btenC}{\bar{\tenC}}
\newcommand{\btenD}{\bar{\tenD}}
\newcommand{\btenE}{\bar{\tenE}}
\newcommand{\btenF}{\bar{\tenF}}
\newcommand{\btenK}{\bar{\tenK}}
\newcommand{\btenL}{\bar{\tenL}}
\newcommand{\btenM}{\bar{\tenM}}
\newcommand{\btenN}{\bar{\tenN}}
\newcommand{\btenO}{\bar{\tenO}}
\newcommand{\btenP}{\bar{\tenP}}
\newcommand{\btenQ}{\bar{\tenQ}}
\newcommand{\btenR}{\bar{\tenR}}
\newcommand{\btenS}{\bar{\tenS}}
\newcommand{\btenT}{\bar{\tenT}}


\newcommand{\lbtenE}{\Delta\bar{\tenE}}

\newcommand{\hattena}{\ensuremath{ \widehat{\ten{a}} }}
\newcommand{\hattenb}{\ensuremath{ \widehat{\ten{b}} }}
\newcommand{\hattenc}{\ensuremath{ \widehat{\ten{c}} }}
\newcommand{\hattend}{\ensuremath{ \widehat{\ten{d}} }}
\newcommand{\hattene}{\ensuremath{ \widehat{\ten{e}} }}
\newcommand{\hattenf}{\ensuremath{ \widehat{\ten{f}} }}
\newcommand{\hatteng}{\ensuremath{ \widehat{\ten{g}} }}
\newcommand{\hattenh}{\ensuremath{ \widehat{\ten{h}} }}
\newcommand{\hatteni}{\ensuremath{ \widehat{\ten{i}} }}
\newcommand{\hattenj}{\ensuremath{ \widehat{\ten{j}} }}
\newcommand{\hattenk}{\ensuremath{ \widehat{\ten{k}} }}
\newcommand{\hattenl}{\ensuremath{ \widehat{\ten{l}} }}
\newcommand{\hattenm}{\ensuremath{ \widehat{\ten{m}} }}
\newcommand{\hattenn}{\ensuremath{ \widehat{\ten{n}} }}
\newcommand{\hatteno}{\ensuremath{ \widehat{\ten{o}} }}
\newcommand{\hattenp}{\ensuremath{ \widehat{\ten{p}} }}
\newcommand{\hattenq}{\ensuremath{ \widehat{\ten{q}} }}
\newcommand{\hattenr}{\ensuremath{ \widehat{\ten{r}} }}
\newcommand{\hattens}{\ensuremath{ \widehat{\ten{s}} }}
\newcommand{\hattent}{\ensuremath{ \widehat{\ten{t}} }}
\newcommand{\hattenu}{\ensuremath{ \widehat{\ten{u}} }}
\newcommand{\hattenv}{\ensuremath{ \widehat{\ten{v}} }}
\newcommand{\hattenw}{\ensuremath{ \widehat{\ten{w}} }}
\newcommand{\hattenx}{\ensuremath{ \widehat{\ten{x}} }}
\newcommand{\hatteny}{\ensuremath{ \widehat{\ten{y}} }}
\newcommand{\hattenz}{\ensuremath{ \widehat{\ten{z}} }}

\newcommand{\hattenA}{\ensuremath{ \widehat{\ten{A}} }}
\newcommand{\hattenB}{\ensuremath{ \widehat{\ten{B}} }}
\newcommand{\hattenC}{\ensuremath{ \widehat{\ten{C}} }}
\newcommand{\hattenD}{\ensuremath{ \widehat{\ten{D}} }}
\newcommand{\hattenE}{\ensuremath{ \widehat{\ten{E}} }}
\newcommand{\hattenF}{\ensuremath{ \widehat{\ten{F}} }}
\newcommand{\hattenG}{\ensuremath{ \widehat{\ten{G}} }}
\newcommand{\hattenH}{\ensuremath{ \widehat{\ten{H}} }}
\newcommand{\hattenI}{\ensuremath{ \widehat{\ten{I}} }}
\newcommand{\hattenJ}{\ensuremath{ \widehat{\ten{J}} }}
\newcommand{\hattenK}{\ensuremath{ \widehat{\ten{K}} }}
\newcommand{\hattenL}{\ensuremath{ \widehat{\ten{L}} }}
\newcommand{\hattenM}{\ensuremath{ \widehat{\ten{M}} }}
\newcommand{\hattenN}{\ensuremath{ \widehat{\ten{N}} }}
\newcommand{\hattenO}{\ensuremath{ \widehat{\ten{O}} }}
\newcommand{\hattenP}{\ensuremath{ \widehat{\ten{P}} }}
\newcommand{\hattenQ}{\ensuremath{ \widehat{\ten{Q}} }}
\newcommand{\hattenR}{\ensuremath{ \widehat{\ten{R}} }}
\newcommand{\hattenS}{\ensuremath{ \widehat{\ten{S}} }}
\newcommand{\hattenT}{\ensuremath{ \widehat{\ten{T}} }}
\newcommand{\hattenU}{\ensuremath{ \widehat{\ten{U}} }}
\newcommand{\hattenV}{\ensuremath{ \widehat{\ten{V}} }}
\newcommand{\hattenW}{\ensuremath{ \widehat{\ten{W}} }}
\newcommand{\hattenX}{\ensuremath{ \widehat{\ten{X}} }}
\newcommand{\hattenY}{\ensuremath{ \widehat{\ten{Y}} }}
\newcommand{\hattenZ}{\ensuremath{ \widehat{\ten{Z}} }}

\newcommand{\tiltena}{\ensuremath{ \widetilde{\ten{a}} }}
\newcommand{\tiltenb}{\ensuremath{ \widetilde{\ten{b}} }}
\newcommand{\tiltenc}{\ensuremath{ \widetilde{\ten{c}} }}
\newcommand{\tiltend}{\ensuremath{ \widetilde{\ten{d}} }}
\newcommand{\tiltene}{\ensuremath{ \widetilde{\ten{e}} }}
\newcommand{\tiltenf}{\ensuremath{ \widetilde{\ten{f}} }}
\newcommand{\tilteng}{\ensuremath{ \widetilde{\ten{g}} }}
\newcommand{\tiltenh}{\ensuremath{ \widetilde{\ten{h}} }}
\newcommand{\tilteni}{\ensuremath{ \widetilde{\ten{i}} }}
\newcommand{\tiltenj}{\ensuremath{ \widetilde{\ten{j}} }}
\newcommand{\tiltenk}{\ensuremath{ \widetilde{\ten{k}} }}
\newcommand{\tiltenl}{\ensuremath{ \widetilde{\ten{l}} }}
\newcommand{\tiltenm}{\ensuremath{ \widetilde{\ten{m}} }}
\newcommand{\tiltenn}{\ensuremath{ \widetilde{\ten{n}} }}
\newcommand{\tilteno}{\ensuremath{ \widetilde{\ten{o}} }}
\newcommand{\tiltenp}{\ensuremath{ \widetilde{\ten{p}} }}
\newcommand{\tiltenq}{\ensuremath{ \widetilde{\ten{q}} }}
\newcommand{\tiltenr}{\ensuremath{ \widetilde{\ten{r}} }}
\newcommand{\tiltens}{\ensuremath{ \widetilde{\ten{s}} }}
\newcommand{\tiltent}{\ensuremath{ \widetilde{\ten{t}} }}
\newcommand{\tiltenu}{\ensuremath{ \widetilde{\ten{u}} }}
\newcommand{\tiltenv}{\ensuremath{ \widetilde{\ten{v}} }}
\newcommand{\tiltenw}{\ensuremath{ \widetilde{\ten{w}} }}
\newcommand{\tiltenx}{\ensuremath{ \widetilde{\ten{x}} }}
\newcommand{\tilteny}{\ensuremath{ \widetilde{\ten{y}} }}
\newcommand{\tiltenz}{\ensuremath{ \widetilde{\ten{z}} }}

\newcommand{\tiltenA}{\ensuremath{ \widetilde{\ten{A}} }}
\newcommand{\tiltenB}{\ensuremath{ \widetilde{\ten{B}} }}
\newcommand{\tiltenC}{\ensuremath{ \widetilde{\ten{C}} }}
\newcommand{\tiltenD}{\ensuremath{ \widetilde{\ten{D}} }}
\newcommand{\tiltenE}{\ensuremath{ \widetilde{\ten{E}} }}
\newcommand{\tiltenF}{\ensuremath{ \widetilde{\ten{F}} }}
\newcommand{\tiltenG}{\ensuremath{ \widetilde{\ten{G}} }}
\newcommand{\tiltenH}{\ensuremath{ \widetilde{\ten{H}} }}
\newcommand{\tiltenI}{\ensuremath{ \widetilde{\ten{I}} }}
\newcommand{\tiltenJ}{\ensuremath{ \widetilde{\ten{J}} }}
\newcommand{\tiltenK}{\ensuremath{ \widetilde{\ten{K}} }}
\newcommand{\tiltenL}{\ensuremath{ \widetilde{\ten{L}} }}
\newcommand{\tiltenM}{\ensuremath{ \widetilde{\ten{M}} }}
\newcommand{\tiltenN}{\ensuremath{ \widetilde{\ten{N}} }}
\newcommand{\tiltenO}{\ensuremath{ \widetilde{\ten{O}} }}
\newcommand{\tiltenP}{\ensuremath{ \widetilde{\ten{P}} }}
\newcommand{\tiltenQ}{\ensuremath{ \widetilde{\ten{Q}} }}
\newcommand{\tiltenR}{\ensuremath{ \widetilde{\ten{R}} }}
\newcommand{\tiltenS}{\ensuremath{ \widetilde{\ten{S}} }}
\newcommand{\tiltenT}{\ensuremath{ \widetilde{\ten{T}} }}
\newcommand{\tiltenU}{\ensuremath{ \widetilde{\ten{U}} }}
\newcommand{\tiltenV}{\ensuremath{ \widetilde{\ten{V}} }}
\newcommand{\tiltenW}{\ensuremath{ \widetilde{\ten{W}} }}
\newcommand{\tiltenX}{\ensuremath{ \widetilde{\ten{X}} }}
\newcommand{\tiltenY}{\ensuremath{ \widetilde{\ten{Y}} }}
\newcommand{\tiltenZ}{\ensuremath{ \widetilde{\ten{Z}} }}

\newcommand{\veca}{\ensuremath{ \vec{a} }}
\newcommand{\vecb}{\ensuremath{ \vec{b} }}
\newcommand{\vecc}{\ensuremath{ \vec{c} }}
\newcommand{\vecd}{\ensuremath{ \vec{d} }}
\newcommand{\vece}{\ensuremath{ \vec{e} }}
\newcommand{\vecf}{\ensuremath{ \vec{f} }}
\newcommand{\vecg}{\ensuremath{ \vec{g} }}
\newcommand{\vech}{\ensuremath{ \vec{h} }}
\newcommand{\veci}{\ensuremath{ \vec{i} }}
\newcommand{\vecj}{\ensuremath{ \vec{j} }}
\newcommand{\veck}{\ensuremath{ \vec{k} }}
\newcommand{\vecl}{\ensuremath{ \vec{l} }}
\newcommand{\vecm}{\ensuremath{ \vec{m} }}
\newcommand{\vecn}{\ensuremath{ \vec{n} }}
\newcommand{\veco}{\ensuremath{ \vec{o} }}
\newcommand{\vecp}{\ensuremath{ \vec{p} }}
\newcommand{\vecq}{\ensuremath{ \vec{q} }}
\newcommand{\vecr}{\ensuremath{ \vec{r} }}
\newcommand{\vecs}{\ensuremath{ \vec{s} }}
\newcommand{\vect}{\ensuremath{ \vec{t} }}
\newcommand{\vecu}{\ensuremath{ \vec{u} }}
\newcommand{\vecv}{\ensuremath{ \vec{v} }}
\newcommand{\vecw}{\ensuremath{ \vec{w} }}
\newcommand{\vecx}{\ensuremath{ \vec{x} }}
\newcommand{\vecy}{\ensuremath{ \vec{y} }}
\newcommand{\vecz}{\ensuremath{ \vec{z} }}

\newcommand{\vecA}{\ensuremath{ \vec{A} }}
\newcommand{\vecB}{\ensuremath{ \vec{B} }}
\newcommand{\vecC}{\ensuremath{ \vec{C} }}
\newcommand{\vecD}{\ensuremath{ \vec{D} }}
\newcommand{\vecE}{\ensuremath{ \vec{E} }}
\newcommand{\vecF}{\ensuremath{ \vec{F} }}
\newcommand{\vecG}{\ensuremath{ \vec{G} }}
\newcommand{\vecH}{\ensuremath{ \vec{H} }}
\newcommand{\vecI}{\ensuremath{ \vec{I} }}
\newcommand{\vecJ}{\ensuremath{ \vec{J} }}
\newcommand{\vecK}{\ensuremath{ \vec{K} }}
\newcommand{\vecL}{\ensuremath{ \vec{L} }}
\newcommand{\vecM}{\ensuremath{ \vec{M} }}
\newcommand{\vecN}{\ensuremath{ \vec{N} }}
\newcommand{\vecO}{\ensuremath{ \vec{O} }}
\newcommand{\vecP}{\ensuremath{ \vec{P} }}
\newcommand{\vecQ}{\ensuremath{ \vec{Q} }}
\newcommand{\vecR}{\ensuremath{ \vec{R} }}
\newcommand{\vecS}{\ensuremath{ \vec{S} }}
\newcommand{\vecT}{\ensuremath{ \vec{T} }}
\newcommand{\vecU}{\ensuremath{ \vec{U} }}
\newcommand{\vecV}{\ensuremath{ \vec{V} }}
\newcommand{\vecW}{\ensuremath{ \vec{W} }}
\newcommand{\vecX}{\ensuremath{ \vec{X} }}
\newcommand{\vecY}{\ensuremath{ \vec{Y} }}
\newcommand{\vecZ}{\ensuremath{ \vec{Z} }}

\newcommand{\tveca}{\ensuremath{ \dot{\vec{a}} }}
\newcommand{\tvecb}{\dot{\vecb}}
\newcommand{\tvecc}{\ensuremath{ \vec{c} }}
\newcommand{\tvecd}{\ensuremath{ \vec{d} }}
\newcommand{\tvece}{\ensuremath{ \vec{e} }}
\newcommand{\tvecf}{\ensuremath{ \vec{f} }}
\newcommand{\tvecg}{\ensuremath{ \dot{\vec{g}} }}
\newcommand{\tvech}{\ensuremath{ \vec{h} }}
\newcommand{\tveci}{\ensuremath{ \vec{i} }}
\newcommand{\tvecj}{\ensuremath{ \vec{j} }}
\newcommand{\tveck}{\ensuremath{ \vec{k} }}
\newcommand{\tvecl}{\ensuremath{ \vec{l} }}
\newcommand{\tvecm}{\ensuremath{ \vec{m} }}
\newcommand{\tvecn}{\dot{\ensuremath{ \vec{n} }}}
\newcommand{\tveco}{\dot{\ensuremath{ \vec{o} }}}
\newcommand{\tvecp}{\dot{\ensuremath{ \vec{p} }}}
\newcommand{\tvecq}{\dot{\ensuremath{ \vec{q} }}}
\newcommand{\tvecr}{\ensuremath{ \vec{r} }}
\newcommand{\tvecs}{\ensuremath{ \vec{s} }}
\newcommand{\tvect}{\ensuremath{ \vec{t} }}
\newcommand{\tvecu}{\dot{\ensuremath{ \vec{u} }}}
\newcommand{\tvecv}{\dot{\ensuremath{ \vec{v} }}}
\newcommand{\tvecw}{\dot{\ensuremath{ \vec{w} }}}
\newcommand{\tvecx}{\dot{\ensuremath{ \vec{x} }}}
\newcommand{\tvecy}{\dot{\ensuremath{ \vec{y} }}}
\newcommand{\tvecz}{\dot{\ensuremath{ \vec{z} }}}

\newcommand{\tvecA}{\ensuremath{ \vec{A} }}
\newcommand{\tvecB}{\ensuremath{ \vec{B} }}
\newcommand{\tvecC}{\ensuremath{ \vec{C} }}
\newcommand{\tvecD}{\ensuremath{ \vec{D} }}
\newcommand{\tvecE}{\ensuremath{ \vec{E} }}
\newcommand{\tvecF}{\ensuremath{ \vec{F} }}
\newcommand{\tvecG}{\ensuremath{ \vec{G} }}
\newcommand{\tvecH}{\ensuremath{ \vec{H} }}
\newcommand{\tvecI}{\ensuremath{ \vec{I} }}
\newcommand{\tvecJ}{\ensuremath{ \vec{J} }}
\newcommand{\tvecK}{\ensuremath{ \vec{K} }}
\newcommand{\tvecL}{\ensuremath{ \vec{L} }}
\newcommand{\tvecM}{\ensuremath{ \vec{M} }}
\newcommand{\tvecN}{\ensuremath{ \vec{N} }}
\newcommand{\tvecO}{\ensuremath{ \vec{O} }}
\newcommand{\tvecP}{\ensuremath{ \vec{P} }}
\newcommand{\tvecQ}{\ensuremath{ \vec{Q} }}
\newcommand{\tvecR}{\ensuremath{ \vec{R} }}
\newcommand{\tvecS}{\ensuremath{ \vec{S} }}
\newcommand{\tvecT}{\ensuremath{ \vec{T} }}
\newcommand{\tvecU}{\ensuremath{ \vec{U} }}
\newcommand{\tvecV}{\ensuremath{ \vec{V} }}
\newcommand{\tvecW}{\ensuremath{ \vec{W} }}
\newcommand{\tvecX}{\ensuremath{ \vec{X} }}
\newcommand{\tvecY}{\ensuremath{ \vec{Y} }}
\newcommand{\tvecZ}{\ensuremath{ \vec{Z} }}

\newcommand{\no}{\hspace{-0.14cm}\,}

\newcommand{\lveca}{\Delta\veca}
\newcommand{\lvecb}{\Delta\vecb}
\newcommand{\lvecc}{\Delta\vecc}
\newcommand{\lvecd}{\Delta\vecd}
\newcommand{\lvece}{\Delta\vece}
\newcommand{\lvecf}{\Delta\vecf}
\newcommand{\lvecg}{\Delta\vecg}
\newcommand{\lvech}{\Delta\vech}
\newcommand{\lveci}{\Delta\veci}
\newcommand{\lvecj}{\Delta\vecj}
\newcommand{\lveck}{\Delta\veck}
\newcommand{\lvecl}{\Delta\vecl}
\newcommand{\lvecm}{\Delta\vecm}
\newcommand{\lvecn}{\Delta\vecn}
\newcommand{\lveco}{\Delta\veco}
\newcommand{\lvecp}{\Delta\vecp}
\newcommand{\lvecq}{\Delta\vecq}
\newcommand{\lvecr}{\Delta\vecr}
\newcommand{\lvecs}{\Delta\vecs}
\newcommand{\lvect}{\Delta\vect}
\newcommand{\lvecu}{\Delta\vecu}
\newcommand{\lvecv}{\Delta\vecv}
\newcommand{\lvecw}{\Delta\vecw}
\newcommand{\lvecx}{\Delta\vecx}
\newcommand{\lvecy}{\Delta\vecy}
\newcommand{\lvecz}{\Delta\vecz}

\newcommand{\lvecN}{\Delta\vecN}


\newcommand{\dveca}{\delta\veca}
\newcommand{\dvecb}{\delta\vecb}
\newcommand{\dvecc}{\delta\vecc}
\newcommand{\dvecd}{\delta\vecd}
\newcommand{\dvece}{\delta\vece}
\newcommand{\dvecf}{\delta\vecf}
\newcommand{\dvecg}{\delta\vecg}
\newcommand{\dvech}{\delta\vech}
\newcommand{\dveci}{\delta\veci}
\newcommand{\dvecj}{\delta\vecj}
\newcommand{\dveck}{\delta\veck}
\newcommand{\dvecl}{\delta\vecl}
\newcommand{\dvecm}{\delta\vecm}
\newcommand{\dvecu}{\delta\vecu}
\newcommand{\dvecv}{\delta\vecv}
\newcommand{\dvecw}{\delta\vecw}
\newcommand{\dvecx}{\delta\vecx}
\newcommand{\dvecy}{\delta\vecy}
\newcommand{\dvecz}{\delta\vecz}

\newcommand{\bveca}{\bar{\veca}}
\newcommand{\bvecb}{\bar{\vecb}}
\newcommand{\bvecc}{\bar{\vecc}}
\newcommand{\bvecd}{\bar{\vecd}}
\newcommand{\bvece}{\bar{\vece}}
\newcommand{\bvecf}{\bar{\vecf}}
\newcommand{\bvecg}{\bar{\vecg}}
\newcommand{\bvecm}{\bar{\vecm}}
\newcommand{\bvecn}{\bar{\vecn}}
\newcommand{\bveco}{\bar{\veco}}
\newcommand{\bvecp}{\bar{\vecp}}
\newcommand{\bvecq}{\bar{\vecq}}
\newcommand{\bvecr}{\bar{\vecr}}
\newcommand{\bvecs}{\bar{\vecs}}
\newcommand{\bvect}{\bar{\vect}}
\newcommand{\bvecu}{\bar{\vecu}}
\newcommand{\bvecv}{\bar{\vecv}}
\newcommand{\bvecw}{\bar{\vecw}}
\newcommand{\bvecx}{\bar{\vecx}}
\newcommand{\bvecy}{\bar{\vecy}}
\newcommand{\bvecz}{\bar{\vecz}}

\newcommand{\bvecA}{\bar{\vecA}}
\newcommand{\bvecB}{\bar{\vecB}}
\newcommand{\bvecC}{\bar{\vecC}}
\newcommand{\bvecD}{\bar{\vecD}}
\newcommand{\bvecE}{\bar{\vecE}}
\newcommand{\bvecF}{\bar{\vecF}}
\newcommand{\bvecG}{\bar{\vecG}}
\newcommand{\bvecH}{\bar{\vecH}}
\newcommand{\bvecI}{\bar{\vecI}}
\newcommand{\bvecJ}{\bar{\vecJ}}
\newcommand{\bvecK}{\bar{\vecK}}
\newcommand{\bvecL}{\bar{\vecL}}
\newcommand{\bvecM}{\bar{\vecM}}
\newcommand{\bvecN}{\bar{\vecN}}
\newcommand{\bvecO}{\bar{\vecO}}
\newcommand{\bvecP}{\bar{\vecP}}
\newcommand{\bvecR}{\bar{\vecR}}
\newcommand{\bvecS}{\bar{\vecS}}
\newcommand{\bvecT}{\bar{\vecT}}
\newcommand{\bvecU}{\bar{\vecU}}


\newcommand{\lbveca}{\Delta\bar{\veca}}
\newcommand{\lbvecb}{\Delta\bar{\vecb}}
\newcommand{\lbvecc}{\Delta\bar{\vecc}}
\newcommand{\lbvecd}{\Delta\bar{\vecd}}
\newcommand{\lbvece}{\Delta\bar{\vece}}
\newcommand{\lbvecf}{\Delta\bar{\vecf}}
\newcommand{\lbvecg}{\Delta\bar{\vecg}}
\newcommand{\lbvech}{\Delta\bar{\vech}}
\newcommand{\lbveci}{\Delta\bar{\veci}}
\newcommand{\lbvecj}{\Delta\bar{\vecj}}
\newcommand{\lbvect}{\Delta\bar{\vect}}
\newcommand{\lbvecu}{\Delta\bar{\vecu}}
\newcommand{\lbvecx}{\Delta\bar{\vecx}}


\newcommand{\dbveca}{\delta\bar{\veca}}
\newcommand{\dbvecb}{\delta\bar{\vecb}}
\newcommand{\dbvecc}{\delta\bar{\vecc}}
\newcommand{\dbvecd}{\delta\bar{\vecd}}
\newcommand{\dbvece}{\delta\bar{\vece}}
\newcommand{\dbvecf}{\delta\bar{\vecf}}
\newcommand{\dbvecg}{\delta\bar{\vecg}}
\newcommand{\dbvech}{\delta\bar{\vech}}
\newcommand{\dbveci}{\delta\bar{\veci}}
\newcommand{\dbvecj}{\delta\bar{\vecj}}
\newcommand{\dbvect}{\delta\bar{\vect}}
\newcommand{\dbvecu}{\delta\bar{\vecu}}
\newcommand{\dbvecx}{\delta\bar{\vecx}}

\newcommand{\hatveca}{\ensuremath{ \widehat{\vec{a}} }}
\newcommand{\hatvecb}{\ensuremath{ \widehat{\vec{b}} }}
\newcommand{\hatvecc}{\ensuremath{ \widehat{\vec{c}} }}
\newcommand{\hatvecd}{\ensuremath{ \widehat{\vec{d}} }}
\newcommand{\hatvece}{\ensuremath{ \widehat{\vec{e}} }}
\newcommand{\hatvecf}{\ensuremath{ \widehat{\vec{f}} }}
\newcommand{\hatvecg}{\ensuremath{ \widehat{\vec{g}} }}
\newcommand{\hatvech}{\ensuremath{ \widehat{\vec{h}} }}
\newcommand{\hatveci}{\ensuremath{ \widehat{\vec{i}} }}
\newcommand{\hatvecj}{\ensuremath{ \widehat{\vec{j}} }}
\newcommand{\hatveck}{\ensuremath{ \widehat{\vec{k}} }}
\newcommand{\hatvecl}{\ensuremath{ \widehat{\vec{l}} }}
\newcommand{\hatvecm}{\ensuremath{ \widehat{\vec{m}} }}
\newcommand{\hatvecn}{\ensuremath{ \widehat{\vec{n}} }}
\newcommand{\hatveco}{\ensuremath{ \widehat{\vec{o}} }}
\newcommand{\hatvecp}{\ensuremath{ \widehat{\vec{p}} }}
\newcommand{\hatvecq}{\ensuremath{ \widehat{\vec{q}} }}
\newcommand{\hatvecr}{\ensuremath{ \widehat{\vec{r}} }}
\newcommand{\hatvecs}{\ensuremath{ \widehat{\vec{s}} }}
\newcommand{\hatvect}{\ensuremath{ \widehat{\vec{t}} }}
\newcommand{\hatvecu}{\ensuremath{ \widehat{\vec{u}} }}
\newcommand{\hatvecv}{\ensuremath{ \widehat{\vec{v}} }}
\newcommand{\hatvecw}{\ensuremath{ \widehat{\vec{w}} }}
\newcommand{\hatvecx}{\ensuremath{ \widehat{\vec{x}} }}
\newcommand{\hatvecy}{\ensuremath{ \widehat{\vec{y}} }}
\newcommand{\hatvecz}{\ensuremath{ \widehat{\vec{z}} }}

\newcommand{\hatvecA}{\ensuremath{ \widehat{\vec{A}} }}
\newcommand{\hatvecB}{\ensuremath{ \widehat{\vec{B}} }}
\newcommand{\hatvecC}{\ensuremath{ \widehat{\vec{C}} }}
\newcommand{\hatvecD}{\ensuremath{ \widehat{\vec{D}} }}
\newcommand{\hatvecE}{\ensuremath{ \widehat{\vec{E}} }}
\newcommand{\hatvecF}{\ensuremath{ \widehat{\vec{F}} }}
\newcommand{\hatvecG}{\ensuremath{ \widehat{\vec{G}} }}
\newcommand{\hatvecH}{\ensuremath{ \widehat{\vec{H}} }}
\newcommand{\hatvecI}{\ensuremath{ \widehat{\vec{I}} }}
\newcommand{\hatvecJ}{\ensuremath{ \widehat{\vec{J}} }}
\newcommand{\hatvecK}{\ensuremath{ \widehat{\vec{K}} }}
\newcommand{\hatvecL}{\ensuremath{ \widehat{\vec{L}} }}
\newcommand{\hatvecM}{\ensuremath{ \widehat{\vec{M}} }}
\newcommand{\hatvecN}{\ensuremath{ \widehat{\vec{N}} }}
\newcommand{\hatvecO}{\ensuremath{ \widehat{\vec{O}} }}
\newcommand{\hatvecP}{\ensuremath{ \widehat{\vec{P}} }}
\newcommand{\hatvecQ}{\ensuremath{ \widehat{\vec{Q}} }}
\newcommand{\hatvecR}{\ensuremath{ \widehat{\vec{R}} }}
\newcommand{\hatvecS}{\ensuremath{ \widehat{\vec{S}} }}
\newcommand{\hatvecT}{\ensuremath{ \widehat{\vec{T}} }}
\newcommand{\hatvecU}{\ensuremath{ \widehat{\vec{U}} }}
\newcommand{\hatvecV}{\ensuremath{ \widehat{\vec{V}} }}
\newcommand{\hatvecW}{\ensuremath{ \widehat{\vec{W}} }}
\newcommand{\hatvecX}{\ensuremath{ \widehat{\vec{X}} }}
\newcommand{\hatvecY}{\ensuremath{ \widehat{\vec{Y}} }}
\newcommand{\hatvecZ}{\ensuremath{ \widehat{\vec{Z}} }}

\newcommand{\tilveca}{\ensuremath{ \widetilde{\vec{a}} }}
\newcommand{\tilvecb}{\ensuremath{ \widetilde{\vec{b}} }}
\newcommand{\tilvecc}{\ensuremath{ \widetilde{\vec{c}} }}
\newcommand{\tilvecd}{\ensuremath{ \widetilde{\vec{d}} }}
\newcommand{\tilvece}{\ensuremath{ \widetilde{\vec{e}} }}
\newcommand{\tilvecf}{\ensuremath{ \widetilde{\vec{f}} }}
\newcommand{\tilvecg}{\ensuremath{ \widetilde{\vec{g}} }}
\newcommand{\tilvech}{\ensuremath{ \widetilde{\vec{h}} }}
\newcommand{\tilveci}{\ensuremath{ \widetilde{\vec{i}} }}
\newcommand{\tilvecj}{\ensuremath{ \widetilde{\vec{j}} }}
\newcommand{\tilveck}{\ensuremath{ \widetilde{\vec{k}} }}
\newcommand{\tilvecl}{\ensuremath{ \widetilde{\vec{l}} }}
\newcommand{\tilvecm}{\ensuremath{ \widetilde{\vec{m}} }}
\newcommand{\tilvecn}{\ensuremath{ \widetilde{\vec{n}} }}
\newcommand{\tilveco}{\ensuremath{ \widetilde{\vec{o}} }}
\newcommand{\tilvecp}{\ensuremath{ \widetilde{\vec{p}} }}
\newcommand{\tilvecq}{\ensuremath{ \widetilde{\vec{q}} }}
\newcommand{\tilvecr}{\ensuremath{ \widetilde{\vec{r}} }}
\newcommand{\tilvecs}{\ensuremath{ \widetilde{\vec{s}} }}
\newcommand{\tilvect}{\ensuremath{ \widetilde{\vec{t}} }}
\newcommand{\tilvecu}{\ensuremath{ \widetilde{\vec{u}} }}
\newcommand{\tilvecv}{\ensuremath{ \widetilde{\vec{v}} }}
\newcommand{\tilvecw}{\ensuremath{ \widetilde{\vec{w}} }}
\newcommand{\tilvecx}{\ensuremath{ \widetilde{\vec{x}} }}
\newcommand{\tilvecy}{\ensuremath{ \widetilde{\vec{y}} }}
\newcommand{\tilvecz}{\ensuremath{ \widetilde{\vec{z}} }}

\newcommand{\tilvecA}{\ensuremath{ \widetilde{\vec{A}} }}
\newcommand{\tilvecB}{\ensuremath{ \widetilde{\vec{B}} }}
\newcommand{\tilvecC}{\ensuremath{ \widetilde{\vec{C}} }}
\newcommand{\tilvecD}{\ensuremath{ \widetilde{\vec{D}} }}
\newcommand{\tilvecE}{\ensuremath{ \widetilde{\vec{E}} }}
\newcommand{\tilvecF}{\ensuremath{ \widetilde{\vec{F}} }}
\newcommand{\tilvecG}{\ensuremath{ \widetilde{\vec{G}} }}
\newcommand{\tilvecH}{\ensuremath{ \widetilde{\vec{H}} }}
\newcommand{\tilvecI}{\ensuremath{ \widetilde{\vec{I}} }}
\newcommand{\tilvecJ}{\ensuremath{ \widetilde{\vec{J}} }}
\newcommand{\tilvecK}{\ensuremath{ \widetilde{\vec{K}} }}
\newcommand{\tilvecL}{\ensuremath{ \widetilde{\vec{L}} }}
\newcommand{\tilvecM}{\ensuremath{ \widetilde{\vec{M}} }}
\newcommand{\tilvecN}{\ensuremath{ \widetilde{\vec{N}} }}
\newcommand{\tilvecO}{\ensuremath{ \widetilde{\vec{O}} }}
\newcommand{\tilvecP}{\ensuremath{ \widetilde{\vec{P}} }}
\newcommand{\tilvecQ}{\ensuremath{ \widetilde{\vec{Q}} }}
\newcommand{\tilvecR}{\ensuremath{ \widetilde{\vec{R}} }}
\newcommand{\tilvecS}{\ensuremath{ \widetilde{\vec{S}} }}
\newcommand{\tilvecT}{\ensuremath{ \widetilde{\vec{T}} }}
\newcommand{\tilvecU}{\ensuremath{ \widetilde{\vec{U}} }}
\newcommand{\tilvecV}{\ensuremath{ \widetilde{\vec{V}} }}
\newcommand{\tilvecW}{\ensuremath{ \widetilde{\vec{W}} }}
\newcommand{\tilvecX}{\ensuremath{ \widetilde{\vec{X}} }}
\newcommand{\tilvecY}{\ensuremath{ \widetilde{\vec{Y}} }}
\newcommand{\tilvecZ}{\ensuremath{ \widetilde{\vec{Z}} }}

\newcommand{\tilveclam}{\ensuremath{ \widetilde{\gvec{\lambda}} }}

\newcommand{\ttvecd}{\ensuremath{ \ddot{\vec{d}} }}
\newcommand{\ttvect}{\ensuremath{ \ddot{\vec{t}} }}
\newcommand{\ttvecu}{\ensuremath{ \ddot{\vec{u}} }}
\newcommand{\ttvecv}{\ensuremath{ \ddot{\vec{v}} }}
\newcommand{\ttvecw}{\ensuremath{ \ddot{\vec{w}} }}
\newcommand{\ttvecx}{\ensuremath{ \ddot{\vec{x}} }}
\newcommand{\ttvecy}{\ensuremath{ \ddot{\vec{y}} }}
\newcommand{\ttvecz}{\ensuremath{ \ddot{\vec{z}} }}

\newcommand{\scaa}{\ensuremath{\mathrm{a}}}
\newcommand{\scab}{\ensuremath{\mathrm{b}}}
\newcommand{\scac}{\ensuremath{\mathrm{c}}}
\newcommand{\scad}{\ensuremath{\mathrm{d}}}
\newcommand{\scae}{\ensuremath{\mathrm{e}}}
\newcommand{\scaf}{\ensuremath{\mathrm{f}}}
\newcommand{\scag}{\ensuremath{\mathrm{g}}}
\newcommand{\scah}{\ensuremath{\mathrm{h}}}
\newcommand{\scai}{\ensuremath{\mathrm{i}}}
\newcommand{\scaj}{\ensuremath{\mathrm{j}}}
\newcommand{\scak}{\ensuremath{\mathrm{k}}}
\newcommand{\scal}{\ensuremath{\mathrm{l}}}
\newcommand{\scam}{\ensuremath{\mathrm{m}}}
\newcommand{\scan}{\ensuremath{\mathrm{n}}}
\newcommand{\scao}{\ensuremath{\mathrm{o}}}
\newcommand{\scap}{\ensuremath{\mathrm{p}}}
\newcommand{\scaq}{\ensuremath{\mathrm{q}}}
\newcommand{\scar}{\ensuremath{\mathrm{r}}}
\newcommand{\scas}{\ensuremath{\mathrm{s}}}
\newcommand{\scat}{\ensuremath{\mathrm{t}}}
\newcommand{\scau}{\ensuremath{\mathrm{u}}}
\newcommand{\scav}{\ensuremath{\mathrm{v}}}
\newcommand{\scaw}{\ensuremath{\mathrm{w}}}
\newcommand{\scax}{\ensuremath{\mathrm{x}}}
\newcommand{\scay}{\ensuremath{\mathrm{y}}}
\newcommand{\scaz}{\ensuremath{\mathrm{z}}}

\newcommand{\scaA}{\ensuremath{\mathrm{A}}}
\newcommand{\scaB}{\ensuremath{\mathrm{B}}}
\newcommand{\scaC}{\ensuremath{\mathrm{C}}}
\newcommand{\scaD}{\ensuremath{\mathrm{D}}}
\newcommand{\scaE}{\ensuremath{\mathrm{E}}}
\newcommand{\scaF}{\ensuremath{\mathrm{F}}}
\newcommand{\scaG}{\ensuremath{\mathrm{G}}}
\newcommand{\scaH}{\ensuremath{\mathrm{H}}}
\newcommand{\scaI}{\ensuremath{\mathrm{I}}}
\newcommand{\scaJ}{\ensuremath{\mathrm{J}}}
\newcommand{\scaK}{\ensuremath{\mathrm{K}}}
\newcommand{\scaL}{\ensuremath{\mathrm{L}}}
\newcommand{\scaM}{\ensuremath{\mathrm{M}}}
\newcommand{\scaN}{\ensuremath{\mathrm{N}}}
\newcommand{\scaO}{\ensuremath{\mathrm{O}}}
\newcommand{\scaP}{\ensuremath{\mathrm{P}}}
\newcommand{\scaQ}{\ensuremath{\mathrm{Q}}}
\newcommand{\scaR}{\ensuremath{\mathrm{R}}}
\newcommand{\scaS}{\ensuremath{\mathrm{S}}}
\newcommand{\scaT}{\ensuremath{\mathrm{T}}}
\newcommand{\scaU}{\ensuremath{\mathrm{U}}}
\newcommand{\scaV}{\ensuremath{\mathrm{V}}}
\newcommand{\scaW}{\ensuremath{\mathrm{W}}}
\newcommand{\scaX}{\ensuremath{\mathrm{X}}}
\newcommand{\scaY}{\ensuremath{\mathrm{Y}}}
\newcommand{\scaZ}{\ensuremath{\mathrm{Z}}}

\newcommand{\bscaa}{\bar{\scaa}}
\newcommand{\bscab}{\bar{\scab}}
\newcommand{\bscac}{\bar{\scac}}
\newcommand{\bscad}{\bar{\scad}}
\newcommand{\bscae}{\bar{\scae}}
\newcommand{\bscaf}{\bar{\scaf}}
\newcommand{\bscag}{\bar{\scag}}
\newcommand{\bscah}{\bar{\scah}}
\newcommand{\bscai}{\bar{\scai}}
\newcommand{\bscaj}{\bar{\scaj}}
\newcommand{\bscak}{\bar{\scak}}
\newcommand{\bscal}{\bar{\scal}}
\newcommand{\bscam}{\bar{\scam}}
\newcommand{\bscan}{\bar{\scan}}
\newcommand{\bscao}{\bar{\scao}}
\newcommand{\bscap}{\bar{\scap}}
\newcommand{\bscaq}{\bar{\scaq}}
\newcommand{\bscar}{\bar{\scar}}
\newcommand{\bscas}{\bar{\scas}}
\newcommand{\bscat}{\bar{\scat}}

\newcommand{\bscaA}{\bar{\scaA}}
\newcommand{\bscaB}{\bar{\scaB}}
\newcommand{\bscaC}{\bar{\scaC}}
\newcommand{\bscaD}{\bar{\scaD}}
\newcommand{\bscaE}{\bar{\scaE}}
\newcommand{\bscaF}{\bar{\scaF}}
\newcommand{\bscaG}{\bar{\scaG}}
\newcommand{\bscaH}{\bar{\scaH}}
\newcommand{\bscaI}{\bar{\scaI}}
\newcommand{\bscaJ}{\bar{\scaJ}}
\newcommand{\bscaK}{\bar{\scaK}}
\newcommand{\bscaL}{\bar{\scaL}}
\newcommand{\bscaM}{\bar{\scaM}}
\newcommand{\bscaN}{\bar{\scaN}}
\newcommand{\bscaO}{\bar{\scaO}}
\newcommand{\bscaP}{\bar{\scaP}}
\newcommand{\bscaQ}{\bar{\scaQ}}
\newcommand{\bscaR}{\bar{\scaR}}
\newcommand{\bscaS}{\bar{\scaS}}
\newcommand{\bscaT}{\bar{\scaT}}

\newcommand{\hatscan}{\ensuremath{\widehat{\scan}}}

\newcommand{\tilscag}{\ensuremath{\widetilde{\scag}}}
\newcommand{\tilscat}{\ensuremath{\widetilde{\scat}}}

\newcommand{\tillam}{\ensuremath{\widetilde{\lambda}}}

\newcommand{\tscag}{\ensuremath{\dot{\scag}}}

\newcommand{\lscaa}{\Delta\scaa}
\newcommand{\lscab}{\Delta\scab}
\newcommand{\lscac}{\Delta\scac}
\newcommand{\lscad}{\Delta\scad}
\newcommand{\lscae}{\Delta\scae}
\newcommand{\lscaf}{\Delta\scaf}
\newcommand{\lscag}{\Delta\scag}
\newcommand{\lscah}{\Delta\scah}
\newcommand{\lscai}{\Delta\scai}
\newcommand{\lscaj}{\Delta\scaj}
\newcommand{\lscak}{\Delta\scak}
\newcommand{\lscal}{\Delta\scal}
\newcommand{\lscam}{\Delta\scam}
\newcommand{\lscan}{\Delta\scan}
\newcommand{\lscao}{\Delta\scao}
\newcommand{\lscap}{\Delta\scap}
\newcommand{\lscaq}{\Delta\scaq}
\newcommand{\lscar}{\Delta\scar}
\newcommand{\lscas}{\Delta\scas}
\newcommand{\lscat}{\Delta\scat}

\newcommand{\lscaA}{\Delta\scaA}
\newcommand{\lscaD}{\Delta\scaD}
\newcommand{\lscaM}{\Delta\scaM}
\newcommand{\lscaN}{\Delta\scaN}

\newcommand{\balpha     }{\bar{\alpha}}
\newcommand{\bbeta      }{\bar{\beta}}
\newcommand{\bgamma     }{\bar{\gamma}}
\newcommand{\bdelta     }{\bar{\delta}}
\newcommand{\bepsilon   }{\bar{\epsilon}}
\newcommand{\bvareps    }{\bar{\varepsilon}}
\newcommand{\blambda    }{\bar{\lambda}}
\newcommand{\bxi        }{\bar{\xi}}
\newcommand{\bsigma     }{\bar{\sigma}}
\newcommand{\bvarsigma  }{\bar{\varsigma}}
\newcommand{\btau       }{\bar{\tau}}

\newcommand{\tileps     }{\widetilde{\epsilon}}
\newcommand{\tillambda  }{\widetilde{\lambda}}
\newcommand{\tilsigma   }{\widetilde{\sigma}}

\newcommand{\vecalpha     }{\ensuremath{ \gvec{\alpha} }}
\newcommand{\vecbeta      }{\ensuremath{ \gvec{\beta} }}
\newcommand{\vecgamma     }{\ensuremath{ \gvec{\gamma} }}
\newcommand{\vecdelta     }{\ensuremath{ \gvec{\delta} }}
\newcommand{\vecepsilon   }{\ensuremath{ \gvec{\epsilon} }}
\newcommand{\vecvarepsilon}{\ensuremath{ \gvec{\varepsilon} }}
\newcommand{\veczeta      }{\ensuremath{ \gvec{\zeta} }}
\newcommand{\veceta       }{\ensuremath{ \gvec{\eta} }}
\newcommand{\vectheta     }{\ensuremath{ \gvec{\theta} }}
\newcommand{\vecvartheta  }{\ensuremath{ \gvec{\vartheta} }}
\newcommand{\veciota      }{\ensuremath{ \gvec{\iota} }}
\newcommand{\veckappa     }{\ensuremath{ \gvec{\kappa} }}
\newcommand{\veclam       }{\ensuremath{ \gvec{\lambda} }}
\newcommand{\vecmu        }{\ensuremath{ \gvec{\mu} }}
\newcommand{\vecnu        }{\ensuremath{ \gvec{\nu} }}
\newcommand{\vecxi        }{\ensuremath{ \gvec{\xi} }}
\newcommand{\vecpi        }{\ensuremath{ \gvec{\pi} }}
\newcommand{\vecvarpi     }{\ensuremath{ \gvec{\varphi} }}
\newcommand{\vecrho       }{\ensuremath{ \gvec{\rho} }}
\newcommand{\vecvarrho    }{\ensuremath{ \gvec{\varrho} }}
\newcommand{\vecsigma     }{\ensuremath{ \gvec{\sigma} }}
\newcommand{\vecvarsigma  }{\ensuremath{ \gvec{\varsigma} }}
\newcommand{\vectau       }{\ensuremath{ \gvec{\tau} }}
\newcommand{\vecupsilon   }{\ensuremath{ \gvec{\upsilon} }}
\newcommand{\vecphi       }{\ensuremath{ \gvec{\phi} }}
\newcommand{\vecvarphi    }{\ensuremath{ \gvec{\varphi} }}
\newcommand{\vecchi       }{\ensuremath{ \gvec{\chi} }}
\newcommand{\vecpsi       }{\ensuremath{ \gvec{\psi} }}
\newcommand{\vecomega     }{\ensuremath{ \gvec{\omega} }}
\newcommand{\vecUpsilon   }{\ensuremath{ \gvec{\Upsilon} }}

\newcommand{\bveceps      }{\ensuremath{ \bar{\gvec{\epsilon}} }}
\newcommand{\bveceta      }{\ensuremath{ \bar{\gvec{\eta}} }}
\newcommand{\bveclam      }{\ensuremath{ \bar{\gvec{\lambda}} }}
\newcommand{\bvecsig      }{\ensuremath{ \bar{\gvec{\sigma}} }}
\newcommand{\bvecvarsigma }{\ensuremath{ \bar{\gvec{\varsigma}} }}
\newcommand{\bvectau      }{\ensuremath{ \bar{\gvec{\tau}} }}
\newcommand{\bvecupsilon  }{\ensuremath{ \bar{\gvec{\upsilon} }}}

\newcommand{\tilveceps    }{\widetilde{\vecepsilon}}
\newcommand{\tilvecsig    }{\widetilde{\vecsigma}}

\newcommand{\tveclam}{\ensuremath{ \dot{\veclam }}}

\newcommand{\lveceps}{\Delta\vecepsilon}
\newcommand{\lveclam}{\Delta\veclam}
\newcommand{\lvecsig}{\Delta\vecsigma}
\newcommand{\lvectau}{\Delta\vectau}
\newcommand{\lvecxi }{\Delta\vecxi}

\newcommand{\dveceps}{\delta\vecepsilon}
\newcommand{\dveclam}{\delta\veclam}
\newcommand{\dvecsig}{\delta\vecsigma}
\newcommand{\dvectau}{\delta\vectau}
\newcommand{\dvecxi }{\delta\vecxi}


\newcommand{\lbveclam}{\Delta\bar{\veclam}}

\newcommand{\tenalpha     }{\ensuremath{ \gten{\alpha} }}
\newcommand{\tenbeta      }{\ensuremath{ \gten{\beta} }}
\newcommand{\tengamma     }{\ensuremath{ \gten{\gamma} }}
\newcommand{\tendelta     }{\ensuremath{ \gten{\delta} }}
\newcommand{\tenepsilon   }{\ensuremath{ \gten{\epsilon} }}
\newcommand{\teneps       }{\ensuremath{ \gten{\varepsilon} }}
\newcommand{\tenzeta      }{\ensuremath{ \gten{\zeta} }}
\newcommand{\teneta       }{\ensuremath{ \gten{\eta} }}
\newcommand{\tentheta     }{\ensuremath{ \gten{\theta} }}
\newcommand{\tenvartheta  }{\ensuremath{ \gten{\vartheta} }}
\newcommand{\teniota      }{\ensuremath{ \gten{\iota} }}
\newcommand{\tenkappa     }{\ensuremath{ \gten{\kappa} }}
\newcommand{\tenlambda    }{\ensuremath{ \gten{\lambda} }}
\newcommand{\tenmu        }{\ensuremath{ \gten{\mu} }}
\newcommand{\tennu        }{\ensuremath{ \gten{\nu} }}
\newcommand{\tenxi        }{\ensuremath{ \gten{\xi} }}
\newcommand{\tenpi        }{\ensuremath{ \gten{\pi} }}
\newcommand{\tenvarpi     }{\ensuremath{ \gten{\varphi} }}
\newcommand{\tenrho       }{\ensuremath{ \gten{\rho} }}
\newcommand{\tenvarrho    }{\ensuremath{ \gten{\varrho} }}
\newcommand{\tensig       }{\ensuremath{ \gten{\sigma} }}
\newcommand{\tenvarsigma  }{\ensuremath{ \gten{\varsigma} }}
\newcommand{\tentau       }{\ensuremath{ \gten{\tau} }}
\newcommand{\tenupsilon   }{\ensuremath{ \gten{\upsilon} }}
\newcommand{\tenphi       }{\ensuremath{ \gten{\phi} }}
\newcommand{\tenvarphi    }{\ensuremath{ \gten{\varphi} }}
\newcommand{\tenchi       }{\ensuremath{ \gten{\chi} }}
\newcommand{\tenpsi       }{\ensuremath{ \gten{\psi} }}
\newcommand{\tenomega     }{\ensuremath{ \gten{\omega} }}

\newcommand{\tenOmega     }{\ensuremath{ \gten{\Omega} }}

\newcommand{\tilteneps    }{\widetilde{\teneps}}
\newcommand{\tiltensig    }{\widetilde{\tensig}}


\newcommand{\bteneps}{\ensuremath{ \bar{\teneps }}}
\newcommand{\btensig}{\ensuremath{ \bar{\tensig }}}


\newcommand{\tteneps}{\ensuremath{ \dot{\teneps }}}
\newcommand{\ttensig}{\ensuremath{ \dot{\tensig }}}


\newcommand{\ltenalpha}{\Delta\tenalpha}
\newcommand{\ltenbeta }{\Delta\tenbeta}
\newcommand{\lteneps  }{\Delta\teneps}
\newcommand{\ltensig  }{\Delta\tensig}

\newcommand{\tgamma}{\ensuremath{ \dot{\gamma} }}
\newcommand{\txi}{\ensuremath{ \dot{\xi} }}
\newcommand{\tlam}{\ensuremath{ \dot{\lambda} }}
\newcommand{\tomega}{\ensuremath{ \dot{\omega} }}

\newcommand{\lgamma}{\Delta\gamma}
\newcommand{\llambda}{\Delta\lambda}
\newcommand{\lxi}{\Delta\xi}
\newcommand{\lsigma}{\stackrel{\triangle}{\sigma}}
\newcommand{\ltau}{\stackrel{\triangle}{\tau}}

\newcommand{\hatalpha     }{\ensuremath{ \widehat{\alpha} }}
\newcommand{\hatbeta      }{\ensuremath{ \widehat{\beta} }}
\newcommand{\hatgamma     }{\ensuremath{ \widehat{\gamma} }}
\newcommand{\hatdelta     }{\ensuremath{ \widehat{\delta} }}
\newcommand{\hatepsilon   }{\ensuremath{ \widehat{\epsilon} }}
\newcommand{\hatvarepsilon}{\ensuremath{ \widehat{\varepsilon} }}
\newcommand{\hatzeta      }{\ensuremath{ \widehat{\zeta} }}
\newcommand{\hateta       }{\ensuremath{ \widehat{\eta} }}
\newcommand{\hattheta     }{\ensuremath{ \widehat{\theta} }}
\newcommand{\hatvartheta  }{\ensuremath{ \widehat{\vartheta} }}
\newcommand{\hatiota      }{\ensuremath{ \widehat{\iota} }}
\newcommand{\hatkappa     }{\ensuremath{ \widehat{\kappa} }}
\newcommand{\hatlambda    }{\ensuremath{ \widehat{\lambda} }}
\newcommand{\hatmu        }{\ensuremath{ \widehat{\mu} }}
\newcommand{\hatnu        }{\ensuremath{ \widehat{\nu} }}
\newcommand{\hatxi        }{\ensuremath{ \widehat{\xi} }}
\newcommand{\hatpi        }{\ensuremath{ \widehat{\pi} }}
\newcommand{\hatvarpi     }{\ensuremath{ \widehat{\varphi} }}
\newcommand{\hatrho       }{\ensuremath{ \widehat{\rho} }}
\newcommand{\hatvarrho    }{\ensuremath{ \widehat{\varrho} }}
\newcommand{\hatsigma     }{\ensuremath{ \widehat{\sigma} }}
\newcommand{\hatvarsigma  }{\ensuremath{ \widehat{\varsigma} }}
\newcommand{\hattau       }{\ensuremath{ \widehat{\tau} }}
\newcommand{\hatupsilon   }{\ensuremath{ \widehat{\upsilon} }}
\newcommand{\hatphi       }{\ensuremath{ \widehat{\phi} }}
\newcommand{\hatvarphi    }{\ensuremath{ \widehat{\varphi} }}
\newcommand{\hatchi       }{\ensuremath{ \widehat{\chi} }}
\newcommand{\hatpsi       }{\ensuremath{ \widehat{\psi} }}
\newcommand{\hatomega     }{\ensuremath{ \widehat{\omega} }}

\newcommand{\hatteneps}{\ensuremath{ \widehat{\teneps} }}
\newcommand{\hattensig}{\ensuremath{ \widehat{\tensig} }}

\newcommand{\ionesi}{\scaI\utensig}
\newcommand{\itwosi}{\scaI\scaI\utensig}
\newcommand{\ithrsi}{\scaI\scaI\scaI\utensig}
\newcommand{\itwos}{\scaI\scaI\utens}
\newcommand{\ithrs}{\scaI\scaI\scaI\utens}

\newcommand{\onetwo}{\frac{1}{2}}
\newcommand{\thrtwo}{\frac{3}{2}}
\newcommand{\onethr}{\frac{1}{3}}
\newcommand{\twothr}{\frac{2}{3}}
\newcommand{\forthr}{\frac{4}{3}}
\newcommand{\onefor}{\frac{1}{4}}
\newcommand{\onesix}{\frac{1}{6}}
\newcommand{\oneeig}{\frac{1}{8}}
\newcommand{\onenin}{\frac{1}{9}}
\newcommand{\onetwe}{\frac{1}{12}}

\newcommand{\tengf}{\teng^{\flat}}
\newcommand{\tengs}{\teng^{\sharp}}

\newcommand{\Lin}{^{Lin}}

\newcommand{\uscan}{_{\mbox{\tiny{N}}}}

\newcommand{\ena}{\ensuremath{^{n+1}}}
\newcommand{\sena}{\ensuremath{^{1\,n+1}}}
\newcommand{\mena}{\ensuremath{^{2\,n+1}}}

\newcommand{\ea}{^{\alpha}}
\newcommand{\eb}{^{\beta}}
\newcommand{\ec}{^{\gamma}}
\newcommand{\ed}{^{\delta}}
\newcommand{\ex}{^{\xi}}

\newcommand{\eat}{^{\alpha T}}
\newcommand{\ebt}{^{\beta T}}
\newcommand{\ect}{^{\gamma T}}
\newcommand{\edt}{^{\delta T}}
\newcommand{\eet}{^{\epsilon T}}

\newcommand{\eaa}{^{\alpha\alpha}}
\newcommand{\eab}{^{\alpha\beta}}
\newcommand{\eac}{^{\alpha\gamma}}
\newcommand{\eba}{^{\beta\alpha}}
\newcommand{\ebc}{^{\beta\gamma}}
\newcommand{\ebd}{^{\beta\delta}}
\newcommand{\ecb}{^{\gamma\beta}}
\newcommand{\ecd}{^{\gamma\delta}}
\newcommand{\edb}{^{\delta\beta}}
\newcommand{\ede}{^{\delta\epsilon}}
\newcommand{\eec}{^{\epsilon\gamma}}
\newcommand{\exx}{^{\xi\xi}}

\newcommand{\ua}{_{\alpha}}
\newcommand{\ub}{_{\beta}}
\newcommand{\uc}{_{\gamma}}
\newcommand{\ud}{_{\delta}}
\newcommand{\ue}{_{\epsilon}}
\newcommand{\ux}{_{\xi}}

\newcommand{\uaa}{_{\alpha\alpha}}
\newcommand{\uab}{_{\alpha\beta}}
\newcommand{\uac}{_{\alpha\gamma}}
\newcommand{\uba}{_{\beta\alpha}}
\newcommand{\ubb}{_{\beta\beta}}
\newcommand{\ubc}{_{\beta\gamma}}
\newcommand{\ubd}{_{\beta\delta}}
\newcommand{\ucd}{_{\gamma\delta}}
\newcommand{\ucb}{_{\gamma\beta}}
\newcommand{\ueb}{_{\epsilon\beta}}
\newcommand{\ued}{_{\epsilon\delta}}
\newcommand{\udb}{_{\delta\beta}}
\newcommand{\uta}{_{{\mbox{\tiny{T}}}\alpha}}
\newcommand{\utb}{_{{\mbox{\tiny{T}}}\beta}}
\newcommand{\utc}{_{{\mbox{\tiny{T}}}\gamma}}
\newcommand{\uxx}{_{\xi\xi}}

\newcommand{\uka}{_{,\alpha}}
\newcommand{\ukb}{_{,\beta}}
\newcommand{\ukc}{_{,\gamma}}
\newcommand{\ukx}{_{,\xi}}

\newcommand{\uakb}{_{\alpha,\beta}}
\newcommand{\uakc}{_{\alpha,\gamma}}
\newcommand{\ubkc}{_{\beta,\gamma}}
\newcommand{\ubkd}{_{\beta,\delta}}
\newcommand{\ubke}{_{\beta,\epsilon}}
\newcommand{\uckd}{_{\gamma,\delta}}
\newcommand{\udke}{_{\delta,\epsilon}}

\newcommand{\ukaa}{_{,\alpha\alpha}}
\newcommand{\ukab}{_{,\alpha\beta}}
\newcommand{\ukba}{_{,\beta\alpha}}
\newcommand{\ukbb}{_{,\beta\beta}}
\newcommand{\ukbc}{_{,\beta\gamma}}
\newcommand{\ukxx}{_{,\xi\xi}}
\newcommand{\uxkx}{_{\xi,\xi}}
\newcommand{\uxkxx}{_{\xi,\xi\xi}}

\newcommand{\ubkcd}{_{\beta,\gamma\delta}}

\newcommand{\uga}{_{g\alpha}}
\newcommand{\ugb}{_{g\beta}}
\newcommand{\ugc}{_{g\gamma}}
\newcommand{\ugd}{_{g\delta}}
\newcommand{\ugka}{_{g,\alpha}}
\newcommand{\ugkb}{_{g,\beta}}
\newcommand{\ugkc}{_{g,\gamma}}
\newcommand{\ugkx}{_{g,\xi}}
\newcommand{\ugakb}{_{g\alpha,\beta}}
\newcommand{\ugbkc}{_{g\beta,\gamma}}

\newcommand{\uana}{_{\alpha\,n+1}}
\newcommand{\ubna}{_{\beta\,n+1}}

\newcommand{\ukana}{_{,\alpha\,n+1}}
\newcommand{\ukano}{_{,\alpha\,n}}

\newcommand{\uano}{_{\alpha\,n}}
\newcommand{\ubno}{_{\beta\,n}}

\newcommand{\umN}{_{\mbox{\tiny{N}}}}
\newcommand{\umT}{_{\mbox{\tiny{T}}}}

\newcommand{\utenb}{_{\tenb}}
\newcommand{\utenp}{_{\tenp}}
\newcommand{\utens}{_{\tens}}
\newcommand{\utenC}{_{\tenC}}
\newcommand{\utenE}{_{\tenE}}

\newcommand{\utensna}{_{\tens\,n+1}}

\newcommand{\uteneps}{_{\teneps}}
\newcommand{\utenepse}{_{\teneps^e}}
\newcommand{\utenepsp}{_{\teneps^p}}
\newcommand{\utensig}{_{\tensig}}
\newcommand{\utensigsig}{_{\tensig\tensig}}

\newcommand{\utenepsna}{_{\teneps\,n+1}}
\newcommand{\utensigna}{_{\tensig\,n+1}}

\newcommand{\gvecx}{\grave{\vecx}}


\newcommand{\mska}{_{s,\alpha}}
\newcommand{\mskb}{_{s,\beta}}
\newcommand{\mskc}{_{s,\gamma}}
\newcommand{\mfka}{_{f,\alpha}}
\newcommand{\mfkb}{_{f,\beta}}
\newcommand{\mfkc}{_{f,\gamma}}


\newcommand{\dmska}{_{Ag,\alpha}}
\newcommand{\dmskb}{_{Ag,\beta}}
\newcommand{\dmskc}{_{Ag,\gamma}}
\newcommand{\dmfka}{_{Ag,\alpha}}
\newcommand{\dmfkb}{_{Ag,\beta}}
\newcommand{\dmfkc}{_{Ag,\gamma}}

\newcommand{\gmska}{_{g,\alpha}}
\newcommand{\gmskb}{_{g,\beta}}
\newcommand{\gmskc}{_{g,\gamma}}
\newcommand{\gmfka}{_{g,\alpha}}
\newcommand{\gmfkb}{_{g,\beta}}
\newcommand{\gmfkc}{_{g,\gamma}}

\newcommand{\sumgp}{\sum_{g=1}^{n_{gp}}}
\newcommand{\sumni}{\sum_{I=1}^{n_{I}}}
\newcommand{\sumnj}{\sum_{J=1}^{n_{J}}}
\newcommand{\sumseg}{\sum_{seg}}
\newcommand{\sumel}{\sum_{e=1}^{n_{el}}}

\newcommand{\suma}{\sum_{\alpha=1}^{n_{\alpha}}}

\newcommand{\gng}{\scag_{N\,g}}
\newcommand{\gtg}{g_{T\,g}}
\newcommand{\lng}{\lambda_{N\,g}}
\newcommand{\ltg}{\lambda_{T\,g}}
\newcommand{\ttg}{t_{T\,g}}

\renewcommand{\d}[1]{\text{$\hspace{0.1cm}$d $\hspace{-0.11cm}#1$}}
\newcommand{\del}{\ensuremath{\partial}}
\newcommand{\divx}[1]{\text{$\hspace{0.1cm}$div$\left(#1\right)$}}
\newcommand{\divX}[1]{\text{$\hspace{0.1cm}$Div$\left(#1\right)$}}
\newcommand{\grad}[1]{\ensuremath{ \boldsymbol{\nabla}{#1}}}
\newcommand{\gradx}[1]{\ensuremath{ \boldsymbol{\nabla}_{\vecx}{#1}}}
\newcommand{\gradX}[1]{\ensuremath{ \boldsymbol{\nabla}_{\vecX}{#1}}}
\newcommand{\parder}[2]{\ensuremath{ \frac{\del #1}{\del #2} }}
\newcommand{\tder}[1]{\ensuremath{ \frac{\d{#1}}{\d{} \hspace{0.05cm}{t}} }}
\newcommand{\dx}{\ensuremath{ \d{\vecx} }}
\newcommand{\dX}{\ensuremath{ \d{\vecX} }}
\newcommand{\da}{\ensuremath{ \d{a} }}
\newcommand{\dA}{\ensuremath{ \d{A} }}
\newcommand{\dv}{\ensuremath{ \d{v} }}
\newcommand{\dV}{\ensuremath{ \d{V} }}
\newcommand{\dxis}{\ensuremath{ \d{\xi} }}
\newcommand{\lda}{\stackrel{\triangle}{\da}}

\newcommand{\dr}{\ensuremath{ \d{r} }}
\newcommand{\dphi}{\ensuremath{ \d{\phi} }}
\newcommand{\dz}{\ensuremath{\d{z}}}
\newcommand{\du}{\ensuremath{\d{u}}}
\newcommand{\dy}{\ensuremath{\d{y}}}
\newcommand{\dxscal}{\ensuremath{\d{x}}}

\renewcommand{\cos}[1]{ \text{cos}\hspace{0.0cm}\left( {#1} \right) }
\renewcommand{\sin}[1]{ \text{sin}\hspace{0.0cm}\left( {#1} \right) }
\renewcommand{\ln}[1]{\text{$\hspace{0.1cm}$ln$\left(#1\right)$}}
\renewcommand{\exp}[1]{\ensuremath{ \,\text{exp}{\left( #1 \right)} }}

\newcommand{\define}{\ensuremath{\stackrel{\mathrm{def}}{=}}}
\newcommand{\p}{\ensuremath{ ^{\prime} }}
\newcommand{\pp}{\ensuremath{ ^{\prime \prime} }}
\newcommand{\first}{$\ensuremath{ 1^{\text{st}} }\,$}
\newcommand{\second}{$\ensuremath{ 2^{\text{nd}} }\,$}
\newcommand{\lin}[1]{\ensuremath{ \calL[#1] }}

\newcommand{\tenfour}[1]{ \ensuremath {\boldsymbol{\mathcal{#1}} } }
\newcommand{\tr}[1]{\text{$\hspace{0.1cm}$tr$\left(#1\right)$}}
\newcommand{\dev}{\ensuremath{ ^{\prime} }}
\renewcommand{\det}[1]{\text{$\hspace{0.1cm}$det$\left(#1\right)$}}
\newcommand{\inv}{\ensuremath{ ^{-1} }}

\renewcommand{\it}{\ensuremath{ ^{-T} }}
\newcommand{\sym}{\ensuremath{ ^{\text{sym}} }}
\newcommand{\skw}{\ensuremath{ ^{\text{skw}} }}
\newcommand{\adj}{\ensuremath{ ^{\sharp}  }}
\newcommand{\mg}[1]{\ensuremath{ \left\| #1 \right\| }}
\newcommand{\mgv}[1]{\ensuremath{ \left| #1 \right| }}
\newcommand{\s}{\ensuremath{ ^{2}  }}
\newcommand{\ione}[1]{\ensuremath{ I_{#1} }}
\newcommand{\itwo}[1]{\ensuremath{ I\hspace{-0.1cm}I_{#1} }}
\newcommand{\ithree}[1]{\ensuremath{ I\hspace{-0.1cm}I\hspace{-0.1cm}I_{#1} }}
\newcommand{\ionep}[1]{\ensuremath{ I_{#1}\p }}
\newcommand{\itwop}[1]{\ensuremath{ I\hspace{-0.1cm}I_{#1}\p }}
\newcommand{\ithreep}[1]{\ensuremath{ I\hspace{-0.1cm}I\hspace{-0.1cm}I_{#1}\p }}
\newcommand{\ionepp}[1]{\ensuremath{ I_{#1}\pp }}
\newcommand{\itwopp}[1]{\ensuremath{ I\hspace{-0.1cm}I_{#1}\pp }}
\newcommand{\ithreepp}[1]{\ensuremath{ I\hspace{-0.1cm}I\hspace{-0.1cm}I_{#1}\pp }}

\newcommand{\gus}{\frac{\partial\scag}{\partial\tensig}}
\newcommand{\guss}{\frac{\partial^2\scag}{\partial^2\tensig}}

\newcommand{\percent}{\ensuremath{ \%  }}
\newcommand{\IE}{\ensuremath{I\hspace{-0.12cm}E  }}
\newcommand{\II}{\ensuremath{1\hspace{-0.12cm}1  }}
\newcommand{\tenIE}{\ensuremath{ \ten{\IE}  }}
\newcommand{\tenII}{\ensuremath{ \ten{\II}  }}
\newcommand{\hattenIE}{\ensuremath{ \widehat{\tenIE}  }}
\newcommand{\IC}{\ensuremath{C\hspace{-0.22cm}C  }}
\newcommand{\cc}{\ensuremath{c\hspace{-0.22cm}c  }}
\newcommand{\tenIC}{\ensuremath{ \ten{\IC}  }}
\newcommand{\tencc}{\ensuremath{ \ten{\cc}  }}
\newcommand{\lsup}[1]{\ensuremath{ {}^{[#1]} \hspace{-0.05cm} }}
\newcommand{\lsub}[1]{\ensuremath{ {}_{[#1]} \hspace{-0.05cm} }}
\newcommand{\lsupb}[2]{\ensuremath{ {}^{#1}_{#2} \hspace{-0.05cm} }}
\newcommand{\Phihat}{\ensuremath{ \widehat{\Phi} }}
\newcommand{\Psihat}{\ensuremath{ \widehat{\Psi} }}
\renewcommand{\s}{\ensuremath{ ^*} }
\renewcommand{\k}{\ensuremath{ ^K} }
\newcommand{\kp}{\ensuremath{ ^{K+1}} }
\newcommand{\km}{\ensuremath{ ^{K-1}} }
\newcommand{\tol}[1]{\ensuremath{ \text{TOL}_{#1} } }
\newcommand{\pen}{\ensuremath{ \calK }}
\newcommand{\tenone}{\ensuremath{ \ten{1} }}
\newcommand{\err}{\ensuremath{ \pounds }}
\newcommand{\li}{\ensuremath{\left[\hspace{-0.15cm}\left[ }}
\newcommand{\ri}{\ensuremath{\right]\hspace{-0.15cm}\right] }}
\renewcommand{\t}{\ensuremath{ ^{\scaT} }}
\newcommand{\tria}{\ensuremath{ ^{tr} }}

\newcommand{\iso}[1]{\ensuremath{ \widetilde{#1} }}
\newcommand{\tenFisox}{\ensuremath{ \tenF_{\vecx} } }
\newcommand{\tenFisoX}{\ensuremath{ \tenF_{\vecX} } }
\newcommand{\tenCisox}{\ensuremath{ \tenC_{\vecx} } }
\newcommand{\tenCisoX}{\ensuremath{ \tenC_{\vecX} } }
\newcommand{\Jisox}{\ensuremath{ J_{\vecx} } }
\newcommand{\JisoX}{\ensuremath{ J_{\vecX} } }
\newcommand{\calRiso}{ \ensuremath{ \iso{\calR}}}
\newcommand{\dzeta}{\ensuremath{ \d{\veczeta} } }
\newcommand{\daiso}{\ensuremath{ \d{\iso{a}} }}
\newcommand{\dviso}{\ensuremath{ \d{\iso{v}} }}
\newcommand{\vecniso}{\ensuremath{ \iso{\vecn} } }

\newcommand{\dgn}{\delta \bar{g}_{\mbox{\tiny{N}}}}
\newcommand{\lgn}{\Delta \bar{g}_{\mbox{\tiny{N}}}}
\newcommand{\ldgn}{\Delta\delta \bar{g}_{\mbox{\tiny{N}}}}

\newcommand{\dbxi}{\delta \bar{\xi}}
\newcommand{\lbxi}{\Delta \bar{\xi}}
\newcommand{\ldbxi}{\Delta\delta \bar{\xi}}

\newcommand{\intbco}{\int\limits_{\varphi\left(B^1\right)}}
\newcommand{\intbct}{\int\limits_{\varphi\left(B^2\right)}}
\newcommand{\intbcc}{\int\limits_{\varphi\left(B_c\right)}}
\newcommand{\intpbco}{\int\limits_{\varphi\left(\partial B^1\right)}}
\newcommand{\intpbct}{\int\limits_{\varphi\left(\partial B^2\right)}}
\newcommand{\intpbcon}{\int\limits_{\varphi\left(\partial B^1_n\right)}}
\newcommand{\intpbctn}{\int\limits_{\varphi\left(\partial B^2_n\right)}}
\newcommand{\intpbcc}{\int\limits_{\Gamma_c}}
\newcommand{\intbro}{\int\limits_{B^1}}
\newcommand{\intbrt}{\int\limits_{B^2}}
\newcommand{\intbrc}{\int\limits_{B_c}}

\newcommand{\intprco}{\int\limits_{\partial B^1}}
\newcommand{\intprct}{\int\limits_{\partial B^2}}
\newcommand{\intprcon}{\int\limits_{\partial B^1_n}}
\newcommand{\intprctn}{\int\limits_{\partial B^2_n}}
\newcommand{\intprcc}{\int\limits_{\partial B_c}}

\newcommand{\mnab}{n_{AB}}
\newcommand{\mnac}{n_{AC}}
\newcommand{\mnad}{n_{AD}}
\newcommand{\mnae}{n_{AE}}

\newcommand{\hthr}{\frac{h}{3}}
\newcommand{\hsix}{\frac{h}{6}}


\newcommand{\pna}{_{p\,n+1}}
\newcommand{\pno}{_{p\,n}}
\newcommand{\Ina}{_{I\,n+1}}
\newcommand{\Ino}{_{I\,n}}
\newcommand{\Jna}{_{J\,n+1}}
\newcommand{\Jno}{_{J\,n}}
\newcommand{\sal}{^{1\,\alpha}}
\newcommand{\sbl}{^{1\,\beta}}
\newcommand{\mal}{^{2\,\alpha}}
\newcommand{\xisg}{\left(\xi^1_{g\,n+1}\right)}
\newcommand{\ximg}{\left(\xi^2_{g\,n+1}\right)}
\newcommand{\xisp}{\left(\xi^1_{p\,n+1}\right)}
\newcommand{\ximp}{\left(\xi^2_{p\,n+1}\right)}
\newcommand{\xisgo}{\left(\xi^1_{g\,n}\right)}
\newcommand{\xiso}{\left(\xi^1_{p\,n}\right)}
\newcommand{\ximo}{\left(\xi^2_{p\,n}\right)}
\newcommand{\xise}{\left(\xi^1_g\left(\eta\right)\right)}
\newcommand{\xime}{\left(\xi^2_g\left(\eta\right)\right)}
\newcommand{\xipe}{\left(\xi^1_p\left(\eta\right)\right)}
\newcommand{\xisa}{\left(\xi^1_A\right)}
\newcommand{\xima}{\left(\xi^2_A\right)}
\newcommand{\xisq}{\left(\xi^1_Q\right)}
\newcommand{\ximq}{\left(\xi^2_Q\right)}

\newcommand{\etag}{\left(\eta\right)}
\newcommand{\xig}{\left(\xi_g\right)}

\newcommand{\xisf}{\left(\xi^1_1\right)}
\newcommand{\ximf}{\left(\xi^2_1\right)}
\newcommand{\xiss}{\left(\xi^1_2\right)}
\newcommand{\xims}{\left(\xi^2_2\right)}

\newcommand{\ana}{\ensuremath{ _{\alpha\,n+1} }}
\newcommand{\bna}{\ensuremath{ _{\beta\,n+1} }}

\newcommand{\sna}{\ensuremath{ _{s\,n+1} }}
\newcommand{\nna}{\ensuremath{ _{\mbox{\tiny{N}}\,n+1} }}
\newcommand{\tna}{\ensuremath{ _{\mbox{\tiny{T}}\,n+1} }}
\newcommand{\tnaa}{\ensuremath{_{\mbox{\tiny{T}}\alpha\,n+1\,}}}
\newcommand{\tnab}{\ensuremath{_{\mbox{\tiny{T}}\,n+1\,\beta}}}
\newcommand{\tnax}{\ensuremath{_{\mbox{\tiny{T}}\,n+1\,\xi}}}

\newcommand{\nA}{\ensuremath{ _{\mbox{\tiny{N}} A} }}
\newcommand{\Ng}{\ensuremath{ _{\mbox{\tiny{N}} g} }}
\newcommand{\tA}{\ensuremath{ _{\mbox{\tiny{T}} A} }}
\newcommand{\ta}{\ensuremath{ _{\mbox{\tiny{T}} \alpha} }}
\newcommand{\Tag}{\ensuremath{ _{\mbox{\tiny{T}} \alpha g} }}
\newcommand{\taA}{\ensuremath{ _{\mbox{\tiny{T}}\alpha A} }}
\newcommand{\tbA}{\ensuremath{ _{\mbox{\tiny{T}}\beta A} }}

\newcommand{\tana}{\ensuremath{ _{\mbox{\tiny{T}} A\,n+1} }}

\newcommand{\taana}{\ensuremath{ _{\mbox{\tiny{T}}\alpha A\,n+1} }}
\newcommand{\tbana}{\ensuremath{ _{\mbox{\tiny{T}}\beta A\,n+1} }}

\newcommand{\taano}{\ensuremath{ _{\mbox{\tiny{T}}\alpha A\,n} }}

\newcommand{\iana}{\ensuremath{ _{i A\,n+1} }}
\newcommand{\nana}{\ensuremath{ _{\mbox{\tiny{N}} A\,n+1} }}
\newcommand{\nano}{\ensuremath{ _{\mbox{\tiny{N}} A\,n} }}
\newcommand{\dana}{\ensuremath{ _{\mbox{\tiny{D}} A\,n+1} }}
\newcommand{\tno}{\ensuremath{ _{\mbox{\tiny{T}}\,n} }}
\newcommand{\tano}{\ensuremath{ _{\mbox{\tiny{T}}A\,n} }}
\newcommand{\na}{\ensuremath{ _{n+1} }}
\newcommand{\ina}{\ensuremath{ _{i\,n+1} }}
\newcommand{\jna}{\ensuremath{ _{j\,n+1} }}

\newcommand{\gna}{\ensuremath{ _{g\,n+1} }}
\newcommand{\qna}{\ensuremath{ _{q\,n+1} }}
\newcommand{\gno}{\ensuremath{ _{g\,n} }}

\newcommand{\Ana}{\ensuremath{ _{A\,n+1} }}
\newcommand{\Bna}{\ensuremath{ _{B\,n+1} }}
\newcommand{\Cna}{\ensuremath{ _{C\,n+1} }}
\newcommand{\Ano}{\ensuremath{ _{A\,n} }}
\newcommand{\Bno}{\ensuremath{ _{B\,n} }}
\newcommand{\Cno}{\ensuremath{ _{C\,n} }}

\newcommand{\nBna}{\ensuremath{ _{\mbox{\tiny{N}}B\,n+1} }}
\newcommand{\tBna}{\ensuremath{ _{\mbox{\tiny{T}}B\,n+1} }}

\newcommand{\xina}{\left(\vecxi_{n+1}\right)}
\newcommand{\xino}{\left(\vecxi_{n}\right)}
\newcommand{\xigna}{\left(\vecxi_{g\,n+1}\right)}
\newcommand{\xigno}{\left(\vecxi_{g\,n}\right)}

\newcommand{\agna}{\ensuremath{ _{\alpha\,g\,n+1} }}
\newcommand{\kagna}{\ensuremath{ _{,\alpha\,g\,n+1} }}
\newcommand{\agno}{\ensuremath{ _{\alpha\,g\,n} }}
\newcommand{\kagno}{\ensuremath{ _{,\alpha\,g\,n} }}

\newcommand{\ogna}{\ensuremath{ _{1\,g\,n+1} }}
\newcommand{\kogna}{\ensuremath{ _{,1\,g\,n+1} }}
\newcommand{\ogno}{\ensuremath{ _{1\,g\,n} }}
\newcommand{\kogno}{\ensuremath{ _{,1\,g\,n} }}

\newcommand{\tgna}{\ensuremath{ _{2\,g\,n+1} }}
\newcommand{\ktgna}{\ensuremath{ _{,2\,g\,n+1} }}
\newcommand{\tgno}{\ensuremath{ _{2\,g\,n} }}
\newcommand{\ktgno}{\ensuremath{ _{,2\,g\,n} }}

\newcommand{\aAna}{\ensuremath{ _{\alpha\,A\,n+1} }}
\newcommand{\bAna}{\ensuremath{ _{\beta\,A\,n+1} }}
\newcommand{\aBna}{\ensuremath{ _{\alpha\,B\,n+1} }}

\newcommand{\nnAna}{\ensuremath{ _{33\,A\,n+1} }}
\newcommand{\aaAna}{\ensuremath{ _{\alpha\alpha\,A\,n+1} }}
\newcommand{\abAna}{\ensuremath{ _{12\,A\,n+1} }}
\newcommand{\anAna}{\ensuremath{ _{\alpha3\,A\,n+1} }}

\newcommand{\iAna}{\ensuremath{ _{i\,A\,n+1} }}
\newcommand{\jAna}{\ensuremath{ _{j\,A\,n+1} }}
\newcommand{\kAna}{\ensuremath{ _{k\,A\,n+1} }}
\newcommand{\kBna}{\ensuremath{ _{k\,B\,n+1} }}

\newcommand{\dthetc}{\delta\bar{\vartheta}^1}
\newcommand{\dthets}{\delta\vartheta^2}
\newcommand{\dphic}{\delta\bar{\varphi}^1}
\newcommand{\dphis}{\delta\varphi^2}

\newcommand{\Dthetc}{\Delta\bar{\vartheta}^1}
\newcommand{\Dthets}{\Delta\vartheta^2}
\newcommand{\Dphic}{\Delta\bar{\varphi}^1}
\newcommand{\Dphis}{\Delta\varphi^2}

\newcommand{\dxi}{\delta\bar{\xi}}
\newcommand{\dxia}{\delta\bar{\xi}^{\alpha}}
\newcommand{\dxib}{\delta\bar{\xi}^{\beta}}
\newcommand{\dxic}{\delta\bar{\xi}^{\gamma}}
\newcommand{\dxid}{\delta\bar{\xi}^{\delta}}
\newcommand{\dxie}{\delta\bar{\xi}^{\epsilon}}
\newcommand{\dxif}{\delta\bar{\xi}^{\eta}}

\newcommand{\Dxi}{\Delta\bar{\xi}}
\newcommand{\Dxia}{\Delta\bar{\xi}^{\alpha}}
\newcommand{\Dxib}{\Delta\bar{\xi}^{\beta}}
\newcommand{\Dxic}{\Delta\bar{\xi}^{\gamma}}
\newcommand{\Dxid}{\Delta\bar{\xi}^{\delta}}
\newcommand{\Dxie}{\Delta\bar{\xi}^{\epsilon}}
\newcommand{\Dxif}{\Delta\bar{\xi}^{\eta}}

\newcommand{\Ddxi}{\Delta\delta\bar{\xi}}
\newcommand{\Ddxia}{\Delta\delta\bar{\xi}^{\alpha}}
\newcommand{\Ddxib}{\Delta\delta\bar{\xi}^{\beta}}
\newcommand{\Ddxic}{\Delta\delta\bar{\xi}^{\gamma}}
\newcommand{\Ddxid}{\Delta\delta\bar{\xi}^{\delta}}
\newcommand{\Ddxie}{\Delta\delta\bar{\xi}^{\epsilon}}
\newcommand{\Ddxif}{\Delta\delta\bar{\xi}^{\eta}}

\newcommand{\dvecaa}{\delta\bar{\veca}_{\alpha}^1}
\newcommand{\dvecab}{\delta\bar{\veca}_{\beta}^1}

\newcommand{\dvecaat}{\delta\bar{\veca}^{1\alpha}}
\newcommand{\dvecabt}{\delta\bar{\veca}^{1\beta}}

\newcommand{\Dvecaa}{\Delta\bar{\veca}_{\alpha}^1}
\newcommand{\Dvecab}{\Delta\bar{\veca}_{\beta}^1}
\newcommand{\Dvecac}{\Delta\bar{\veca}_{\gamma}^1}

\newcommand{\Dvecaat}{\Delta\bar{\veca}^{1\alpha}}
\newcommand{\Dvecabt}{\Delta\bar{\veca}^{1\beta}}

\newcommand{\dvecn}{\delta\bar{\vecn}^1}
\newcommand{\Dvecn}{\Delta\bar{\vecn}^1}
\newcommand{\Ddvecn}{\Delta\delta\bar{\vecn}^1}

\newcommand{\dvecua}{\delta\bar{\vecu}_{,\alpha}^2}
\newcommand{\dvecub}{\delta\bar{\vecu}_{,\beta}^2}
\newcommand{\dvecuc}{\delta\bar{\vecu}_{,\gamma}^2}
\newcommand{\dvecud}{\delta\bar{\vecu}_{,\vartheta}^2}
\newcommand{\dvecue}{\delta\bar{\vecu}_{,\theta}^2}

\newcommand{\Dvecua}{\Delta\bar{\vecu}_{,\alpha}^2}
\newcommand{\Dvecub}{\Delta\bar{\vecu}_{,\beta}^2}
\newcommand{\Dvecuc}{\Delta\bar{\vecu}_{,\gamma}^2}
\newcommand{\Dvecud}{\Delta\bar{\vecu}_{,\vartheta}^2}
\newcommand{\Dvecue}{\Delta\bar{\vecu}_{,\theta}^2}

\newcommand{\dvecuab}{\delta\bar{\vecu}_{,\alpha\beta}^2}
\newcommand{\dvecuac}{\delta\bar{\vecu}_{,\alpha\gamma}^2}
\newcommand{\dvecubc}{\delta\bar{\vecu}_{,\beta\gamma}^2}
\newcommand{\dvecuad}{\delta\bar{\vecu}_{,\alpha\vartheta}^2}
\newcommand{\dvecuae}{\delta\bar{\vecu}_{,\alpha\theta}^2}

\newcommand{\Dvecuab}{\Delta\bar{\vecu}_{,\alpha\beta}^2}
\newcommand{\Dvecuac}{\Delta\bar{\vecu}_{,\alpha\gamma}^2}
\newcommand{\Dvecubc}{\Delta\bar{\vecu}_{,\beta\gamma}^2}
\newcommand{\Dvecuad}{\Delta\bar{\vecu}_{,\alpha\vartheta}^2}
\newcommand{\Dvecuae}{\Delta\bar{\vecu}_{,\alpha\theta}^2}

\newcommand{\dvecus}{\delta\vecu^1}
\newcommand{\Dvecus}{\Delta\vecu^1}
\newcommand{\vecxa}{\bar{\vecx}_{,\alpha}^2}
\newcommand{\vecxb}{\bar{\vecx}_{,\beta}^2}
\newcommand{\vecxab}{\bar{\vecx}_{,\alpha\beta}^2}

\newcommand{\gthet}{g_{\vartheta}}
\newcommand{\dgthet}{\delta g_{\vartheta}}
\newcommand{\Dgthet}{\Delta g_{\vartheta}}

\newcommand{\thetg}{\theta_G}
\newcommand{\dthetg}{\delta\theta_G}
\newcommand{\Dthetg}{\Delta\theta_G}

\newcommand{\gphi}{g_{\varphi}}
\newcommand{\dgphi}{\delta g_{\varphi}}
\newcommand{\Dgphi}{\Delta g_{\varphi}}

\newcommand{\tta}{t_{T\alpha}}
\newcommand{\ttb}{t_{T\beta}}
\newcommand{\ttc}{t_{T\gamma}}

\newcommand{\ttx}{t_{T\xi}}

\newcommand{\ttat}{t_{T}^{\alpha}}

\newcommand{\Dtta}{\Delta t_{T\alpha}}
\newcommand{\Dttb}{\Delta t_{T\beta}}
\newcommand{\Dttc}{\Delta t_{T\gamma}}

\newcommand{\ttra}{t_{t\alpha}^{tr}}
\newcommand{\ttrb}{t_{t\beta}^{tr}}
\newcommand{\ttrc}{t_{t\gamma}^{tr}}

\newcommand{\ttrat}{t_{T}^{trial\alpha}}
\newcommand{\ttrbt}{t_{T}^{trial\beta}}
\newcommand{\ttrct}{t_{T}^{trial\gamma}}
\newcommand{\ttrdt}{t_{T}^{trial\vartheta}}

\newcommand{\Dttra}{\Delta t_{T\alpha}^{trial}}
\newcommand{\Dttrb}{\Delta t_{T\beta}^{trial}}
\newcommand{\Dttrc}{\Delta t_{T\gamma}^{trial}}

\newpage
\section{Introduction}\label{sec:intro}

In automotive engineering, engineers must rely on datasets to accurately parameterize and validate the generalization capabilities of drivetrain simulation models. 
Ideally, these datasets should cover the variance of drivetrain hardware characteristics across the vehicle fleet and the broad variety of driving situations. However, the issue is that measurement collection in the real world can be very expensive \cite{Jin2019estimationVehicleDyn}. Consequently, simulation engineers often have to work with sparse datasets that, for example, do not fully capture the variance in the jerk signal behavior across the vehicle fleet. 
{A vehicle's drivetrain encompasses the components that deliver power from the engine/motor to the wheels, while jerk refers to the rate of change of acceleration, a key factor in perceived ride comfort.}
Vehicles with different mileage or drivetrain components with slightly different clearances can exhibit significant differences in their jerk behavior. A sparse or poorly distributed dataset, therefore, limits the generalization ability of a drivetrain model.

To mitigate this issue, synthetic measurements can be generated using non-generative methods that rely on models and their corresponding inputs to simulate new signals of interest \cite{Adegbohun2021PowertrainModeling}. Such a supervised approach to data augmentation can be greatly limited if the required input data for a physics-based or data-driven model are missing in the recorded measurement files or cannot be measured at all \cite{Zhao2015KalmanFilter}. Another major drawback of this conventional approach is that it requires drivetrain engineers to deeply understand the underlying physical phenomena to determine which input data are necessary for a non-generative model to simulate additional valuable signals \cite{Gurusamy2023PredictionEVDriving}. 
Furthermore, the more physically relevant input information {is} fed into the model for precise simulation results, the more accurate the simulation will likely be.
This input-output imbalance necessitates a large number of input parameters (like torque and speed of the electric machine) to simulate a single particular output, such as vehicle acceleration.

These limitations can be tackled with {approaches from} generative artificial intelligence (AI) \cite{Fotiadis2023GenModels,Gilpin2024GenerativeNLDyn}. One {such family of deep generative models includes} architectures such as Variational Autoencoders (VAEs) \cite{Caillon2021RAVE,Pastor2021Autoencoder}.
\rev{Within the broad family of generative models, in this work, we focus on VAEs because they combine stable likelihood-based training, an explicit latent representation, and the possibility of straightforward conditional extensions, which is advantageous for small and noisy engineering datasets when compared with more adversarial training paradigms such as GANs or more data-hungry alternatives such as diffusion models \cite{AutoencodersSurvey,Harshvardhan2020GenModels}.}
VAEs can be trained to generate new signals solely from a limited amount of available signals collected from {physics-based}, simulations or a real system and do not require additional parameterization with respect to the system configuration.
\rev{This self-supervised data-augmentation perspective is especially attractive in drivetrain engineering because it reduces dependence on complete simulator inputs while still allowing physically interpretable post-analysis in the latent space.}
While unconditional data augmentation can be highly advantageous for drivetrain engineers, it is also important to consider the value of conditional AI, for example, conditional VAEs (CVAEs) \cite{Kingma2013VAEs,Zhang2021conditionalVAE,Guo2020CVAE}. In this case, the generation of synthetic data is additionally conditioned on other inputs, similar to supervised data augmentation. 
This can significantly minimize the need for real-world on-road testing, making generative AI a powerful tool for addressing incomplete datasets.

The black-box nature of generative models could lead to poor acceptance among drivetrain experts if the generated signals cannot be explained and validated from an engineering perspective. 
Therefore, a key aspect of using generative AI, in particular VAEs, in the context of drivetrain simulation is the analysis and interpretation of the information encoded in the latent space \cite{Miles2021VAEs,Ahmed2023VAELatentSpace,Criscuolo2025interpretingLatentSpace,EDNMF}. This interpretation step ensures the generation of new signals that are physically plausible. It also helps identify regions in the latent space that match specific simulation requirements, such as a particular driver torque demand.
\rev{In this sense, the main contribution of this paper is to demonstrate that VAEs can serve as effective generative models for drivetrain simulation, bypassing the need for exhaustive manual system parametrization by interpreting the latent space or conditioning the generation of new data on specific desired features, while maintaining physical plausibility.
Our study is unique because we applied VAEs to real-world jerk signals collected by a car manufacturer. To date, to the best of our knowledge, there has been no such study on real jerk signals.}

The outline of this paper is as follows. First, we introduce the drivetrain simulation problem in Section~\ref{sec:drivetrain_sim_problem} and explain the different possible approaches to address it. Second, we present the theoretical background for VAEs in Section~\ref{sec:VAE_theory} and extensions that yield conditional VAEs.
Section~\ref{sec:Numerical_results} presents the dataset and the challenges it imposes for VAE training. We then train unconditional and conditional VAEs to address the problem of jerk signal generation and measure their performance using different metrics. We also compare the performance of the trained VAEs with respect to a physics-based and a hybrid model. The manuscript concludes in Section~\ref{sec:conclusions}.

\section{Drivetrain simulation problem}
\label{sec:drivetrain_sim_problem}

A key performance indicator for quantifying vehicle drivability is the jerk, defined as the rate of change of acceleration. During abrupt positive changes in acceleration, known as tip-in load changes, the torsion in the driveshaft fluctuates, stimulating vibrations in the natural frequency range of the drivetrain (0-2 Hz and 8-12 Hz) \cite{Koch2020drivability}. These longitudinal oscillations negatively affect vehicle dynamics, leading to reduced drivability performance and driver discomfort \cite{Koch2020drivability,Scamarcio2020AntiJerkControllers}. Although this issue is present in vehicles with combustion engines, it becomes more pronounced in electric vehicles due to their higher torque gradients \cite{Scamarcio2020AntiJerkControl_Electric}. The primary control variable for managing the jerk is torque demand, although it is also influenced by drivetrain parameters such as gear ratio, vehicle mass, and tire dynamics. Furthermore, parameters such as electric machine (ELM) speed and road conditions--referred to as driving scenario parameters--significantly impact jerk behavior.

Simulation of vehicle drivability under varying driving scenarios can drastically improve drivetrain development efficiency by reducing the reliance on physical road tests, saving both time and resources \cite{Schmidt2023RevVirtualValidation}. Several models exist to simulate the effects of drivetrain and driving scenarios on jerk behavior \cite{Crowther2005PowertrainModeling,Zeng2018DataDrivenJerk,Pan2021DataDriven,James2020VehicleDynamicsModels}. These models typically predict the jerk as a function of the torque demand. In this context, sudden pedal pressure (tip-in) or release (tip-out) are key driving events. Drivetrain models can generally be categorized into physics-based, data-driven, and hybrid models.\\

\noindent\textbf{Physics-based models} have been the most widely used approach to simulate drivetrain oscillations in various driving scenarios and drivetrain configurations \cite{Crowther2005PowertrainModeling,Lekshmi2019math_mod_ev}. For example, a simplified model of the drive axle of a battery electric vehicle often consists of two masses: the electric machine and the wheel, connected by a finitely rigid driveshaft modeled as a torsion spring \cite{Lekshmi2019math_mod_ev, Franck2021Analysis_ED, Shi2022}. Changes in torque demand, whether during acceleration or braking, alter the torsion in the shaft, exciting vibrations in the natural frequency range of the drivetrain.

These models are typically implemented in environments such as MATLAB/Simulink, as demonstrated, for example, in \cite{Shetty2014,McDonald2012,Butler1999}. They are often integrated within Hardware-in-the-Loop (HIL) setups to simulate real-life driving conditions \cite{Schmidt2023RevVirtualValidation}. They include models of the whole transmission system, covering aspects like torque transmission and gear settings. Physics-based models are straightforward to implement and yield immediate results, making them reliable tools for drivetrain simulations.

However, the accuracy of these models comes at the cost of complexity. Detailed {physics-based} models, which have been iteratively refined by experts over time, require deep domain knowledge and significant experience. Although effective, these traditional models are not always suitable for the fast-paced and disruptive development cycles of modern electric vehicles, where rapid iteration and efficiency are critical \cite{Schmidt2023RevVirtualValidation}.\\

\noindent With the rise of machine learning, particularly in supervised learning, \textbf{data-driven models} have gained prominence in drivetrain simulation \cite{James2020VehicleDynamicsModels,Pan2021DataDriven}. These models are often viewed as ``black box'' approaches, which is advantageous when the underlying physics are too complex to be modeled explicitly \cite{James2020VehicleDynamicsModels}. Typically, data-driven models are trained on annotated datasets where hardware parameters serve as input features and the measured acceleration is used as the target label. The resulting acceleration data is then transformed into jerk signals as required.

Data-driven approaches involve complex calculations and require large neural network architectures with numerous parameters, making them computationally intensive \cite{Zhao2023,Li2022SOHEstimation,Zhou2022State-of-Health}.    
Other limitations are that neural networks are generally less interpretable than physics-based models, which limits their acceptance among simulation engineers. Moreover, these models often struggle with extrapolation \cite{Sun2024DataAugExtrapolation}, as they perform poorly outside the range of training data. To address this, some researchers have explored physics-informed neural networks \cite{Sun2024DataAugExtrapolation,Raissi2019PINNs,Kontolati2024OpLearningDySys,Hu2024StateSpaceModelsOpLearning}, which incorporate physical constraints into the machine learning model to improve performance and reliability.\\

\noindent \textbf{Hybrid models} combine the strengths of both data-driven and physics-based approaches \cite{James2020VehicleDynamicsModels,Wang2022hybrid,Kasilingam2024Hybrid,Shah2024hybrid}. In the context of drivetrain simulation, they typically begin with a simplified {physics-based} model that captures the most critical physical phenomena. This ensures that the essential dynamics are grounded in physics. A data-driven correction term is added to account for any modeling discrepancies or to refine the model \cite{Skull2024}.

Hybrid models not only require a solid understanding of the underlying physics but are also input-intensive \cite{Skull2024}. In addition to torque demand, several other control variables, such as hardware configurations and driving conditions, must be incorporated into the hybrid model. This comprehensive set of inputs is necessary for both the initial physical simulation and the subsequent data-driven correction, making these models highly versatile but computationally demanding.\\

\noindent {In summary}, the advent of battery electric vehicles has created a significant demand for more flexible, accurate, and efficient drivetrain modeling beyond state-of-the-art physics-based approaches. Although supervised data-driven and hybrid models offer good accuracy, they are often data-intensive, which limits their efficiency. To address this challenge, this work explores generative models, more precisely VAEs, as a promising solution to simulate drivetrain behavior, with the aim of balancing accuracy and resource efficiency.

\section{Variational autoencoders as generative models for data augmentation}
\label{sec:VAE_theory}

\begin{figure}[!htb]
    \centering
    \includegraphics[width=0.98\textwidth]{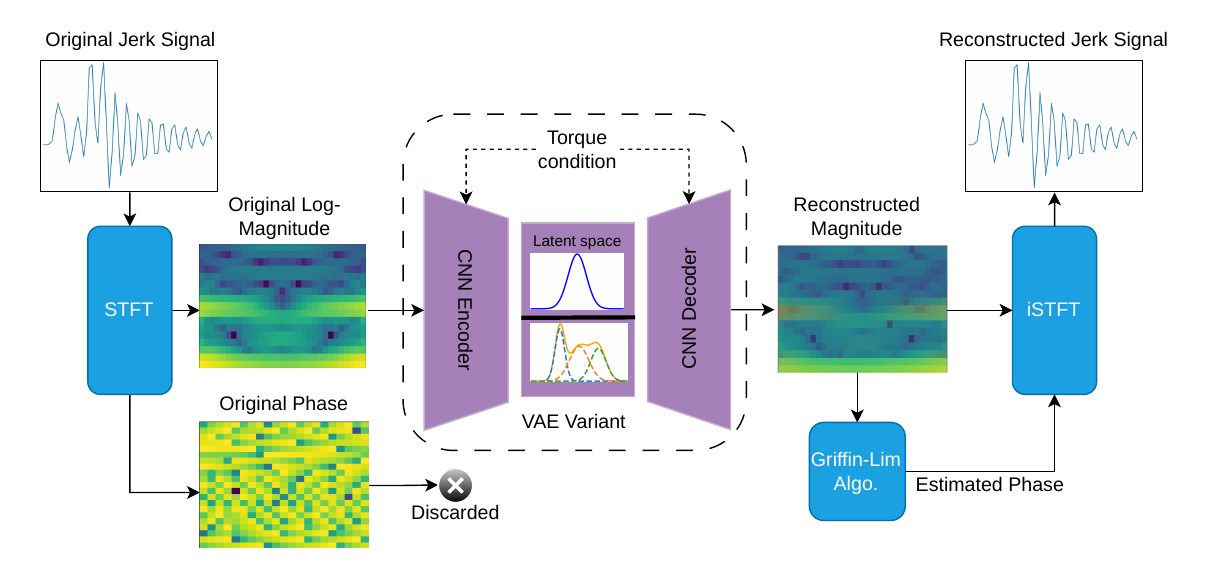}
    \caption{
    \rev{\textbf{Schematic of the drivetrain VAE pipeline:} measured jerk windows are transformed into spectrograms, encoded into a latent representation, optionally conditioned on torque and vehicle information, and decoded back to spectrograms before inverse transformation to the jerk domain.}}
    \label{fig:model_schematic_revision}
\end{figure}

This section provides a theoretical foundation for the use of VAEs as generative models in the context of drivetrain simulations. First, we discuss the transformation of measured jerk signals into spectrograms in Section~\ref{sec:trafo}, which are used as input for the various VAE architectures.
In Section~\ref{subsec:unc_VAEs_theory}, unconditional VAEs are introduced, focusing on their encoder-decoder structure, probabilistic latent space representation, and the objective function that drives their training. The concepts of CVAE and Gaussian mixture model (GMM)-CVAE are presented in Sections~\ref{subsec:CVAEs_theory} and \ref{subsec:GMM-CVAEs_theory}, respectively, \rev{which explain} how conditioning on auxiliary variables and the use of mixture models in the latent space enhance flexibility and control over new \rev{spectrogram} generation. \rev{Figure~\ref{fig:model_schematic_revision} summarizes this pipeline schematically.}

\subsection{Transformation between time domain jerk signal and spectrogram}\label{sec:trafo}

The measurement data that we aim to augment in this study consists of time-domain jerk signals. Prior to their augmentation via the different VAE architectures considered, these signals are transformed into spectrograms, that is, represented in the time-frequency domain using a short-time Fourier transform (STFT) \cite{Griffin1984,Oppenheim1999STFT}. After augmentation, the generated spectrograms are converted back into time-domain jerk signals using the inverse short-time Fourier transform (iSTFT). In this work, we rely on the PyTorch implementation for STFT and iSTFT \cite{Paszke2019pytorch}. 

\rev{Projection of }{time-domain signals into the spectral domain (spectrograms) was chosen as it offers several advantages for processing with VAEs, especially those utilizing convolutional neural network (CNN) encoders and decoders. Spectrograms provide a rich time-frequency representation, allowing the model to learn and capture localized temporal and spectral patterns simultaneously. This can be more effective than using raw time-series data, as CNNs can more readily extract relevant features from the 2D structure of a spectrogram, potentially simplifying the network architecture and improving the ability to model complex signal characteristics.}

\textbf{Time domain jerk signal to spectrogram:} \rev{Let \(s[n]\) denote the discrete jerk signal sampled at uniform time intervals, where \(n \in \{0,\dots,N-1\}\) is the discrete sample index. The discrete STFT of \(s[n]\) is defined as}
\begin{equation}\label{eq:STFT} 
\rev{S[k,m] = \sum_{n=0}^{N_w-1} s[n+mH] \, w[n] \, e^{-j 2 \pi kn/N_w},}
\end{equation}
\rev{where \(m\) is the frame index, \(k\) is the frequency-bin index, \(N_w\) is the window length, \(H\) is the hop size between two consecutive frames, and \(w[n]\) is the analysis window (e.g., a Hann window). Thus, the STFT is fully defined in terms of the discrete signal samples used in the implementation.} From \(S\rev{[k,m]}\), the magnitude and phase can be extracted as follows,
\begin{equation}\label{eq:STFTmagnitude} 
\begin{aligned}
    |\rev{S[k,m]}| &= \sqrt{\operatorname{Re}(\rev{S[k,m]})^2 + \operatorname{Im}(\rev{S[k,m]})^2}, \\
    \varphi\rev{[k,m]} &= \arg(\rev{S[k,m]}) = \tan^{-1}\left(\frac{\operatorname{Im}(\rev{S[k,m]})}{\operatorname{Re}(\rev{S[k,m]})}\right). 
\end{aligned}
\end{equation} 
where $\operatorname{Re}(\rev{S[k,m]})$ and $\operatorname{Im}(\rev{S[k,m]})$ are the real and imaginary parts of $\rev{S[k,m]}$, respectively. 

To compress the dynamic range of the magnitude spectrogram, a logarithmic scaling is applied,
\begin{equation}\label{eq:log_scaled} 
    \rev{x_{km}} = \log(|\rev{S[k,m]}| + \epsilon), 
\end{equation} 
where $\epsilon$ is a small constant which prevents us from taking the logarithm of zero. The log-scaled magnitude spectrogram \(\mathbf{x} \in \mathbb{R}^{F \times T}\), with \rev{\(k\in\{0,\dots,F-1\}\)} the frequency-bin index, \rev{\(m\in\{0,\dots,T-1\}\)} the frame index, and components \(\rev{x_{km}}\), is used as input for the VAE architectures. In the following, for the sake of brevity, we will refer to the log-scaled magnitude of a spectrogram simply as the spectrogram $\vecx$.

{This logarithmic scaling is a common practice in signal processing as it helps to enhance the details of lower-energy components in the spectrogram and makes the distribution of the input features more suitable for neural network training by preventing high-magnitude values from disproportionately influencing the learning process. For jerk signals, oscillations at lower frequencies can be particularly important for drivability perception, and log-scaling can aid in better representing these nuances.}

\textbf{Spectrogram to time domain jerk signal:} In the interest of as-compact-as-possible VAE architectures, herein, we only augment the magnitude of the spectrogram, while the phase is estimated in a post-processing step after augmentation. This approach simplifies the design of the neural networks for the encoder and decoder in the VAEs, as reconstructing both the magnitude and phase simultaneously within the same network can be challenging. Additionally, the phase of jerk signals typically exhibits minimal variation, further justifying this simplification.
Thus, when the VAE is used as a generative model, the phase $\varphi$ is unknown and is estimated using the Griffin-Lim algorithm \cite{Griffin1984}. 

That is, given a generated log-scaled magnitude spectrogram ${\mathbf{x}}$ and estimated phase $\hat{\varphi}$, first, we recover the non-log-scaled magnitude spectrogram as
\begin{equation}
    |\hat{\rev{S}}[k,m]| = \exp{\rev{x_{km}}} - \epsilon.
\end{equation}
Then, the discrete jerk signal \(\hat{s}\rev{[n]}\) is reconstructed from the magnitude and phase using the iSTFT.

\subsection{Variational Autoencoder (VAE)}
\label{subsec:unc_VAEs_theory}

Generative models aim to learn the underlying probability distribution of observed data \(\mathbf{x}\), enabling density estimation and the generation of new samples that resemble the original data or its reconstruction \cite{Harshvardhan2020GenModels}. 
The VAE is a generative model first introduced in \cite{Kingma2013VAEs} that learns a lower-dimensional latent representation \(\mathbf{z}\) to capture the essential features and structure of high-dimensional input data in a compressed way \cite{Doersch2016tutorialVAEs}, which in the current context are spectrograms $\vecx$ as introduced in the above Section~\ref{sec:trafo}. 
\rev{The probabilistic formulation and training objective summarized in this subsection follow the VAE framework introduced in \cite{Kingma2013VAEs} and the tutorial exposition in \cite{Doersch2016tutorialVAEs}. In particular, the latent-variable model, the variational approximation of the posterior, and the ELBO-based objective are specialized to the spectrogram representation of jerk signals introduced in Section~\ref{sec:trafo}.}
\rev{Whenever a Gaussian density is written for \(\mathbf{x}\), the spectrogram is understood in vectorized form, i.e., \(\mathrm{vec}(\mathbf{x}) \in \mathbb{R}^{n_x}\) with \(n_x = FT\). This makes the dimensions of the corresponding covariance matrices precise.}

The hidden random generative process of $\mathbf{x}$ consists of two steps: First, the latent variable $\mathbf{z}$ is sampled from some prior distribution $p(\mathbf{z})$. {This prior $p(\mathbf{z})$ is chosen to be a standard Gaussian distribution, $p(\mathbf{z}) = \mathcal{N}(\mathbf{z} \mid \mathbf{0}, \mathbf{I})$, where $\mathbf{I}$ is the identity matrix. This choice simplifies calculations, particularly for the regularization term in the objective function, and encourages a well-behaved latent space.} Thus, $\mathbf{z} \sim p(\mathbf{z})$. Second, a value $\mathbf{x}$ is generated by some conditional distribution $p_{\vectheta}(\mathbf{x}\mid\mathbf{z})$, i.e. $\mathbf{x} \sim p_{\vectheta}(\mathbf{x}\mid\mathbf{z})$.
We aim to find $p_{\vectheta}(\mathbf{x}\mid\mathbf{z})$ (the parameters $\vectheta$), such that we can sample from $p(\mathbf{z})$ and with high probability, $\mathbf{x} \sim p_{\vectheta}(\mathbf{x}\mid\mathbf{z})$ follows the same distribution as the $\vecx$ in our dataset \cite{Kingma2013VAEs}.
From a mathematical point of view, this can be achieved by maximizing the marginal log-likelihood of \(\mathbf{x}\) (also termed the log evidence). 
\rev{Following the standard VAE derivation in \cite{Kingma2013VAEs,Doersch2016tutorialVAEs}, the model parameters are learned by maximizing a variational lower bound of the marginal log-likelihood of \(\mathbf{x}\). This variational lower bound is constructed from the marginal log-likelihood of \(\mathbf{x}\), also termed the log evidence,}
\begin{equation}\label{eq:deflogpx}
    \log p_{\vectheta}(\mathbf{x}) = \log \int p_{\vectheta}(\mathbf{x}, \mathbf{z}) \, \text{d}\mathbf{z} = \log \int \, p_{\vectheta}(\mathbf{x}\mid\mathbf{z})p(\mathbf{z}) \, \text{d}\mathbf{z}.
\end{equation}
{Here, $p_{\vectheta}(\mathbf{x}\mid\mathbf{z})$ represents the likelihood of observing data $\mathbf{x}$ given latent variables $\mathbf{z}$ and model parameters $\vectheta$, while $p(\mathbf{z})$ is the prior over $\mathbf{z}$. Maximizing $\log p_{\vectheta}(\mathbf{x})$ with respect to $\vectheta$ inherently optimizes the parameters of the likelihood function $p_{\vectheta}(\mathbf{x}\mid\mathbf{z})$. This is because, to render the observed data $\mathbf{x}$ maximally probable under the integral, the likelihood $p_{\vectheta}(\mathbf{x}\mid\mathbf{z})$ must assign high probabilities to $\mathbf{x}$ for latent variables $\mathbf{z}$ that are themselves likely under the prior $p(\mathbf{z})$. Thus, this process compels the model to learn a likelihood function that effectively explains the generation of the observed data.}
This marginal likelihood \(p_{\vectheta}(\mathbf{x})\) is challenging to compute because:
\begin{itemize}
    \item The latent variable \(\mathbf{z}\) may be high-dimensional, making Monte Carlo-based integration over \(\mathbf{z}\) practically infeasible.
    \item {While the likelihood $p_{\vectheta}(\mathbf{x} \mid \mathbf{z})$ (parameterized by the decoder) is typically tractable for a given $\mathbf{z}$, the integration $\int p(\mathbf{z}) \, p_{\vectheta}(\mathbf{x}\mid\mathbf{z}) \, \text{d}\mathbf{z}$ for computing the marginal likelihood $p_{\vectheta}(\mathbf{x})$ is often intractable, especially when $p_{\vectheta}(\mathbf{x} \mid \mathbf{z})$ is a complex function (e.g., a deep neural network) or when no closed-form solution for the integral exists.}
\end{itemize}
Therefore, it was suggested in \cite{Kingma2013VAEs} to compute $\log p_{\vectheta}(\mathbf{x})$ in terms of the expectation $\mathbb{E}_{p_{\vectheta}(\mathbf{z}\mid\mathbf{x})}[\bullet] = \int p_{\vectheta}(\mathbf{z}\mid\mathbf{x}) (\bullet) \, d\mathbf{z}$ w.r.t. the posterior $p_{\vectheta}(\vecz\mid\vecx)$ such that,
\begin{equation}\label{eq:expectationPost}
    \operatorname{log} p_{\vectheta}(\mathbf{x}) = \log \mathbb{E}_{p_{\vectheta}(\mathbf{z}|\mathbf{x})}\bigg[\dfrac{p_{\vectheta}(\vecx, \vecz)}{p_{\vectheta}(\vecz\mid\vecx)}\bigg].
\end{equation}
The issue is that as $p_{\vectheta}(\mathbf{x})$ is intractable, \(p_{\vectheta}(\mathbf{z}\mid\mathbf{x})\) is also intractable.
Therefore, to approximate the intractable posterior \(p_{\vectheta}(\mathbf{z}\mid\mathbf{x})\), a variational distribution \(q_{\vecphi}(\mathbf{z}\mid\mathbf{x})\) is introduced, parameterized by \(\vecphi\) (e.g., through a neural network). The goal will be to make \(q_{\vecphi}(\mathbf{z}\mid\mathbf{x})\) as close as possible to \(p_{\vectheta}(\mathbf{z}\mid\mathbf{x})\), i.e., \(q_{\vecphi}(\mathbf{z}\mid\mathbf{x}) \approx p_{\vectheta}(\mathbf{z}\mid\mathbf{x})\). In VAEs, the posterior \(q_{\vecphi}(\mathbf{z}\mid\mathbf{x})\) and the likelihood \(p_{\vectheta}(\mathbf{x}\mid\mathbf{z})\) are implemented as neural networks: the encoder and decoder, respectively. 

\textbf{Encoder:} 
It approximates the (untractable) posterior \({p}_{\vectheta}(\mathbf{z}\mid\mathbf{x})\) with a variational distribution \(q_{\vecphi}(\mathbf{z}\mid\mathbf{x})\), typically chosen as a Gaussian $\mathcal{N}$ with 
\begin{equation}\label{eq:posterior_vae}
\begin{aligned}
    q_{\vecphi}(\mathbf{z}\mid\mathbf{x}) &= \mathcal{N}\bigl(\mathbf{z} \mid \boldsymbol{\mu}_{\vecphi}(\mathbf{x}), \rev{\boldsymbol{\Sigma}_{\vecphi}(\mathbf{x})}\bigr), \\
    \rev{\boldsymbol{\Sigma}_{\vecphi}(\mathbf{x})} &= \rev{\operatorname{diag}\!\big(\boldsymbol{\sigma}_{\vecphi}^{2}(\mathbf{x})\big).}
\end{aligned}
\end{equation}
Here, \(\boldsymbol{\mu}_{\vecphi}(\mathbf{x}) \in \mathbb{R}^{n_z}\) and \(\boldsymbol{\sigma}_{\vecphi}(\mathbf{x}) \in \mathbb{R}^{n_z}\) are the mean and standard deviation, respectively (with $n_z$ the dimension of the latent space), and are obtained from
\begin{equation}\label{eq:VAE_mu_sigma}
    (\boldsymbol{\mu}_{\vecphi}(\mathbf{x}), \boldsymbol{\sigma}_{\vecphi}(\mathbf{x})) = {\text{NN}}_{\text{enc}}(\mathbf{x}; \vecphi),
\end{equation}
where \(\text{NN}_{\text{enc}}\) is the encoder neural network parametrized in \(\vecphi\), that branches into the two outputs $(\boldsymbol{\mu}_{\vecphi}(\mathbf{x}), \boldsymbol{\sigma}_{\vecphi}(\mathbf{x}))$. The parameter set \(\vecphi\) includes all weights and biases necessary to produce both \(\boldsymbol{\mu}_{\vecphi}(\mathbf{x})\) and \(\boldsymbol{\sigma}_{\vecphi}(\mathbf{x})\). The encoder bridges the data space and the latent space.

\textbf{Decoder:} 
It approximates the likelihood \(p_{\vectheta}({\mathbf{x}}\mid\mathbf{z})\) with a neural network \(\text{NN}_{\mathrm{dec}}(\mathbf{z}; \vectheta)\), parameterized by \(\vectheta\), to output the mean of a Gaussian distribution
\begin{equation}\label{eq:p_VAE}
    p_{\vectheta}({\mathbf{x}}\mid\mathbf{z}) = \mathcal{N}\bigl(\rev{\mathrm{vec}(\mathbf{x})} \mid \rev{\mathrm{vec}\!\big(\text{NN}_{\mathrm{dec}}(\mathbf{z};\vectheta)\big)}, \text{diag}(\boldsymbol{\lambda}^{2})\bigr),
\end{equation}
where \(\rev{\boldsymbol{\lambda} \in \mathbb{R}^{n_x}}\) is a vector that defines the scale of the output covariance. In practice, \(\boldsymbol{\lambda}\) can either be fixed (e.g., \(\boldsymbol{\lambda} = \lambda \mathbf{1}\) for some scalar \(\lambda > 0\)), or learned as part of the model, depending on the modeling assumptions. {In this study, we set $\lambda$ to a constant value of $1.0$ for all output dimensions. This choice makes the negative log-likelihood reconstruction term in the ELBO (Equation~\ref{eq:loss_vae}) equivalent to a scaled Mean Squared Error plus a constant, a common practice that encourages the decoder to learn the mean of the data distribution effectively.}
While more complex output distributions are possible, this choice provides a balance between model expressiveness and computational tractability for many types of continuous data.
Notice that during training, the decoder takes latent variables \(\mathbf{z}\) sampled from the approximate posterior \(q_{\vecphi}(\mathbf{z}\mid\mathbf{x})\) and learns to approximate the likelihood $p_{\vectheta}(\mathbf{x} \mid \mathbf{z})$. 
During generation, it takes \(\mathbf{z}\) sampled from the prior \(p(\mathbf{z})\) and generates new samples ${\mathbf{x}}$. That is, the decoder bridges the latent space and the data space.

\textbf{Objective Function:} 
Coming back to Equation~\eqref{eq:expectationPost} and using the definition of the joint probability in terms of the posterior, that is, $ p(\mathbf{x}, \mathbf{z}) = p(\mathbf{z}\mid\mathbf{x})\, p(\mathbf{x})$, it follows that,
\begin{equation}\label{eq:fulllogpx}
\log p_{\vectheta}(\mathbf{x}) = \mathbb{E}_{q_{\vecphi}(\mathbf{z}\mid\mathbf{x})} \bigg[\log \frac{p_{\vectheta}(\mathbf{x}, \mathbf{z})}{q_{\vecphi}(\mathbf{z}\mid\mathbf{x})}\bigg] + \mathbb{E}_{q_{\vecphi}(\mathbf{z}\mid\mathbf{x})} \bigg[\log \frac{q_{\vecphi}(\mathbf{z}\mid\mathbf{x})}{p_{\vectheta}(\mathbf{z}\mid\mathbf{x})}\bigg].
\end{equation}
The second term is the Kullback-Leibler (KL) divergence between $q_{\vecphi}(\mathbf{z}\mid\mathbf{x})$ and $p_{\vectheta}(\mathbf{z}\mid\mathbf{x})$:
\begin{equation}
D_{\text{KL}} \big( q_{\vecphi}(\mathbf{z}\mid\mathbf{x}) \| p_{\vectheta}(\mathbf{z}|\mathbf{x}) \big) = \mathbb{E}_{q_{\vecphi}(\mathbf{z}\mid\mathbf{x})} \bigg[\log \frac{q_{\vecphi}(\mathbf{z}\mid\mathbf{x})}{p_{\vectheta}(\mathbf{z}\mid\mathbf{x})}\bigg].
\end{equation}
However, since the KL divergence is always greater or equal to zero by definition, it can be neglected and the remaining first term is known as the \textit{Evidence Lower Bound} (ELBO):
\begin{equation}\label{eq:ELBO}
\mathcal{L}_{\vectheta, \vecphi} (\mathbf{x}) = \mathbb{E}_{q_{\vecphi}(\mathbf{z}\mid\mathbf{x})} \big[ \log p_{\vectheta}(\mathbf{x}, \mathbf{z}) - \log q_{\vecphi}(\mathbf{z}\mid\mathbf{x}) \big].
\end{equation}
Therefore, Equation~\eqref{eq:fulllogpx} can be rewritten as,
\begin{equation}\label{eq:logpxreduced}
\begin{aligned}
\log p_{\vectheta}(\mathbf{x}) &\geq \mathbb{E}_{q_{\vecphi}(\mathbf{z}\mid\mathbf{x})} \bigg[ \log \frac{p_{\vectheta}(\mathbf{x}, \mathbf{z})}{q_{\vecphi}(\mathbf{z}\mid\mathbf{x})} \bigg], \\
&\geq \mathcal{L}_{\vectheta, \vecphi} (\mathbf{x}).
\end{aligned}
\end{equation}
Maximizing the marginal log-likelihood is, therefore, equivalent to maximizing the ELBO.
Moreover, to maximize the ELBO, we can equivalently minimize its negative. The negative ELBO is used as the VAE loss function:
\begin{equation}\label{eq:lossELBO}
\lbrace\vecphi^*, \vectheta^*\rbrace = \operatorname{arg} \min_{\vecphi,\vectheta} - \mathcal{L}_{\vectheta, \vecphi} (\mathbf{x}) .
\end{equation}
Substituting the joint distribution \(p_{\vectheta}(\mathbf{x}, \mathbf{z}) = p(\mathbf{z}) \, p_{\vectheta}(\mathbf{x} \mid \mathbf{z})\) into the ELBO, Equation~\eqref{eq:ELBO}, yields,
\begin{equation}
\mathcal{L}_{\vectheta, \vecphi} (\mathbf{x}) = \mathbb{E}_{q_{\vecphi}(\mathbf{z} \mid \mathbf{x})} \big[ \log p_{\vectheta}(\mathbf{x} \mid \mathbf{z}) \big] + \mathbb{E}_{q_{\vecphi}(\mathbf{z} \mid \mathbf{x})} \big[ \log p(\mathbf{z}) - \log q_{\vecphi}(\mathbf{z} \mid \mathbf{x}) \big].
\end{equation}
The second term in the above equation can again be rewritten as a KL divergence. Thus, the ELBO becomes:
\begin{equation}\label{eq:tractableELBO}
\mathcal{L}_{\vectheta, \vecphi} (\mathbf{x}) = \mathbb{E}_{q_{\vecphi}(\mathbf{z} \mid \mathbf{x})} \big[ \log p_{\vectheta}(\mathbf{x} \mid \mathbf{z}) \big] - D_{\text{KL}} \big( q_{\vecphi}(\mathbf{z} \mid \mathbf{x}) \| p(\mathbf{z}) \big),
\end{equation}
where:
\begin{itemize}
    \item The first term, \(\mathbb{E}_{q_{\vecphi}(\mathbf{z}\mid\mathbf{x})} \big[\log p_{\vectheta}(\mathbf{x}\mid\mathbf{z}) \big]\), is the \textbf{reconstruction error term}. It measures how well the decoder \(p_{\vectheta}(\mathbf{x}\mid\mathbf{z})\) can reconstruct the original data \(\mathbf{x}\) from the latent variable \(\mathbf{z}\).
    \item The second term, \(D_{\text{KL}} \big( q_{\vecphi}(\mathbf{z}\mid\mathbf{x}) \| p(\mathbf{z}) \big)\), is the \textbf{regularization error term}, which ensures that the approximate posterior \(q_{\vecphi}(\mathbf{z}\mid\mathbf{x})\) is close to the prior \(p(\mathbf{z})\).
\end{itemize}
Thus, the ELBO given by Equation~\eqref{eq:tractableELBO} is computationally feasible and serves as a surrogate for the initial objective function in Equation~\eqref{eq:fulllogpx}. Expanding the terms, the VAE loss function in Equation~\eqref{eq:lossELBO} becomes:
\begin{equation}\label{eq:loss_vae}
 - \mathcal{L}_{\vectheta, \vecphi} (\mathbf{x}) = \underbrace{\mathbb{E}_{q_{\vecphi}(\mathbf{z}\mid\mathbf{x})} \big[-\log p_{\vectheta}(\mathbf{x}\mid\mathbf{z}) \big]}_{\text{Reconstruction Loss}} + \underbrace{D_{\text{KL}} \big( q_{\vecphi}(\mathbf{z}\mid\mathbf{x}) \| p(\mathbf{z}) \big)}_{\text{Regularization Loss}}.
\end{equation}
So far, we presented the derivation of the objective function assuming $\mathbf{x}$ to be one datapoint (one spectrogram) in our dataset, i.e., $\mathbf{x} = \mathbf{x}^{(i)}$, where our whole dataset is $\mathbf{X} = \lbrace \mathbf{x}^{(i)} \rbrace_{i=1}^{n_d}$ (with $n_d$ the number of spectrograms in the dataset).
Consequently, given the multiple spectrograms $\mathbf{x}^{(i)}$ in the dataset, we need to construct an estimator of the marginal likelihood lower bound of the full dataset as
\begin{equation}\label{eq:totalloss}
 - \mathcal{L}_{\vectheta, \vecphi} (\mathbf{X}) = - \sum_{i=1}^{n_d}  \mathcal{L}_{\vectheta, \vecphi} (\mathbf{x}^{(i)}).
\end{equation}
By minimizing this loss function \eqref{eq:totalloss} for the neural network parameters $\vecphi$ and $\vectheta$, the VAE learns both a probabilistic latent representation $\mathbf{z}$ of the data $\mathbf{x}$ and a generative model capable of reconstructing or sampling new data points $\mathbf{x}$. 
This approach leads to latent spaces that are well-structured, continuous, and, in some cases, interpretable. 

\textbf{Reparameterization Trick:} To enable gradient-based optimization through stochastic sampling, the VAE employs the reparameterization trick. Instead of sampling \(\mathbf{z}\) directly from \(\mathcal{N}(\boldsymbol{\mu}_{\vecphi}(\mathbf{x}), \rev{\boldsymbol{\Sigma}_{\vecphi}(\mathbf{x})})\), we consider
\begin{equation}\label{eq:reparametrization_trick}
    \mathbf{z} = \boldsymbol{\mu}_{\vecphi}(\mathbf{x}) + \boldsymbol{\sigma}_{\vecphi}(\mathbf{x}) \odot \boldsymbol{\epsilon},
\end{equation}
where \(\boldsymbol{\epsilon} \sim \mathcal{N}(\mathbf{0}, \mathbf{I})\) is an auxiliary noise variable, and \(\odot\) denotes element-wise multiplication. This formulation isolates the randomness in \(\boldsymbol{\epsilon}\), allowing gradients to flow through \(\boldsymbol{\mu}_{\vecphi}(\mathbf{x})\) and \(\boldsymbol{\sigma}_{\vecphi}(\mathbf{x})\).

\subsection{Conditional Variational Autoencoder (CVAE)}
\label{subsec:CVAEs_theory}

Conditional VAE (CVAE) extends standard (unconditional) VAE by conditioning the encoder and decoder on additional information, such as class labels or auxiliary data \cite{Kingma2013VAEs,Doersch2016tutorialVAEs,Kingma_2019}. In this case, the observed data \(\mathbf{x}\) is generated from a conditional likelihood distribution \(p_{\vectheta}(\mathbf{x} \mid \mathbf{z}, \mathbf{c})\), parameterized by \(\vectheta\), i.e., \(\mathbf{x} \sim p_{\vectheta}(\mathbf{x} \mid \mathbf{z}, \mathbf{c})\). 
The goal is to learn the parameters \(\vectheta\) of the conditional likelihood \(p_{\vectheta}(\mathbf{x} \mid \mathbf{z}, \mathbf{c})\) such that, by sampling \(\mathbf{z} \sim p(\mathbf{z})\) and conditioning on \(\mathbf{c}\), the generated samples \(\mathbf{x} \sim p_{\vectheta}(\mathbf{x} \mid \mathbf{z}, \mathbf{c})\) resemble the data distribution of \(\mathbf{x}\) in the dataset for the given condition \( \mathbf{c} \in \mathbb{R}^{n_c} \) (with $n_c$ the number of conditions).

From a mathematical perspective, this can be achieved by maximizing the marginal log-likelihood of \(\mathbf{x}\), conditioned on \(\mathbf{c}\),
\begin{equation}\label{eq:deflogpx_conditional}
    \log p_{\vectheta}(\mathbf{x} \mid \mathbf{c}) = \log \int p_{\vectheta}(\mathbf{x}, \mathbf{z} \mid \mathbf{c}) \, \mathrm{d}\mathbf{z} = \log \int p(\mathbf{z}) \, p_{\vectheta}(\mathbf{x} \mid \mathbf{z}, \mathbf{c}) \, \mathrm{d}\mathbf{z}.
\end{equation}
As with the unconditional VAE, computing the marginal likelihood \(p_{\vectheta}(\mathbf{x} \mid \mathbf{c})\) is challenging.
To address this challenge, the posterior distribution \(p_{\vectheta}(\mathbf{z} \mid \mathbf{x}, \mathbf{c})\) is approximated by a variational distribution \(q_{\vecphi}(\mathbf{z} \mid \mathbf{x}, \mathbf{c})\), parameterized by \(\vecphi\) (e.g., through an encoder network). 

The conditional information is integrated into the encoder-decoder as follows:
\begin{enumerate}
    \item The posterior from Equation~(\ref{eq:posterior_vae}) becomes
    \begin{equation}
        q_{\vecphi}(\mathbf{z}\mid\mathbf{x}, \mathbf{c}) = \mathcal{N}(\mathbf{z}\mid \boldsymbol{\mu}_{\vecphi}(\mathbf{x}, \mathbf{c}), \rev{\operatorname{diag}(\boldsymbol{\sigma}_{\vecphi}^2(\mathbf{x}, \mathbf{c}))}),
    \end{equation}
    and the \textbf{encoder} defined in Equation~(\ref{eq:VAE_mu_sigma}) modifies to
    \begin{equation}\label{eq:encoder_cvae}
        {\text{NN}}_{\text{enc}}(\vecx, \vecc; \vecphi)=(\boldsymbol{\mu}_{\vecphi}(\mathbf{x}, \mathbf{c}), \boldsymbol{\sigma}_{\vecphi}(\mathbf{x}, \mathbf{c})).
    \end{equation} 
    
    \item The \textbf{decoder} takes ($\mathbf{z}, \mathbf{c}$) to generate ${\mathbf{x}}$, and the likelihood becomes
    \begin{equation}\label{eq:p_CVAE}
    p_{\vectheta}({\mathbf{x}}\mid\mathbf{z}, \mathbf{c}) = \mathcal{N}(\rev{\mathrm{vec}(\mathbf{x})} \mid \rev{\mathrm{vec}\!\big({\text{NN}}_{\text{dec}}(\mathbf{z}, \mathbf{c}; \vectheta)\big)}, \text{diag}(\boldsymbol{\lambda}^{2})).
    \end{equation}
    \item The \textbf{objective function} from Equation~(\ref{eq:loss_vae}) is modified to account for the condition $\vecc$ as follows,
    \begin{equation}\label{eq:loss_cvae}
            - \mathcal{L}_{\vectheta, \vecphi}^{CVAE} (\mathbf{x}) = \mathbb{E}_{q_{\vecphi}(\mathbf{z}\mid\mathbf{x},\mathbf{c})} \big[-\log p_{\vectheta}(\mathbf{x}\mid\mathbf{z}, \mathbf{c}) \big] + D_{\text{KL}} \big( q_{\vecphi}(\mathbf{z}\mid\mathbf{x}, \mathbf{c}) \| p(\mathbf{z}) \big).
    \end{equation}
\end{enumerate}
In the context of this paper, CVAEs will allow the generation of new driving scenarios under specific conditions $\vecc$, performing the same task as simulator models or traditional simulation methods. This conditioning is achieved by incorporating torque demand as the conditioning variable in CVAEs during training.

\subsection{Gaussian Mixture Model Conditional Variational Autoencoder (GMM-CVAE)}
\label{subsec:GMM-CVAEs_theory}

The GMM-CVAE further extends the standard VAE by modeling the latent space using a {mixture of Gaussian distributions} instead of a single Gaussian \cite{Guo2020CVAE}. In this approach, the latent space \( \mathbf{z} \) is modeled as a mixture of \( K \) Gaussian components, which defines the prior on the latent space, that is,
\begin{equation}\label{eq:p_GMMCVAE_prior}
    p(\mathbf{z}) = \sum_{k=1}^K \pi_k \mathcal{N}\bigl(\mathbf{z} \mid \boldsymbol{\mu}_k, \rev{\operatorname{diag}(\boldsymbol{\sigma}_k^2)}\bigr).
\end{equation}
Here, \( \pi_k \) represents the mixture weights, where $\sum_{k=1}^K \pi_k = 1$, and \(\boldsymbol{\mu}_k\), \rev{\(\operatorname{diag}(\boldsymbol{\sigma}_k^2)\)} are the mean and covariance of the \( k \)-th Gaussian component. The mixture weights \( \pi_k \) represent the prior probability of each Gaussian component. They encode global information about how the data is distributed across the latent space. For instance, if one component corresponds to a cluster of data with higher frequency, \( \pi_k \) determines its relative importance in the mixture.

During training, the encoder network \( {\text{NN}}_{enc}(\vecx, \vecc; \vecphi) \) learns the conditioned mean and covariance
\rev{\((\boldsymbol{\mu}_{\vecphi, k}(\mathbf{x},\mathbf{c}), \operatorname{diag}(\boldsymbol{\sigma}_{\vecphi, k}^{2}(\mathbf{x},\mathbf{c})))\)}
of the variational posterior \( q_{\vecphi}(\mathbf{z} \mid \mathbf{x}, \mathbf{c}) \), which is defined as,
\begin{equation}\label{eq:q_gmm-cvae}
q_{\vecphi}(\mathbf{z}\mid\mathbf{x}, \mathbf{c}) = \sum_{k=1}^K \gamma_k(\mathbf{x},\mathbf{c}) \mathcal{N}\bigl(\mathbf{z}\mid \boldsymbol{\mu}_{\vecphi, k}(\mathbf{x},\mathbf{c}), \rev{\operatorname{diag}(\boldsymbol{\sigma}_{\vecphi, k}^{2}(\mathbf{x},\mathbf{c}))}\bigr).
\end{equation}
The mixture weights \( \gamma_k(\mathbf{x}, \mathbf{c}) \), which represent the responsibility of the \( k \)-th component for the data point \( \mathbf{x} \), are computed as
\begin{equation}
    \gamma_k(\mathbf{x}, \mathbf{c}) = \frac{\pi_k \mathcal{N}\bigl(\mathbf{z}\mid \boldsymbol{\mu}_{\vecphi, k}(\mathbf{x}, \mathbf{c}), \rev{\operatorname{diag}(\boldsymbol{\sigma}_{\vecphi, k}^{2}(\mathbf{x},\mathbf{c}))}\bigr)}{\sum_{j=1}^K \pi_j \mathcal{N}\bigl(\mathbf{z}\mid \boldsymbol{\mu}_{\vecphi, j}(\mathbf{x},\mathbf{c}), \rev{\operatorname{diag}(\boldsymbol{\sigma}_{\vecphi, j}^{2}(\mathbf{x},\mathbf{c}))}\bigr)}. 
\end{equation}
{Here, the notation $\mathcal{N}(\mathbf{z}\mid \boldsymbol{\mu}, \boldsymbol{\Sigma})$ refers to the probability density function of the Gaussian distribution evaluated at a sample $\mathbf{z}$. The responsibilities $\gamma_k(\mathbf{x}, \mathbf{c})$ effectively denote the posterior probability of the latent variable $\mathbf{z}$ (associated with input $\mathbf{x}$ under condition $\mathbf{c}$) belonging to the $k$-th Gaussian component of the mixture. During training, to sample from $q_{\vecphi}(\mathbf{z}\mid\mathbf{x}, \mathbf{c})$, a component $k$ is typically chosen according to these responsibilities (interpreted as $P(k \mid \mathbf{x}, \mathbf{c})$), and then $\mathbf{z}$ is sampled from $\mathcal{N}\bigl(\mathbf{z}\mid \boldsymbol{\mu}_{\vecphi, k}(\mathbf{x},\mathbf{c}), \rev{\operatorname{diag}(\boldsymbol{\sigma}_{\vecphi, k}^{2}(\mathbf{x},\mathbf{c}))}\bigr)$.}

Notice that the responsibilities \( \gamma_k(\mathbf{x}, \mathbf{c}) \) depend on \( \pi_k \), as the posterior distribution is a push-forward of the prior for specific input data.
The term  \( \pi_k \) scales the contribution of each Gaussian component in the posterior. This ensures that the posterior aligns with both the global prior (encoded by  \( \pi_k \)) and the specific data point $\mathbf{x}$.
By combining the prior \( p(\mathbf{z}) \) with the learned variational posterior \( q_{\vecphi}(\mathbf{z}\mid\mathbf{x},\mathbf{c})\), the GMM-CVAE effectively models latent distributions and captures multimodal or clustered data structures.

The objective for the GMM-CVAE remains the same as in Equation~\eqref{eq:loss_cvae}, but $q_{\vecphi}(\mathbf{z}\mid\mathbf{x}, \mathbf{c})$ is now defined as in Equation~\eqref{eq:q_gmm-cvae} to include the mixture of Gaussians in the KL divergence.

\section{Numerical results}
\label{sec:Numerical_results}

In this section, we present the results obtained for the various VAE models introduced in Section \ref{sec:VAE_theory}. 
First, we provide an overview of the dataset used to train the VAE models in Section~\ref{subsec:data_analysis}. Next, we study the performance of the VAE models in Section~\ref{subsec:VAEs_results}, focusing on their ability to accurately reproduce the spectrograms. Sections~\ref{subsec:unc_VAE} and~\ref{subsec:cond_VAE} analyze the unconditional and conditional VAE, respectively.
We analyze the latent space learned by the VAEs and demonstrate the generation of new jerk signals by sampling from this latent space.
The performance of the VAE, CVAE, and GMM-CVAE is quantitatively evaluated using several metrics. These metrics are computed to assess the quality of reconstruction and generative ability of the models.
A comparison between the three models is made based on the computed metrics.
Section~\ref{sec:VAE_vs_PhysicsHybrid} compares the predictive performance of a physics-based drivetrain (hardware) model, a hybrid model, and the CVAE generative model.

\subsection{\rev{Data acquisition and preprocessing}}
\label{subsec:data_analysis}

\textbf{Data acquisition:}
\rev{The study uses experimental powertrain-test-bench measurements from two fully electric SUV variants that mainly differ in maximum torque delivery and in the number of driven axles. The dataset comprises 1,098 rear-axle jerk windows.}

\rev{The tests were performed on a drivetrain bench with independently actuated electric machines, which enabled repeatable tip-in and tip-out torque steps at prescribed operating points. For each operating point, the measured acceleration was differentiated to obtain jerk. Each jerk window contains 60 samples acquired at 50~Hz, corresponding to a 1.2~s time interval. The associated operating conditions span torque values from $-225$ to $820$~Nm and electric machine speeds from 207 to 14,886~rpm.}

\begin{figure}[!htb]
    \center{\includegraphics[width=0.6\textwidth]{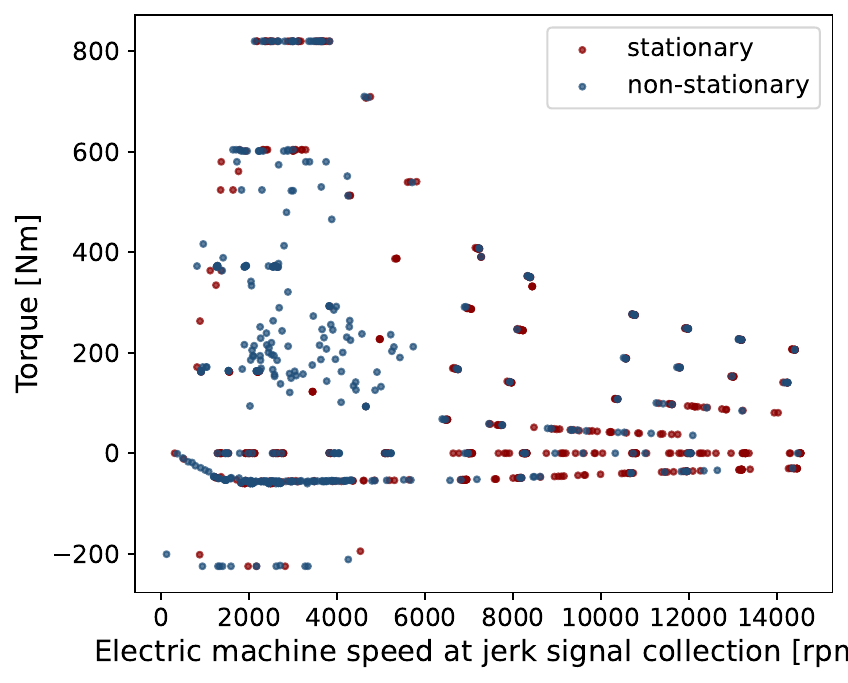}}
    \caption{\rev{\textbf{Distribution of the experimental jerk-signal data:} torque $M$ [Nm] versus electric machine speed $v$ [rpm] for all measured operating points. Only the stationary jerk windows derived from these measurements were used for VAE training.}}
    \label{fig:Dataset_torque_rpm}
\end{figure}

\begin{figure}[!htb]
    \centering
    \includegraphics[width=0.95\textwidth]{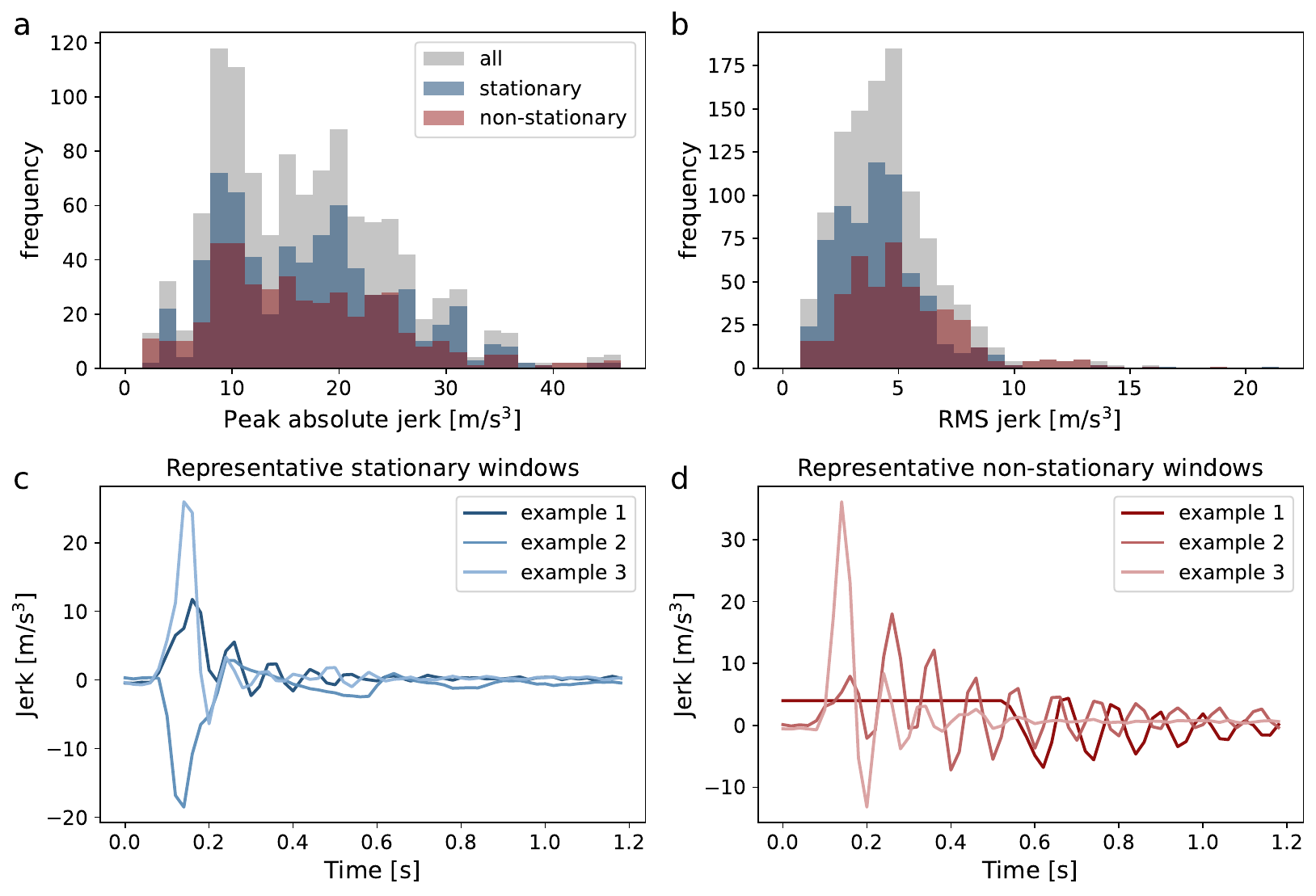}
    \caption{\rev{\textbf{Distribution of experimental jerk windows.} \textbf{a} Peak absolute jerk for all, stationary, and non-stationary windows. \textbf{b} RMS jerk for the same subsets. \textbf{c} Representative stationary jerk windows. \textbf{d} Representative non-stationary jerk windows.}}
    \label{fig:jerk_distribution_overview}
\end{figure}

\rev{Figure~\ref{fig:Dataset_torque_rpm} shows the operating-point coverage in the torque--speed plane, whereas Figure~\ref{fig:jerk_distribution_overview} summarizes the experimental jerk windows in terms of peak amplitude, RMS amplitude, and representative time-domain traces for the stationary and non-stationary subsets. The non-stationary windows exhibit broader RMS distributions and more persistent oscillatory behavior, while the stationary subset remains concentrated in the lower-energy regime used for spectral modeling. The data coverage is markedly non-uniform: mid-range torques dominate the stationary subset, while very high positive torques and strong negative torques are comparatively scarce. This imbalance is important for interpretation because it means the strongest quantitative support is obtained in the densely sampled operating region, while the remaining measurements still provide coverage across the full experimental envelope.} \\

\noindent \textbf{Stationarity:}
\rev{Stationarity of the jerk windows is checked before applying the STFT because the chosen FFT size of 32 samples favors frequency resolution over aggressive temporal localization. A stationary signal is understood here as a signal whose mean, variance, and autocorrelation do not change appreciably over the analysis window. Statistical tests such as Augmented Dickey--Fuller (ADF) \cite{Mushtaq2011} and Kwiatkowski--Phillips--Schmidt--Shin (KPSS) \cite{Kokoszka2016} can be used for this purpose; we use the ADF test as implemented in statsmodels \cite{Seabold2010statsmodels} and retain windows with $p<0.05$.}

\begin{figure}[!htb]
    \centering
    \includegraphics[width=0.6\textwidth]{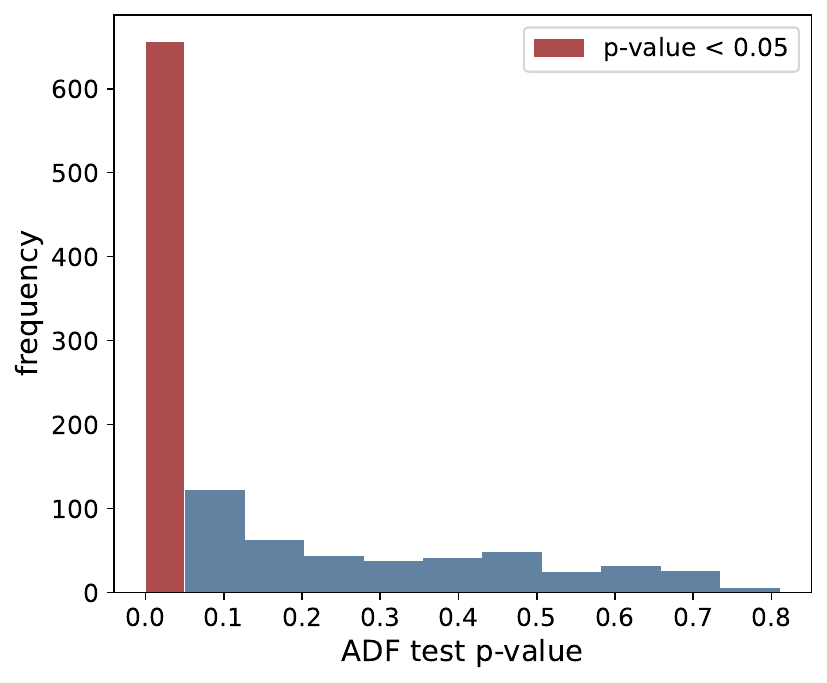}
    \caption{\rev{\textbf{Stability analysis:} ADF stationarity test showing that 655 jerk windows fulfill the criterion $p<0.05$ and are treated as stationary for spectral modeling.}}
    \label{fig:stationary_test}
\end{figure}

\rev{Figure~\ref{fig:stationary_test} shows that 655 of the 1,098 jerk windows satisfy this criterion, while 443 are classified as non-stationary. Only the stationary subset is retained for STFT-based modeling so that all models are trained and evaluated on a signal class that is consistent with the underlying assumptions of the spectral representation.} \\

\noindent \textbf{Signal transformation:} 
\rev{We compute spectrograms for the remaining $n_s = 655$ stationary jerk windows using the STFT described in Section~\ref{sec:trafo}. Table~\ref{tab:STFT_hyperparameters} lists the corresponding hyperparameters. The resulting magnitude spectrograms have size $17 \times 39$ and are normalized using statistics computed on the training split only.} 
\begin{table}[!htb]
\centering
\begin{tabular}{r|c}
  \rev{\textbf{Parameter}} & \rev{\textbf{Value}}  \\
  \hline
 \rev{Hop size} & \rev{2 samples}  \\ 
 \rev{Window} & \rev{Hann window}  \\ 
 \rev{FFT / window size} & \rev{32 samples}\\
 \rev{Signal length} & \rev{60 samples ($1.2$ s)}\\
 \rev{Sampling frequency} & \rev{50 Hz} \\
 \rev{Spectrogram size} & \rev{$17 \times 39$ pixels} \\
\end{tabular}
\caption{\rev{\textbf{STFT hyperparameters:} parameters used to transform the jerk windows into log-magnitude spectrograms.}}
\label{tab:STFT_hyperparameters}
\end{table}

\begin{figure}[!htb]
    \center{\includegraphics[width=0.98\textwidth]{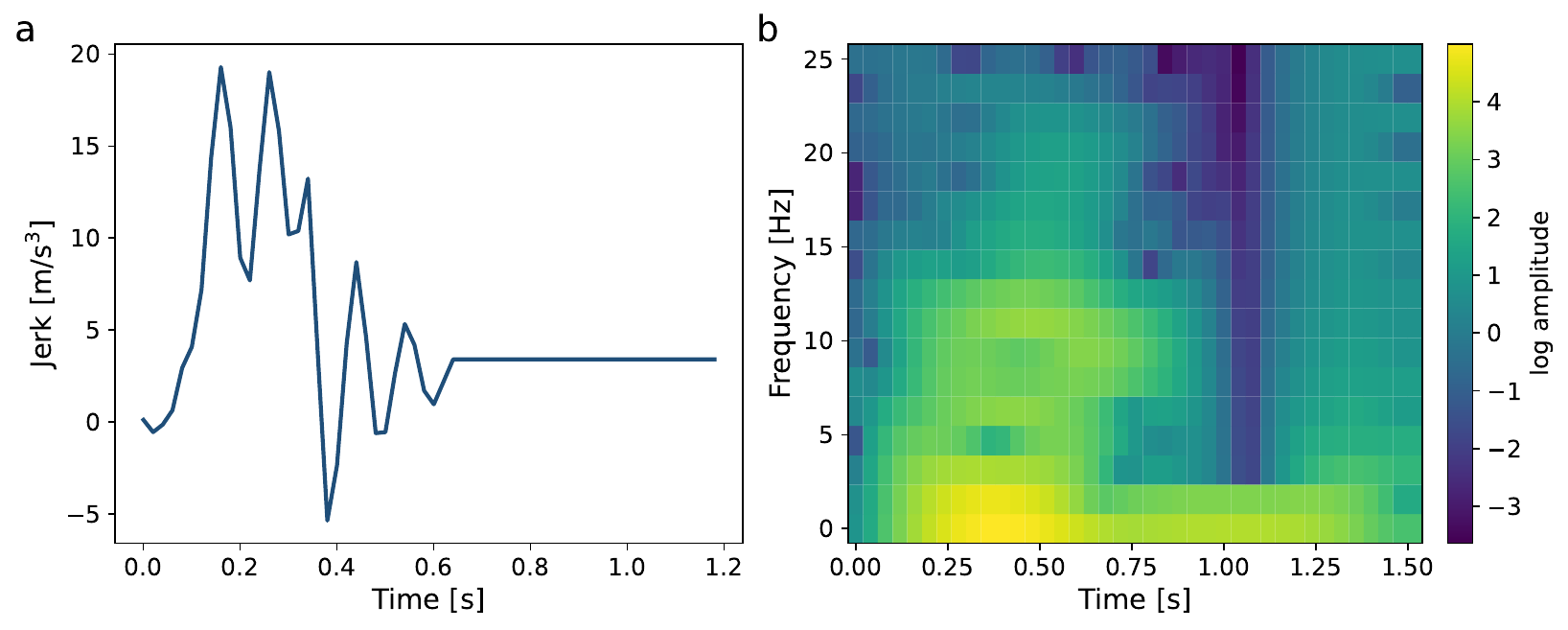}}
    \caption{\rev{\textbf{Signal transformation:} \textbf{a} jerk signal in the time domain (time in seconds) and \textbf{b} corresponding log-magnitude spectrogram (time in seconds, frequency in Hz).}}
    \label{fig:jerk_to_spectrogram}
\end{figure}

\rev{Figure~\ref{fig:jerk_to_spectrogram} illustrates the transformation for one representative window. The logarithmic scaling compresses the magnitude range and makes the spectrogram amplitude unitless, so the colorbar is intentionally shown without physical units. The stationary dataset is then split into 458 training, 66 validation, and 131 test samples, corresponding to a 0.7/0.1/0.2 split. The split is stratified by hardware label so that both vehicle variants remain proportionally represented in all three subsets, and the spectrogram normalization statistics are computed from the training set only before being applied to the validation and test sets.}

\rev{The time axis of the spectrogram differs from the raw-signal axis because the STFT represents overlapping windowed segments. With a hop size of two samples, adjacent frames overlap strongly, which increases time resolution in the spectrogram without indicating any mismatch in the underlying signal length.} \\

\noindent \textbf{Spectrograms characterization:}
\rev{The spectrograms exhibit a broad dynamic range and non-negligible measurement noise, so any generative model must separate persistent frequency structure from acquisition artifacts. The operating-point imbalance remains visible after stationarity filtering: 415 of the 655 stationary windows lie between $-100$ and $100$~Nm, whereas only 69 windows exceed $400$~Nm and only 5 windows are below $-100$~Nm. Consequently, the quantitative trends are supported most strongly in the densely sampled operating region, while the remaining measured torque intervals provide complementary coverage across the full experimental range.}

\subsection{Variational autoencoders as generative models for data augmentation}
\label{subsec:VAEs_results}

\rev{We next report the numerical results for the VAE variants introduced in Section~\ref{sec:VAE_theory}. All models were implemented in Python/PyTorch \cite{Paszke2019pytorch} and trained on the stationary-data split described above.}

\rev{All three models share the same convolutional backbone: four encoder convolution blocks with 32, 64, 96, and 128 channels, followed by batch normalization, ReLU activations, and max pooling; the decoder mirrors this structure through upsampling and convolution until a single spectrogram channel is reconstructed. For the CVAE, torque and hardware information are concatenated with the latent representation through a small conditioning branch, while the GMM-CVAE augments the latent space with a two-component categorical mixture variable. }

\rev{Hyperparameters were selected on the basis of validation experiments and are reported in Table~\ref{tab:VAE_hyperparameters}. In particular, the latent-dimension ablation shown in Table~\ref{tab:latent_ablation} for the unconditional VAE backbone demonstrates that 64 latent variables yield the lowest validation spectrogram MSE (0.4505), followed by 16 (0.4627) and 32 (0.4667). This result suggests that additional latent capacity can improve unconditional VAE reconstruction on the present dataset. Accordingly, the comparative experiments reported below use a 64-dimensional latent space for the VAE, CVAE, and GMM-CVAE.}

\begin{table}[!htb]
\centering
\begin{tabular}{r|c} 
     \rev{\textbf{Hyperparameter}} & \rev{\textbf{Value}}  \\
    \hline
    \rev{Epochs} & \rev{150}  \\ 
    \rev{Learning rate} & \rev{$10^{-4}$}   \\ 
    \rev{Optimizer} & \rev{Adam ($\beta_1=0.9$, $\beta_2=0.999$)}  \\
    \rev{Loss weights} & \rev{\shortstack{unit weights for\\reconstruction and KL terms}}  \\
    \rev{Training batch size} & \rev{152}  \\
    \rev{Validation / test batch size} & \rev{66 / 131}  \\
    \rev{Encoder / decoder conv blocks} & \rev{4 / 4}  \\
    \rev{\shortstack{Latent dimension\\(VAE / CVAE / GMM-CVAE)}} & \rev{64 / 64 / 64}\\
    \rev{GMM components} & \rev{2}\\
    \rev{\shortstack{Trainable parameters\\(VAE / CVAE / GMM-CVAE)}} & \rev{\shortstack{347,297 / 360,513 /\\451,907}}\\
\end{tabular}
\caption{\rev{\textbf{VAE hyperparameters:} training and architecture settings used for the reported experiments.}}
\label{tab:VAE_hyperparameters}
\end{table}


\begin{table}[!htb]
\centering
\begin{tabular}{r|c|c}
\rev{\textbf{Latent dimension}} & \rev{\textbf{Validation MSE}} & \rev{\textbf{Parameters}} \\ \hline
\rev{16} & \rev{0.4627} & \rev{298,001} \\
\rev{32} & \rev{0.4667} & \rev{314,433} \\
\rev{64} & \rev{0.4505} & \rev{347,297} \\
\end{tabular}
\caption{\rev{\textbf{Latent-space ablation:} validation performance of the unconditional VAE for different latent dimensions.}}
\label{tab:latent_ablation}
\end{table}

\subsubsection{Unconditional VAE}
\label{subsec:unc_VAE}

In this subsection, we focus on the unconditional VAE as introduced in Section~\ref{subsec:unc_VAEs_theory}. \\

\noindent \textbf{Performance evaluation:}
\rev{Table~\ref{tab:metrics_comparison} summarizes the quantitative performance on the 131-sample stationary test set. The unconditional VAE achieves the strongest reconstruction accuracy among the VAE variants, with spectrogram MSE $=0.3495$, MAE $=0.4207$, NMSE $=0.3394$, and SSIM $=0.4249$. In the jerk domain, where the original STFT phase is retained for fair reconstruction assessment, the same model yields jerk MSE $=0.0743$, jerk MAE $=0.1436$, and mean correlation $=0.9605$.}

\rev{The unconditional VAE reconstructs the dominant low-frequency drivetrain structure and the main jerk transients reliably. The spectrogram-domain gains are accompanied by a high jerk-domain correlation, which indicates that the model preserves the waveform components that are most relevant for downstream drivability analysis while smoothing part of the measurement noise.}

\begin{figure}[!htb]
    \centering
    \includegraphics[width=0.95\textwidth]{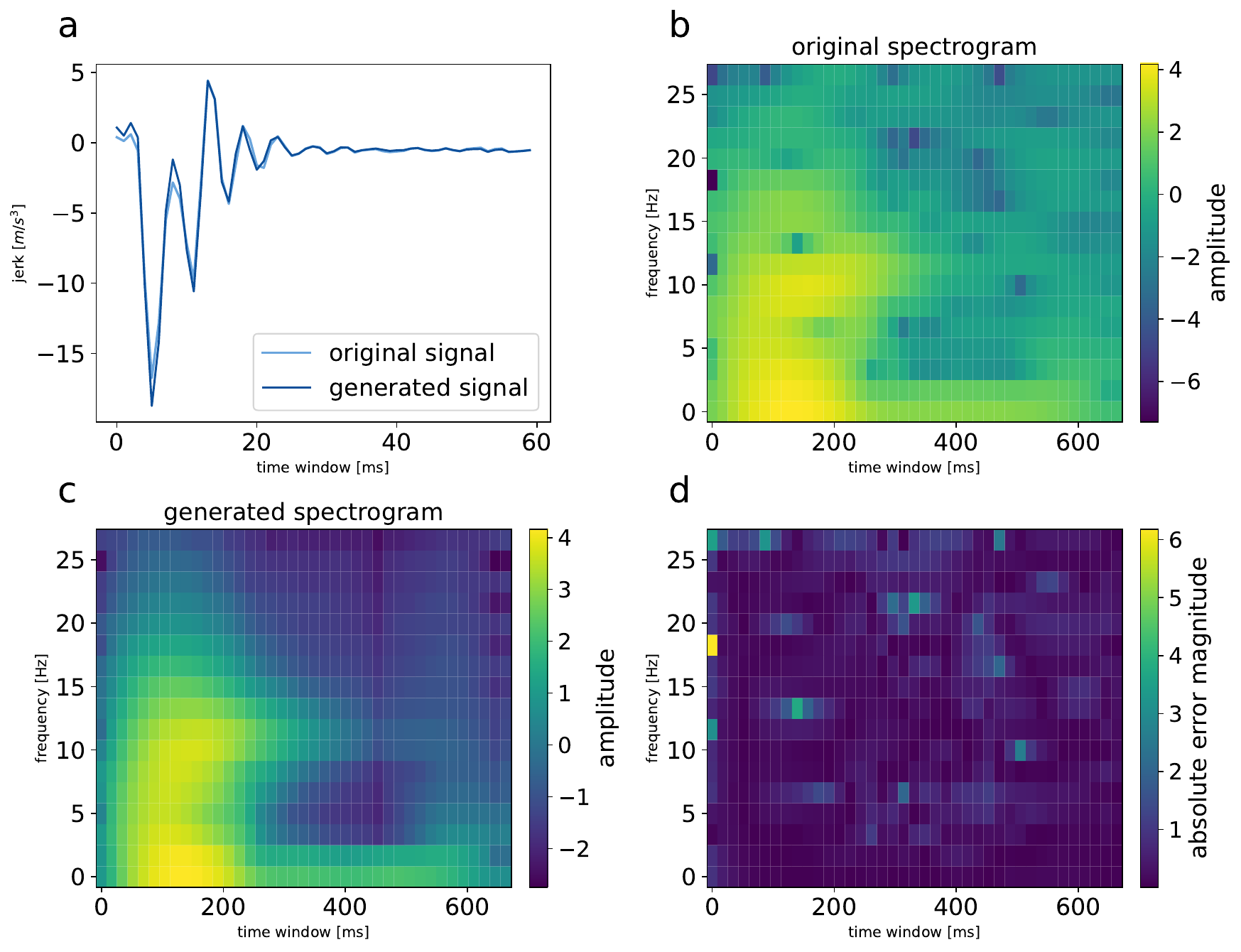}
    \caption{\rev{\textbf{VAE:} model performance evaluation in reconstructing the original spectrogram and jerk signal. \textbf{a} comparison between the original and generated jerk signals in the time domain (time window in ms), \textbf{b} original spectrogram, \textbf{c} generated spectrogram from the VAE, and \textbf{d} absolute error between the original and generated spectrograms (time window in ms, frequency in Hz).}}
    \label{fig:Unc_VAE}
\end{figure}

Figure~\ref{fig:Unc_VAE} illustrates a representative unconditional-VAE reconstruction. The generated jerk waveform in Figure~\ref{fig:Unc_VAE}\textbf{a} closely follows the measured signal, while the reconstructed spectrogram in Figure~\ref{fig:Unc_VAE}\textbf{c} preserves the dominant low-frequency and resonance-band structure visible in the original spectrogram shown in Figure~\ref{fig:Unc_VAE}\textbf{b}. The absolute-error panel in Figure~\ref{fig:Unc_VAE}\textbf{d} shows that the remaining discrepancies are localized mainly in weaker and noisier regions, which is consistent with the smoothing effect induced by the variational bottleneck. \\

\noindent \textbf{Generation of new samples using the reparameterization trick:} 
For a given spectrogram $\vecx$, the latent mean (\(\boldsymbol{\mu}_{\vecphi}(\vecx)\)) and variance (\(\boldsymbol{\sigma}^2_{\vecphi}(\vecx)\)) 
are computed using the encoder network from Equation~(\ref{eq:VAE_mu_sigma}). Making use of the reparameterization trick from Equation~\ref{eq:reparametrization_trick}, we sample the noise vector \(\boldsymbol{\epsilon} \sim \mathcal{N}(\mathbf{0}, \mathbf{I})\) to obtain realizations of the latent variable $\vecz$.
The VAE decodes each sample from the latent distribution into a jerk signal with statistical properties similar to the original.

\begin{figure}[!htb]
\centering
 \includegraphics[width=0.5\textwidth]{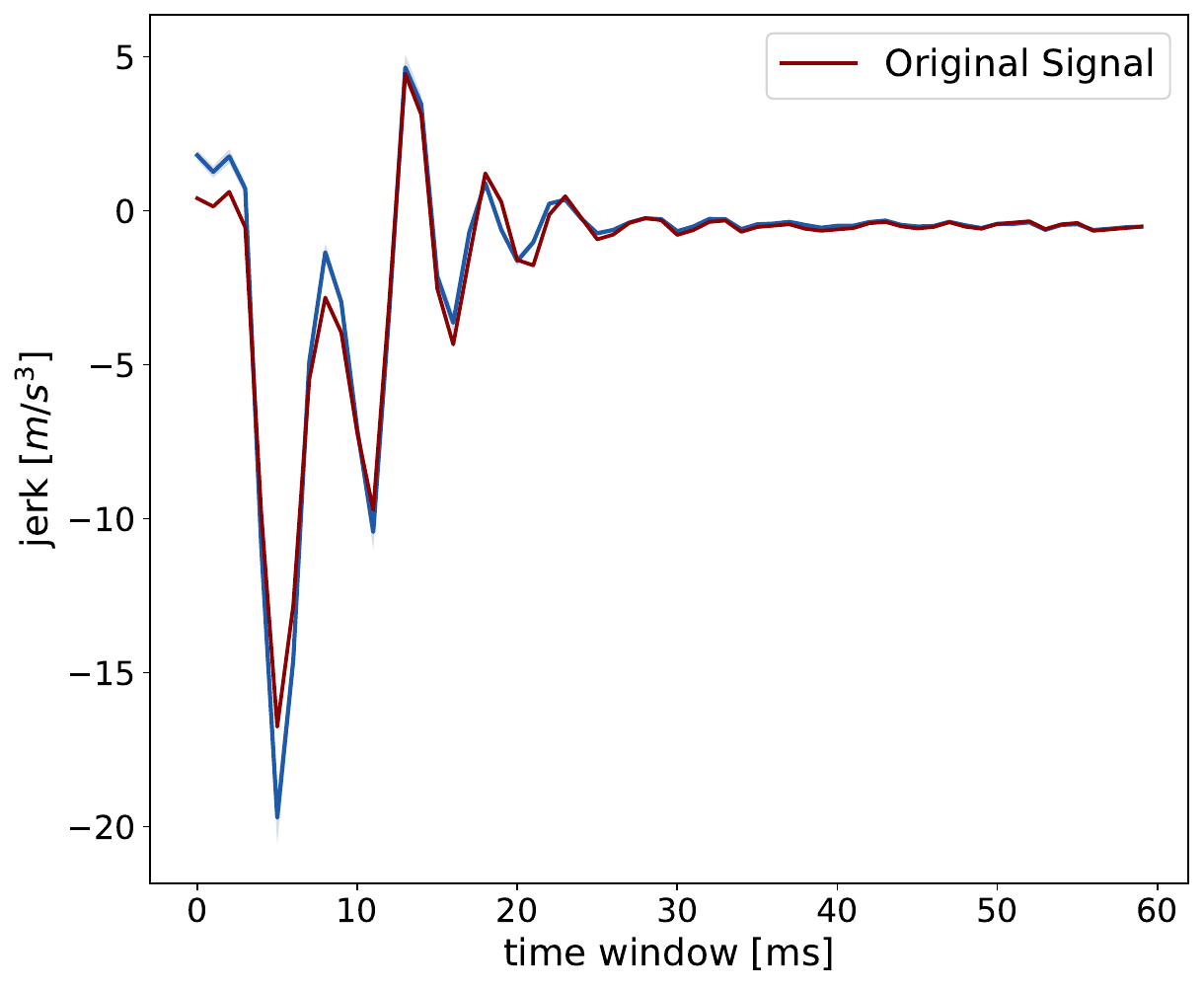}
 \caption{\textbf{Reparameterization trick - VAE generated jerk signal variability:}
 Comparison of the original jerk signal (red) with {the mean (solid blue line) and standard deviation envelope (shaded blue region) of 10,000 jerk signals generated by sampling from the latent posterior $q_{\vecphi}(\mathbf{z}|\mathbf{x}_{\text{original}})$ using the reparameterization trick (Equation~\ref{eq:reparametrization_trick}) and then decoding. The decoder output variance $\lambda$ is fixed (see Section 3.2), so the displayed variability originates from the latent space sampling.}}
 \label{fig:Unc_VAE_sampling_new_jerk_signals}
\end{figure}

Figure~\ref{fig:Unc_VAE_sampling_new_jerk_signals} displays 10,000 realizations of jerk signals. {These signals are generated by first encoding the original jerk signal (red) to obtain its latent distribution $q_{\vecphi}(\mathbf{z}|\mathbf{x}_{\text{original}})$, then repeatedly sampling $\mathbf{z}$ from this distribution using the reparameterization trick (Equation~\ref{eq:reparametrization_trick}), and finally decoding these latent samples. For this local posterior-sampling visualization, the original STFT phase of the reference window is retained so that the displayed spread isolates the effect of latent-space variability.} The close alignment between the original jerk signal and the envelope of these generated signals demonstrates the VAE’s ability to learn a consistent encoding and to represent local variability around a given input. \\

\noindent \textbf{Latent space visualization and qualitative sampling:} VAEs have an inherent limitation: without explicit conditioning, newly generated outputs (spectrograms) cannot be targeted reliably to a desired physical operating point. A {well-structured, that is, well-distributed, latent space still correlates directly with the generative capabilities of the VAE architecture. {Such a latent space ideally exhibits desirable properties like continuity (whereby small changes in the latent vector lead to correspondingly small changes in the generated output), smoothness, and a meaningful organization where different data variations are encoded in a somewhat disentangled or structured manner. These characteristics are crucial for enabling robust interpolation between data points and controlled generation of novel samples.} It influences the variety of features captured when sampling new data points {and the overall quality and coherence of the generated data}.

\rev{To examine what the unconditional VAE has actually learned, we encoded all 655 stationary spectrogram windows and retained the deterministic encoder means $\boldsymbol{\mu}_{\vecphi}(\vecx)\in\mathbb{R}^{64}$. These latent means were standardized and projected to two dimensions with the scikit-learn implementation of t-SNE \cite{Cai2022theoreticaltSNE,Van2008visualizingtSNE,Pedregosa2011ScikitLearn} using perplexity 30, PCA initialization, 2000 iterations, and random seed 42.}

\rev{Figure \ref{fig:tSNE_latent_space} shows the resulting embedding colored by vehicle type (Figure \ref{fig:tSNE_latent_space}\textbf{a}) and by seven torque bins (Figure \ref{fig:tSNE_latent_space}\textbf{b}). To avoid over-interpreting the 2D projection, we also quantified separation directly in the standardized 64-dimensional latent space. The silhouette score is 0.276 for vehicle labels but only 0.003 for torque bins. Hence, the unconditional latent representation separates the two vehicle variants moderately well, whereas torque information remains strongly entangled with other factors. This behavior is consistent with the operating-data imbalance, since the stationary set contains only 5 windows in the $[-300,-100]$~Nm bin but 262 windows in the $[0,100]$~Nm bin, and it suggests that hardware-dependent spectral structure is more dominant than torque alone. Consequently, the t-SNE map is useful for interpretation but does not provide a reliable basis for torque-targeted inverse sampling; that role is fulfilled more robustly by the conditional VAE in Section~\ref{subsec:cond_VAE}.}

\begin{figure}[!htb]
    \centering
    \includegraphics[width=0.95\textwidth]{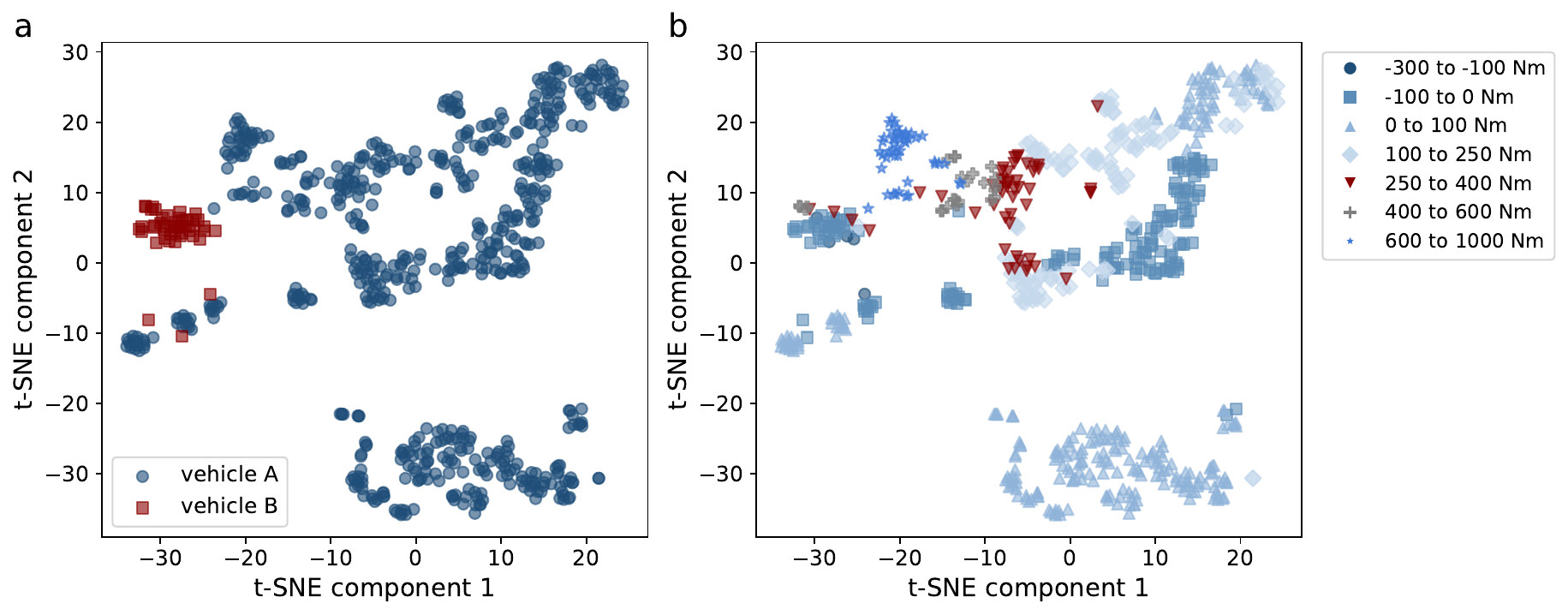}
    \caption{\rev{\textbf{Unconditional VAE latent-space visualization.} Standardized deterministic encoder means for all 655 stationary windows projected to two dimensions with t-SNE (perplexity 30, PCA initialization, 2000 iterations, seed 42). Panel \textbf{a} labels the points by vehicle type, whereas panel \textbf{b} labels the same embedding by seven torque bins. The projection is used for visualization only.}}
    \label{fig:tSNE_latent_space}
\end{figure}

\rev{For the qualitative figure, we selected one representative stationary window from each displayed torque interval, encoded its spectrogram, and drew one latent sample from the corresponding approximate posterior $q_{\vecphi}(\vecz \mid \vecx_{\mathrm{ref}})$. Figure~\ref{fig:Unc_VAE_jerk_signals} therefore shows local variations around known reference operating points. Each row in the figure corresponds to one representative torque interval in $\left[-300, 1000\right]$~Nm and consists of three subplots: \textbf{a} a generated spectrogram ${\mathbf{x}}_k$, \textbf{b} the corresponding jerk signal in the time domain, and \textbf{c} its frequency spectrum.}
 
The spectrograms in Figure~\ref{fig:Unc_VAE_jerk_signals}\textbf{a} display frequency components over time for the generated jerk signals, demonstrating how the VAE can reconstruct plausible frequency representations. 
\rev{Figure~\ref{fig:Unc_VAE_jerk_signals}\textbf{b} shows the corresponding time-domain jerk signals, where the representative operating conditions lead to visibly different waveform shapes.}
Figure~\ref{fig:Unc_VAE_jerk_signals}\textbf{c} shows the frequency spectrum of each jerk signal. The jerk signals in the training and testing datasets used for VAE training consistently exhibit their highest amplitudes around 0-2 Hz and 8-12 Hz. {Notice that these frequencies align with vibrations in the drivetrain's natural frequency range caused by abrupt acceleration changes.} Therefore, we shaded the frequency bands (0–2 Hz and 8–12 Hz) to emphasize that the generated signals are physically plausible.
\rev{For fully generative sampling, where the true phase is unavailable, we reconstruct the time-domain signals using Griffin--Lim \cite{Griffin1984}. A dedicated phase-reconstruction study with 200 iterations yielded a mean waveform RMSE of 4.99, a mean MAE of 2.88, and a mean correlation of $-0.04$ relative to reconstructions using the original phase, indicating that phase recovery is a non-negligible source of time-domain error. For that reason, the reconstruction metrics in Table~\ref{tab:metrics_comparison} use the original STFT phase whenever available, while Griffin--Lim is reserved for genuinely generative examples such as Figure~\ref{fig:Unc_VAE_jerk_signals}.}

\begin{figure}[!hbt]
\centering
 \includegraphics[width=0.78\textwidth]{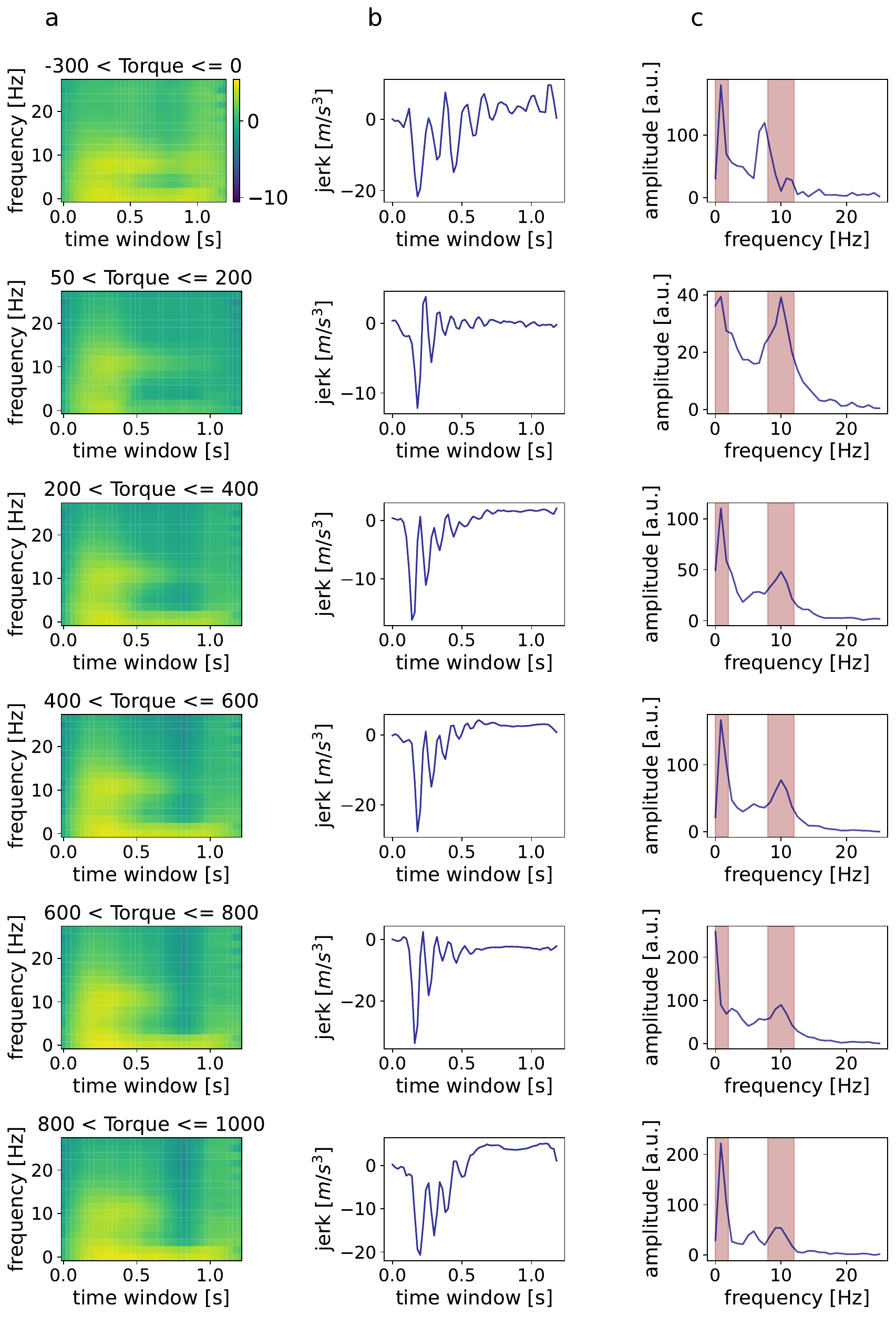}
 \caption{\rev{\textbf{Unconditional VAE: posterior latent samples around representative stationary windows from six torque intervals.} \textbf{a} Generated spectrograms, \textbf{b} generated jerk signals in the time domain, and \textbf{c} the corresponding frequency spectrum. The shaded areas highlight important frequency bands (0-2 Hz and 8-12 Hz). \textbf{Note:} the colorbar displayed in the top subfigure applies to all spectrograms. These examples illustrate local variability around representative operating points and should not be interpreted as strict torque-controlled prior generation.}}
 \label{fig:Unc_VAE_jerk_signals}
\end{figure}

\subsubsection{Conditional VAEs}
\label{subsec:cond_VAE}

This section discusses the results obtained using conditional VAEs, specifically CVAE and GMM-CVAE, using the same network hyperparameters specified in Table~\ref{tab:VAE_hyperparameters}.\\

\noindent \textbf{Performance evaluation:}
\rev{Table~\ref{tab:metrics_comparison} compares the unconditional VAE, CVAE, and GMM-CVAE on the same stationary test set. The unconditional VAE remains the strongest model overall in both the spectrogram and jerk domains, while the CVAE is the best-performing conditional variant with spectrogram MSE $=0.3699$, jerk MSE $=0.1073$, and jerk MAE $=0.1457$. The fact that the conditional models remain slightly weaker in pure reconstruction is not unexpected in this small-data regime. Part of their capacity is dedicated to learning the torque-conditioned manifold, which makes exact spectrogram reconstruction harder even though the conditioning improves controllability.}

\rev{The GMM-CVAE does not improve over the simpler CVAE. In the present implementation, the mixture-assignment logits are learned jointly with the network, but the number of mixture components is fixed at two rather than fully optimized. The resulting parameter increase to 451,907 trainable weights is therefore not justified by a measurable accuracy gain, so we treat the GMM-CVAE as an exploratory extension rather than the preferred final model.}

\begin{table}[!htb]
\centering
\begin{tabular}{r|c|c|c}
\rev{\textbf{Metric}}                              & \rev{\textbf{VAE}} & \rev{\textbf{CVAE}} & \rev{\textbf{GMM-CVAE}} \\ \hline
\rev{Spectrogram MSE}                                  & \rev{0.3495}                     & \rev{0.3699}        & \rev{0.4024}            \\ 
\rev{Spectrogram MAE}                                  & \rev{0.4207}                     & \rev{0.4302}        & \rev{0.4731}            \\ 
\rev{Spectrogram NMSE}                               & \rev{0.3394}                     & \rev{0.3592}        & \rev{0.3908}            \\ 
\rev{Spectrogram NMAE}                               & \rev{0.5398}                     & \rev{0.5519}        & \rev{0.6070}            \\ 
\rev{Spectrogram SSIM}                                 & \rev{0.4249}                     & \rev{0.4074}        & \rev{0.3726}            \\ 
\rev{Spectrogram SNR (dB)}                             & \rev{4.1865}                     & \rev{3.9732}        & \rev{3.5824}            \\ 
\rev{Spectrogram PSNR (dB)}                            & \rev{8.0094}                    & \rev{7.7960}       & \rev{7.4053}           \\ 
\rev{Jerk MSE}                            & \rev{0.0743}                    & \rev{0.1073}       & \rev{0.1495}           \\ 
\rev{Jerk MAE}                            & \rev{0.1436}                    & \rev{0.1457}       & \rev{0.1909}           \\ 
\rev{Jerk correlation}                            & \rev{0.9605}                    & \rev{0.9588}       & \rev{0.9487}           \\ 
\end{tabular}
\caption{\rev{\textbf{VAEs performance evaluation:} comparison of spectrogram-domain and jerk-domain metrics for the unconditional VAE, CVAE, and GMM-CVAE on the stationary test set of 131 samples.}}
\label{tab:metrics_comparison}
\end{table}

Visual inspection of the reconstructed spectrograms and jerk waveforms confirms the same qualitative behavior for the CVAE and GMM-CVAE. The corresponding examples are therefore reported in \ref{appendixD:CVAE_results} in Figures~\ref{fig:CVAE} and \ref{fig:GMM-CVAE}. \\

\noindent \textbf{Generation of new jerk signals by sampling from the latent space:} Conditional generative models allow the decoder to be sampled under an explicit operating condition. 
\rev{Each displayed row is based on one representative stationary window. Its full torque-demand trajectory is used as the conditioning input, and for the CVAE a latent sample $\vecz \sim \mathcal{N}(\ten0,\tenI)$ is drawn from the prior and decoded together with the corresponding vehicle label. The scalar torque shown in the subplot title is the final torque value of the representative conditioning trajectory. Figures~\ref{fig:CVAE_gen} and \ref{fig:GMM-CVAE_gen}, the latter included in {\ref{appendixD:CVAE_results}} to avoid redundancies here, present these qualitative generations for the CVAE and GMM-CVAE, respectively. Figures~\ref{fig:CVAE_gen}\textbf{a} and \ref{fig:GMM-CVAE_gen}\textbf{a} show the generated jerk spectrograms, Figures~\ref{fig:CVAE_gen}\textbf{b} and \ref{fig:GMM-CVAE_gen}\textbf{b} show the corresponding jerk signals, and Figures~\ref{fig:CVAE_gen}\textbf{c} and \ref{fig:GMM-CVAE_gen}\textbf{c} show the associated frequency spectra. The highlighted regions again emphasize the dominant low-frequency and drivetrain-resonance bands.}

\begin{figure}[!htb]
    \centering
    \includegraphics[width=0.78\textwidth]{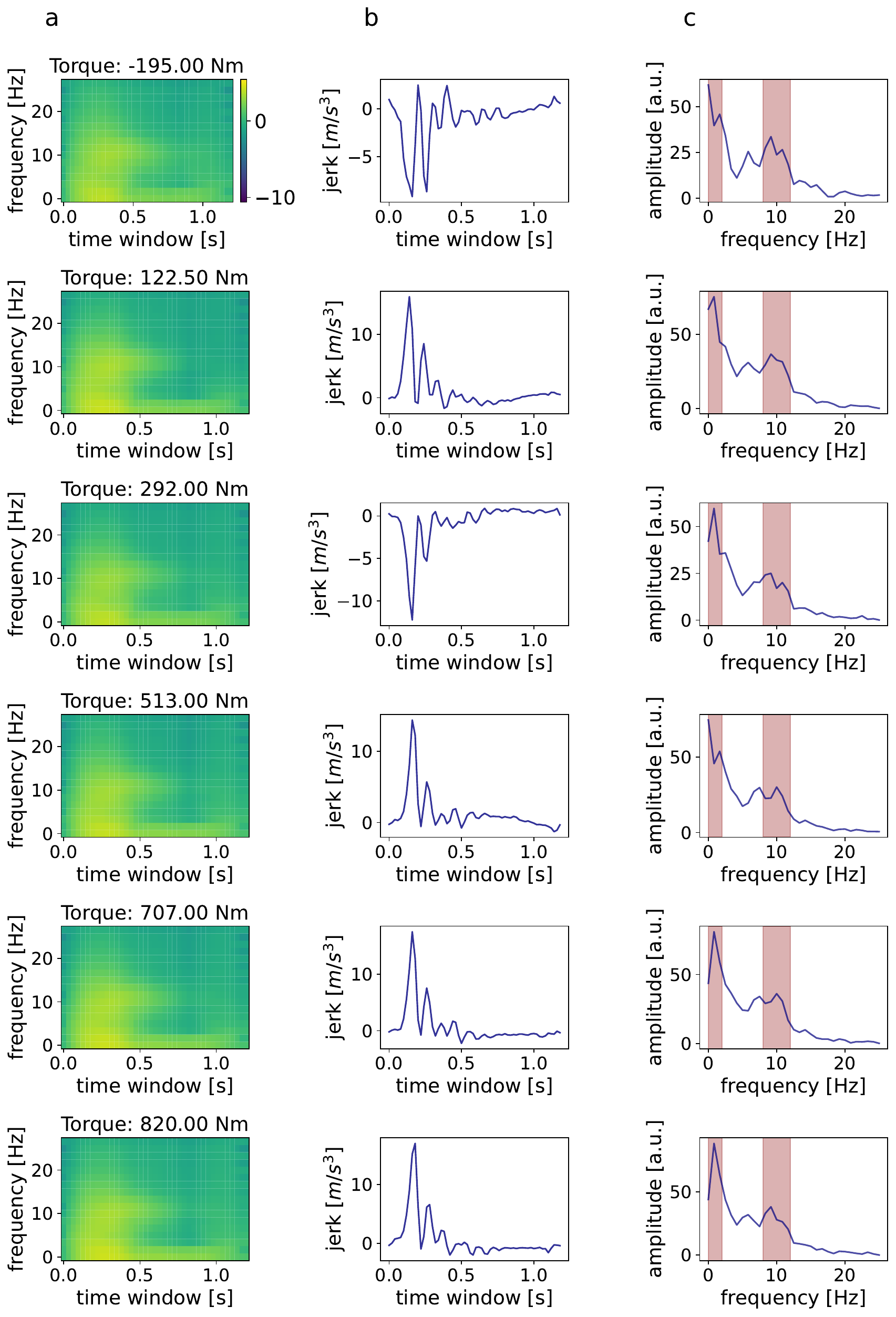}
    \caption{\rev{\textbf{CVAE: conditional prior samples generated under representative torque-demand and hardware conditions.} \textbf{a} Generated spectrograms, \textbf{b} generated jerk signals in the time domain, and \textbf{c} frequency spectrum of each jerk signal. The shaded areas highlight important frequency bands (0-2 Hz and 8-12 Hz). \textbf{Note:} the colorbar displayed in the top subfigure applies to all spectrograms.}}
    \label{fig:CVAE_gen}
\end{figure}

As in the unconditional case, Figures~\ref{fig:CVAE_gen} and \ref{fig:GMM-CVAE_gen} show that the conditional VAEs have effectively learned the underlying distribution of the jerk signals. \rev{Unlike the unconditional VAE figure, which is based on posterior samples around representative reference windows, the CVAE figure is obtained by genuine prior sampling under explicit torque-demand and hardware conditions. This provides a significant advantage when using the VAE as a replacement in drivetrain simulations, where control over specific input parameters is a desired feature.}
The generated signals are also physically plausible in the CVAE and GMM-CVAE cases, as the jerk signals in the dataset used for training correspond to frequency regions with the highest amplitudes around 0-2 Hz and 8-12 Hz.

\rev{When comparing the jerk signals generated by the CVAE (e.g., Figure~\ref{fig:CVAE_gen}) with those from the unconditional VAE for similar torque ranges (e.g., Figure~\ref{fig:Unc_VAE_jerk_signals}), some differences in character may be observed. This can be attributed to the distinct ways the latent spaces are structured and utilized. The unconditional VAE learns a general latent representation, and the samples in Figure~\ref{fig:Unc_VAE_jerk_signals} are local posterior draws around representative reference windows. In contrast, the CVAE's latent space is explicitly conditioned on the torque demand. This conditioning creates a more structured mapping between the desired torque input and the generated output, leading to samples that are more appropriate for targeted generation even if the spectrogram-domain reconstruction metrics are slightly lower in the present small-data regime.}

\subsection{Comparison between conditional generative-, physics-, and hybrid-based models}
\label{sec:VAE_vs_PhysicsHybrid}

In this section, we compare the predictive performance of a {physics-based drivetrain model (referred to as physics-based model)}, a hybrid model, and the CVAE generative model for the dataset presented in Section~\ref{subsec:data_analysis}. The CVAE was chosen here because explicit conditioning is required for the comparison and, among the conditional VAE variants, it provides the best balance between controllability and reconstruction quality; see Table~\ref{tab:metrics_comparison}. In particular, the unconditional VAE has the limitation that the generated jerk signal cannot be conditioned to a specific torque value, while the GMM-CVAE does not provide a significant improvement in performance over the CVAE and introduces additional complexity in sampling from the approximated likelihood.
We refer the reader to {\ref{appendixD:PhysicsHybridModels}} for details of the setup of physics-based and hybrid models.

Table~\ref{tab:VAE_vs_hybrid} summarizes the MSE values for the jerk signals generated by each model. \rev{Unlike Table~\ref{tab:metrics_comparison}, which uses the full stationary test set of 131 samples for VAE-family reconstruction, Table~\ref{tab:VAE_vs_hybrid} is restricted to a matched subset of 32 stationary cases for which outputs from the physics-based model, the hybrid model, and the CVAE were all available.}
As the {physics-based model} is a simplified representation (specifically, the two-mass spring-mass oscillator system mentioned in Section~\ref{sec:drivetrain_sim_problem}), it results in a much lower accuracy compared to the data-driven and hybrid approaches. {This simplified model serves as a basic benchmark and is not intended to be representative of the sophisticated, highly calibrated physics-based simulation tools used in detailed industrial applications. Its performance underscores the difficulty in achieving high fidelity with basic physical formulations alone and motivates the use of data-driven techniques like VAEs or hybrid models to capture more complex system behaviors.} This could be improved by using a more sophisticated {physics-based} model, which was, however, out of scope for this study.
Also note that our goal is to increase the efficiency of the drivetrain development process, and the development of such {physics-based} models is generally time-consuming.

The hybrid model, which utilizes a 7-million-parameter U-Net convolutional neural network architecture as a data-driven correction stage on top of a basic drivetrain physics simulator (see \ref{appendixD:PhysicsHybridModels} for details), demonstrates slightly better accuracy in predicting time-domain jerk signals than the CVAE when the CVAE is used generatively for a given condition (Table~\ref{tab:VAE_vs_hybrid}). 
{We acknowledge that the U-Net in the hybrid model has significantly more parameters (7 million) compared to our VAE architectures. 
This notable difference in model scale (roughly a factor of 15-20) means the comparison is not strictly like-for-like in terms of model complexity or data requirements, although the discrepancy is not as large as three orders of magnitude. 
The primary focus of the VAE development was on generative capabilities and interoperability with more compact models.
The U-Net hybrid model is presented here as a benchmark, representing a high-performance approach for signal prediction in this domain, which helps contextualize the CVAE's generative performance.}
Its substantially higher number of parameters makes it significantly more data-intensive and challenging to train compared to the VAEs tested in this study. 
This complexity underscores the trade-off between accuracy and model efficiency {(referring to aspects like lower data requirements for training, fewer model parameters, and potentially faster inference or training times)}, with VAEs offering a more practical alternative for scenarios with limited data availability {or computational resources}.
{A detailed comparison with hybrid models having a parameter count similar to the VAEs was not performed in this study but could be a direction for future work to assess parameter efficiency more directly for the predictive task.}

\begin{table}[!htb]
\centering
\begin{tabular}{r|c|c|c} 
    \rev{\textbf{Metric}} & \rev{\textbf{Physics-based Model}} & \rev{\textbf{Hybrid Model}} & \rev{\textbf{CVAE}} \\
    \hline
    \rev{Jerk MSE} & \rev{181.2} & \rev{0.4656} & \rev{1.9355}
\end{tabular}
\caption{
\rev{\textbf{Comparison of physics-based model, hybrid model, and CVAE as a generative model.} The MSE was computed as the mean squared error between the measured and generated jerk signals (not spectrograms) over the common matched subset of 32 stationary test signals.}}
\label{tab:VAE_vs_hybrid}
\end{table}

\section{\rev{Conclusion}}
\label{sec:conclusions}

In this study, we showed the potential of {deep generative models, specifically VAEs and their conditional variants,} for drivetrain simulation.
\rev{Using a reproducible preprocessing pipeline, we analyzed 1,098 measured jerk windows, retained 655 stationary samples for spectral modeling, and benchmarked compact unconditional and conditional VAEs on both spectrogram and jerk-domain metrics. The results support the following conclusions:}

\begin{itemize}
    \item \rev{The unconditional VAE is the strongest compact model in this study, achieving spectrogram MSE $=0.3495$, SSIM $=0.4249$, jerk MSE $=0.0743$, and jerk correlation $=0.9605$ on the stationary test set.}

    \item \rev{The CVAE remains the strongest conditional model: although its reconstruction metrics are slightly weaker than those of the unconditional VAE, it outperforms the GMM-CVAE while allowing direct torque-conditioned generation.}

    \item \rev{The GMM-CVAE does not provide a measurable benefit over the simpler CVAE under the present data budget and fixed two-component mixture setup, despite its higher parameter count.}

    \item \rev{The latent-space analysis remains useful for interpretation, but the t-SNE-based sampling procedure should be regarded as a visualization-driven heuristic rather than a physically exact inverse map. Explicit conditioning is therefore the more reliable mechanism for targeted signal generation.}

    \item \rev{Phase recovery is a non-negligible contributor to time-domain error when the original phase is unavailable. Griffin--Lim is therefore suitable for qualitative generative examples, whereas quantitative reconstruction metrics should be computed with the true phase whenever possible.}
\end{itemize}

\rev{Overall, the study confirms that VAEs are a promising compact alternative to more input-intensive physics-based and hybrid approaches for drivetrain jerk modeling. They do not replace detailed simulators in all settings, but they provide a practical generative surrogate when measured data are limited, full physical parameterization is unavailable, and torque-conditioned scenario generation is required.}

\begin{appendix}
\renewcommand{\thefigure}{A\arabic{figure}}
\setcounter{figure}{0}

\appendix
\renewcommand{\thesection}{Appendix \Alph{section}}

\section{Performance metrics to evaluate VAE, CVAE, and GMM-CVAE performance}\label{appendixA}
\label{appendixA:Models_evaluation}

Several metrics are used to evaluate the model performance, as follows.

\textbf{Mean Squared Error (MSE):}
MSE measures the average squared difference between the original and reconstructed data. For a batch of spectrograms $\mathbf{X} = [\mathbf{x}_1, \mathbf{x}_2, \dots, \mathbf{x}_N]$ and their reconstructions $\hat{\mathbf{X}} = [\hat{\mathbf{x}}_1, \hat{\mathbf{x}}_2, \dots, \hat{\mathbf{x}}_N]$, the MSE is given by:
\begin{equation}
    \text{MSE} = \frac{1}{N} \sum_{i=1}^{N} \|\mathbf{X}_i - \hat{\mathbf{X}}_i\|^2
\end{equation}
where $N$ is the size of the batch.

\textbf{Mean Absolute Error (MAE):}
MAE calculates the average absolute difference between the original and reconstructed data. It is defined as:
\begin{equation}
    \text{MAE} = \frac{1}{N} \sum_{i=1}^{N} |\mathbf{X}_i - \hat{\mathbf{X}}_i|.
\end{equation}

\textbf{Normalized MSE (NMSE):}
NMSE normalizes the MSE to account for the variance of the spectrogram data, allowing for better comparison across different datasets. It is defined as:
\begin{equation}
    \text{Normalized MSE} = \frac{\text{MSE}}{\sigma_X^2}
\end{equation}
where $\sigma_X^2$ is the variance of the spectrogram pixel values.

\textbf{Normalized MAE (NMAE):}
Similarly, the normalized MAE accounts for the mean absolute deviation of the spectrogram data:
\begin{equation}
    \text{Normalized MAE} = \frac{\text{MAE}}{\frac{1}{N} \sum_{i=1}^{N} |\mathbf{X}_i - \mu_X|}
\end{equation}
where $\mu_X$ is the mean of the spectrogram pixel values.

\textbf{Structural Similarity Index Measure (SSIM):}
SSIM measures the structural similarity between the original and reconstructed spectrograms, considering luminance, contrast, and structure. For two signals $\mathbf{X}_i$ and $\hat{\mathbf{X}}_i$, SSIM is given by:
\begin{equation}
    \text{SSIM}(\mathbf{X}_i, \hat{\mathbf{X}}_i) = \frac{(2\mu_X \mu_{\hat{X}} + C_1)(2\sigma_{X\hat{X}} + C_2)}{(\mu_X^2 + \mu_{\hat{X}}^2 + C_1)(\sigma_X^2 + \sigma_{\hat{X}}^2 + C_2)}
\end{equation}
where $\mu_X$ and $\mu_{\hat{X}}$ are the means of $\mathbf{X}_i$ and $\hat{\mathbf{X}}_i$, $\sigma_X^2$ and $\sigma_{\hat{X}}^2$ are the variances, $\sigma_{X\hat{X}}$ is the covariance, and $C_1, C_2$ are small constants to stabilize the division.

\textbf{Signal-to-Noise Ratio (SNR):}
SNR is a measure of signal quality, comparing the signal power to the noise power introduced during reconstruction. It is given by:
\begin{equation}
    \text{SNR} = 10 \log_{10} \left( \frac{\text{signal power}}{\text{error power}} \right)
\end{equation}
where:
\begin{equation}
    \text{signal power} = \frac{1}{N} \sum_{i=1}^{N} \mathbf{X}_i^2, \quad \text{error power} = \frac{1}{N} \sum_{i=1}^{N} (\mathbf{X}_i - \hat{\mathbf{X}}_i)^2
\end{equation}

\textbf{Peak Signal-to-Noise Ratio (PSNR):}
PSNR is used to measure the ratio between the maximum possible signal power and the noise power. It is calculated as:
\begin{equation}
    \text{PSNR} = 10 \log_{10} \left( \frac{ \text{MAX}^2}{\text{MSE}} \right)
\end{equation}
where $\text{MAX}$ is the maximum pixel value of the original spectrogram, and MSE is the mean squared error between the original and reconstructed spectrograms.

These metrics provide a comprehensive evaluation of the performance of the VAE in terms of reconstruction quality, structural similarity, and noise handling.

\section{Complementary results for conditional VAEs}
\label{appendixD:CVAE_results}

\rev{Figures~\ref{fig:CVAE} and \ref{fig:GMM-CVAE} illustrate the performance of conditional VAEs. Figures~\ref{fig:CVAE}\textbf{a} and \ref{fig:GMM-CVAE}\textbf{a} compare the original and generated jerk signals in the time domain, showing strong alignment. In Figures~\ref{fig:CVAE}\textbf{b} and \ref{fig:GMM-CVAE}\textbf{b}, the original spectrogram is shown, while Figures~\ref{fig:CVAE}\textbf{c} and \ref{fig:GMM-CVAE}\textbf{c} show the reconstructed spectrograms. Figures~\ref{fig:CVAE}\textbf{d} and \ref{fig:GMM-CVAE}\textbf{d} depict the corresponding absolute spectrogram errors.}

\begin{figure}[!htb]
    \centering
    \includegraphics[width=0.95\textwidth]{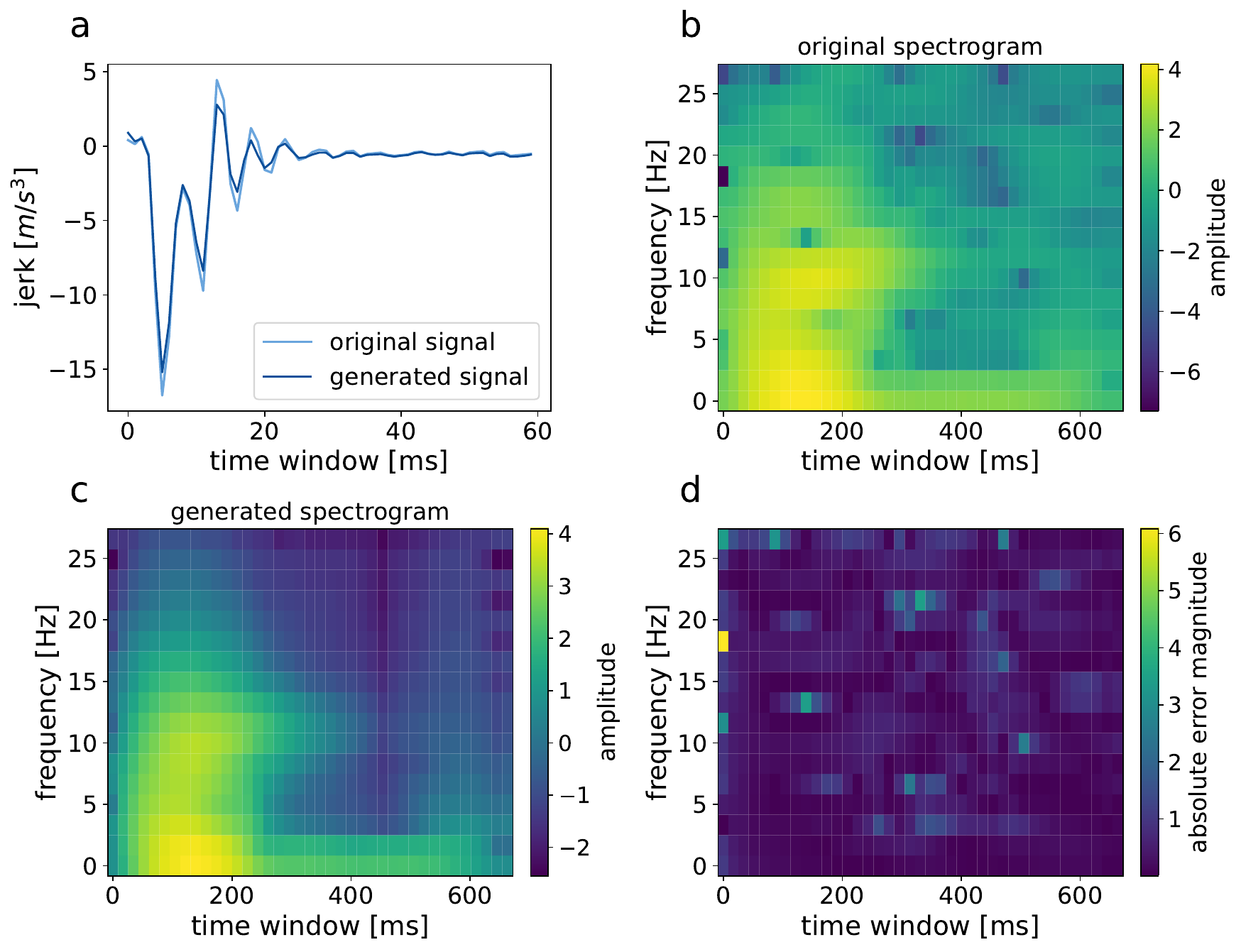}
    \caption{\rev{\textbf{CVAE:} model performance evaluation in reconstructing the original spectrogram and jerk signal. \textbf{a} comparison between the original and generated jerk signals in the time domain (time window in ms), \textbf{b} original spectrogram, \textbf{c} generated spectrogram from the CVAE, and \textbf{d} absolute error between the original and generated spectrograms (time window in ms, frequency in Hz).}}
    \label{fig:CVAE}
\end{figure}

\begin{figure}[!htb]
    \centering
    \includegraphics[width=0.95\textwidth]{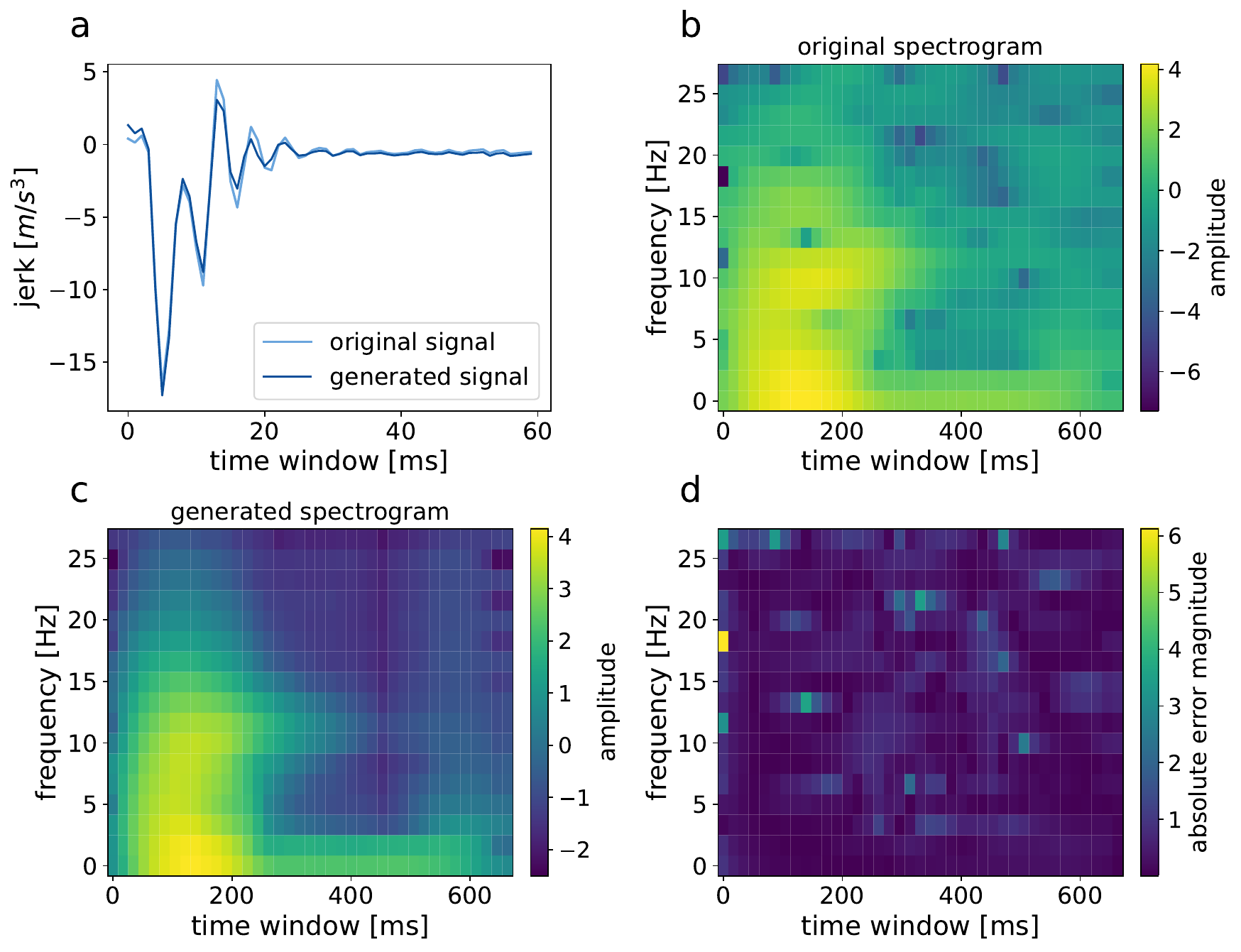}
    \caption{\rev{\textbf{GMM-CVAE:} model performance evaluation in reconstructing the original spectrogram and jerk signal. \textbf{a} comparison between the original and generated jerk signals in the time domain (time window in ms), \textbf{b} original spectrogram, \textbf{c} generated spectrogram from the GMM-CVAE, and \textbf{d} absolute error between the original and generated spectrograms (time window in ms, frequency in Hz).}}
    \label{fig:GMM-CVAE}
\end{figure}

\rev{Figure~\ref{fig:GMM-CVAE_gen}\textbf{a} shows qualitative GMM-CVAE generations for the same representative torque-demand conditions used above. For each row, the most probable mixture component for the representative condition is selected and a latent sample is drawn from the corresponding Gaussian prior before decoding. Figure~\ref{fig:GMM-CVAE_gen}\textbf{b} shows the corresponding jerk signal, while Figure~\ref{fig:GMM-CVAE_gen}\textbf{c} shows the associated frequency spectrum. The highlighted regions again emphasize the dominant low-frequency and drivetrain-resonance bands.}

\begin{figure}[!htb]
    \centering
    \includegraphics[width=0.78\textwidth]{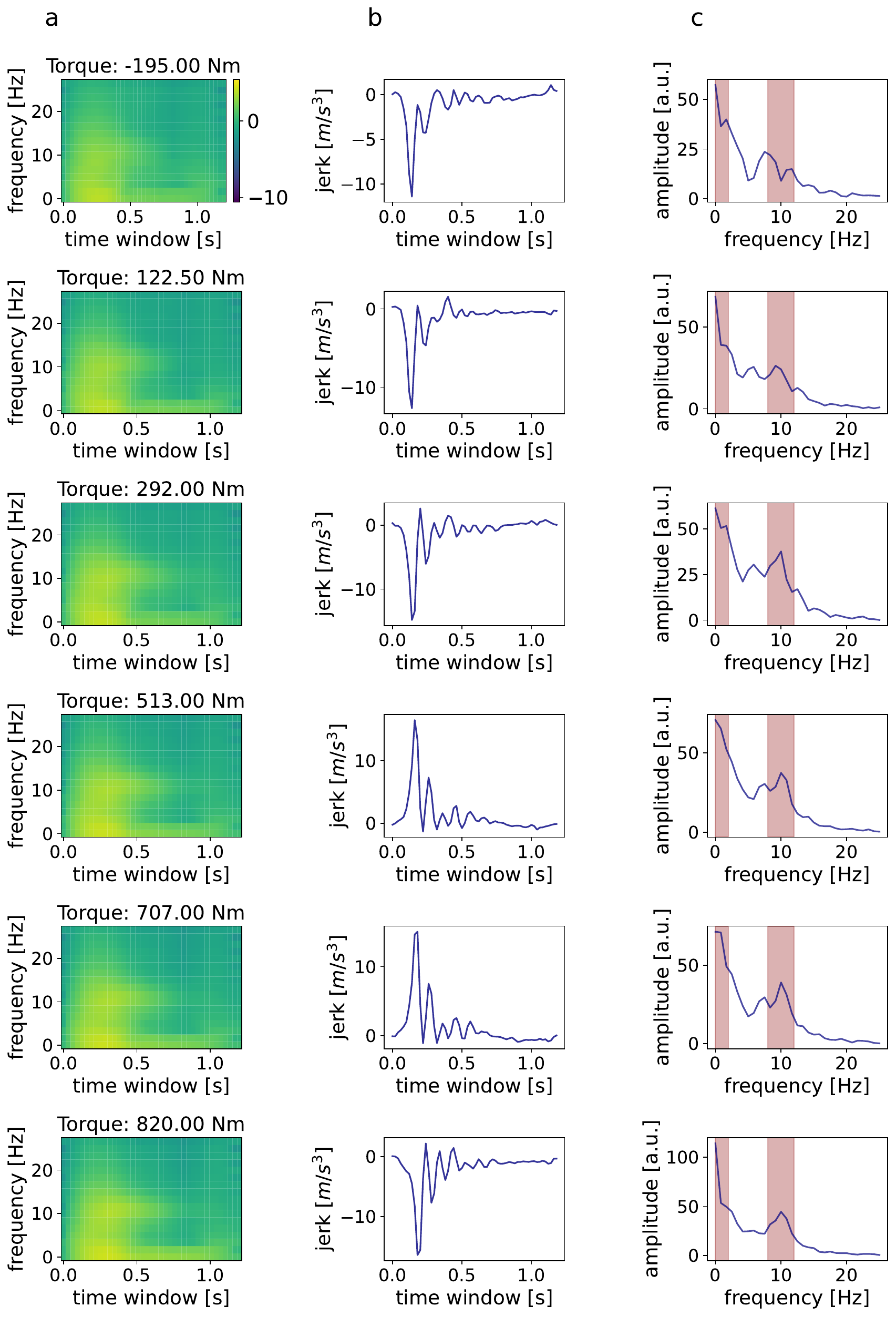}
    \caption{\rev{\textbf{GMM-CVAE: qualitative prior samples generated under representative torque-demand conditions.} \textbf{a} Generated spectrograms, \textbf{b} generated jerk signals in the time domain, and \textbf{c} frequency spectrum of each jerk signal. The shaded areas highlight important frequency bands (0-2 Hz and 8-12 Hz). \textbf{Note:} the colorbar displayed in the top subfigure applies to all spectrograms.}}
    \label{fig:GMM-CVAE_gen}
\end{figure}

\section{Physics-based and hybrid models set up}
\label{appendixD:PhysicsHybridModels}

To remove the unwanted effect of jerk, one can calibrate the engine maps {by building} a physics-based hardware simulator that simulates the jerk and acceleration given a torque signal as input. Theoretically, the simulator could create signals close to the real world, but complex physical interactions affect the accuracy of the simulations and cause them to diverge from real measurements. {As a consequence,} one can build a hybrid hardware simulator by {adding} a correction model \cite{WINKLER2021101072, HARLIM2021109922, Kotter2018X_in_theLoop}. The correction model is used to bring the simulated signal close to reality {by learning complex physical interactions from the data}. 
To this end, we train a neural network on pairs of simulated and measured acceleration signals.

In this work, we use a Matlab Simulink model for our physics-based hardware simulator. The Simulink model implements the basic physics formula to simulate the acceleration signal and takes into consideration hardware parameters such as vehicle mass (kg), gear ratio, rotational inertia ($\text{kg} \cdot \text{m}^2$), stiffness (Nm/degrees), and wheel radius (meters).

Our hybrid simulator approach relies on a correction model on top of a physics-based hardware simulator, and we compare it with the CVAE model. The correction model consists of a neural network (NN) that corrects the simulator output, bridging the gap between simulated and real data.

The input to the NN is composed of 2 channels: the simulated signal (acceleration) and torque demand. There is only one output: the corrected signal that corresponds to the real acceleration. The NN model was trained using data collected from a test bench and real-world measurements.

U-Net is a fully convolutional neural network architecture that evolved from the traditional convolutional neural network and was first designed and applied in 2015 to process biomedical images \cite{ronneberger2015unetconv}.
It is an encoder-decoder architecture with skip connections (residual connections) between the encoder and decoder layers. Although originally developed for image data, the U-Net architecture can be applied to signal processing \cite{stoller2018waveunetnet}. We chose U-Net as the architecture for our neural network model \cite{ronneberger2015unetconv}. The U-Net original design was modified using 1D convolutions with architectural changes inspired by \cite{liu2022convnet2020s}. The main takeaway is that we use the ConvNeXt block \cite{liu2022convnet2020s} with inverted bottleneck \cite{Sandler2018} as the base convolution block. We use a U-Net with 7 million parameters, a convolution base number of filters equal to 96, and a kernel size of 7.

\end{appendix}


\paragraph{Acknowledgments}
The authors acknowledge discussions with David Anton on Bayesian statistics and generative modeling.

\paragraph{Funding Statement}
This research has been financed by the core funding of Porsche and TU Braunschweig.

\paragraph{Competing Interests}
None

\paragraph{Data Availability Statement}
\rev{Data will be made available on request. For access to the experimental drivetrain measurements supporting this study, readers may contact the corresponding author.}

\paragraph{Code Availability Statement}
\rev{The pre-processing and analysis scripts will be made available by the authors upon reasonable request.}

\paragraph{Ethical Standards}
The research meets all ethical guidelines, including adherence to the legal requirements of the study country.

\paragraph{Author Contributions}
\textbf{Conceptualization:} P.S., H.W., M.S., B.B., AD.C. and L.V.;
\textbf{Experimental data acquisition:} L.V., M.S. and AD.C.
\textbf{Software:} P.S., JH.UQ., B.B., AD.C. and L.V.; 
\textbf{Data visualization:} P.S., JH.UQ., B.B. and L.V.; 
\textbf{Formal analysis:} P.S., JH.UQ., B.B., AD.C., H.W. and M.S.; 
\textbf{Writing original draft:} P.S., JH.UQ., AD.C., H.W. and M.S.; 
\textbf{Funding acquisition:} H.W., M.S.; JH.UQ. and B.B. contributed equally as second authors to this work;
All authors approved the final submitted draft.


\begin{thebibliography}{00}


\bibitem{Jin2019estimationVehicleDyn}
X. Jin, G. Yin, N. Chen,
\textit{Advanced estimation techniques for vehicle system dynamic state: A survey},
Sensors,
19 (19) (2019) 4289.

\bibitem{Adegbohun2021PowertrainModeling}
F. Adegbohun, A. Von Jouanne, B. Phillips, E. Agamloh, A. Yokochi,
\textit{High performance electric vehicle powertrain modeling, simulation and validation},
Energies,
14 (5) (2021) 1493.

\bibitem{Zhao2015KalmanFilter}
Z. Zhao, H. Chen, J. Yang, X. Wu, Z. Yu,
\textit{Estimation of the vehicle speed in the driving mode for a hybrid electric car based on an unscented {Kalman} filter},
Proceedings of the Institution of Mechanical Engineers, Part D: Journal of automobile engineering,
229 (4) (2015) 437--456.

\bibitem{Gurusamy2023PredictionEVDriving}
A. Gurusamy, B. Ashok, B. Mason,
\textit{Prediction of electric vehicle driving range and performance characteristics: A review on analytical modeling strategies with its influential factors and improvisation techniques},
IEEE Access,
11 (2023) 131521--131548.

\bibitem{Fotiadis2023GenModels}
S. Fotiadis, M. L. Valencia, S. Hu, S. Garasto, C. D. Cantwell, A. A. Bharath,
\textit{Disentangled generative models for robust prediction of system dynamics},
in: International Conference on Machine Learning,
PMLR, 2023, pp. 10222--10248.

\bibitem{Gilpin2024GenerativeNLDyn}
W. Gilpin,
\textit{Generative learning for nonlinear dynamics},
Nature Reviews Physics,
6 (3) (2024) 194--206.

\bibitem{Caillon2021RAVE}
A. Caillon, P. Esling,
\textit{RAVE: A variational autoencoder for fast and high-quality neural audio synthesis},
arXiv preprint arXiv:2111.05011,
2021.

\bibitem{Pastor2021Autoencoder}
O. Pastor-Serrano, D. Lathouwers, Z. Perk{\'o},
\textit{A semi-supervised autoencoder framework for joint generation and classification of breathing},
Computer Methods and Programs in Biomedicine,
209 (2021) 106312.

\bibitem{AutoencodersSurvey}
M. A. Talha, A. Z. Md. Khairi, L. M. Kamarudin, A. Zakaria, I. Yunas,
\textit{Autoencoders and their applications in machine learning: a survey},
Artificial Intelligence Review,
57 (1) (2024) 28.

\bibitem{Kingma2013VAEs}
D. P. Kingma, M. Welling,
\textit{Auto-encoding variational bayes},
arXiv preprint arXiv:1312.6114,
2013.

\bibitem{Zhang2021conditionalVAE}
C. Zhang, R. Barbano, B. Jin,
\textit{Conditional variational autoencoder for learned image reconstruction},
Computation,
9 (11) (2021) 114.

\bibitem{Guo2020CVAE}
C. Guo, J. Zhou, H. Chen, N. Ying, J. Zhang, D. Zhou,
\textit{Variational autoencoder with optimizing {Gaussian} mixture model priors},
IEEE Access,
8 (2020) 43992--44005.

\bibitem{Miles2021VAEs}
C. Miles, M. R. Carbone, E. J. Sturm, D. Lu, A. Weichselbaum, K. Barros, R. M. Konik,
\textit{Machine learning of {Kondo} physics using variational autoencoders and symbolic regression},
Physical Review B,
104 (23) (2021) 235111.

\bibitem{Ahmed2023VAELatentSpace}
T. Ahmed, L. Longo,
\textit{Latent Space Interpretation and Visualisation for Understanding the Decisions of Convolutional Variational Autoencoders Trained with {EEG} Topographic Maps.},
in: xAI (Late-breaking Work, Demos, Doctoral Consortium),
2023, pp. 65--70.

\bibitem{Criscuolo2025interpretingLatentSpace}
S. Criscuolo, A. Apicella, R. Prevete, L. Longo,
\textit{Interpreting the latent space of a Convolutional Variational Autoencoder for semi-automated eye blink artefact detection in {EEG} signals},
Computer Standards \& Interfaces,
92 (2025) 103897.

\bibitem{EDNMF}
T. Boukhalfa, M. Ghogho, A. Swami,
\textit{Encoder-decoder nonnegative matrix factorization with $\beta$-divergence for data clustering},
Pattern Recognition,
171 (2026) 112211.

\bibitem{Koch2020drivability}
A. Koch, L. Schulz, G. Jakstas, J. Falkenstein,
\textit{Drivability optimization by reducing oscillation of electric vehicle drivetrains},
World Electric Vehicle Journal,
11 (4) (2020) 68.

\bibitem{Scamarcio2020AntiJerkControllers}
A. Scamarcio, P. Gruber, S. {De Pinto}, A. Sorniotti,
\textit{Anti-jerk controllers for automotive applications: A review},
Annual Reviews in Control,
50 (2020) 174-189.

\bibitem{Scamarcio2020AntiJerkControl_Electric}
A. Scamarcio, M. Metzler, P. Gruber, S. De Pinto, A. Sorniotti,
\textit{Comparison of anti-jerk controllers for electric vehicles with on-board motors},
IEEE Transactions on Vehicular Technology,
69 (10) (2020) 10681--10699.

\bibitem{Schmidt2023RevVirtualValidation}
H. Schmidt, K. B{\"u}ttner, G. Prokop,
\textit{Methods for virtual validation of automotive powertrain systems in terms of vehicle drivability—a systematic literature review},
IEEE Access,
11 (2023) 27043--27065.

\bibitem{Crowther2005PowertrainModeling}
A. Crowther, N. Zhang,
\textit{Torsional finite elements and nonlinear numerical modelling in vehicle powertrain dynamics},
Journal of Sound and Vibration,
284 (3-5) (2005) 825--849.

\bibitem{Zeng2018DataDrivenJerk}
X. Zeng, H. Cui, D. Song, N. Yang, T. Liu, H. Chen, Y. Wang, Y. Lei,
\textit{Jerk analysis of a power-split hybrid electric vehicle based on a data-driven vehicle dynamics model},
Energies,
11 (6) (2018) 1537.

\bibitem{Pan2021DataDriven}
Y. Pan, X. Nie, Z. Li, S. Gu,
\textit{Data-driven vehicle modeling of longitudinal dynamics based on a multibody model and deep neural networks},
Measurement,
180 (2021) 109541.

\bibitem{James2020VehicleDynamicsModels}
S. S. James, S. R. Anderson, M. Da Lio,
\textit{Longitudinal vehicle dynamics: A comparison of physical and data-driven models under large-scale real-world driving conditions},
IEEE Access,
8 (2020) 73714--73729.

\bibitem{Lekshmi2019math_mod_ev}
L. S., P.S. Lal Priya,
\textit{Mathematical modeling of Electric vehicles - A survey},
Control Engineering Practice,
92 (2019) 104138.

\bibitem{Franck2021Analysis_ED}
M. Franck, J. P. Rickwärtz, D. Butterweck, M. Nell, K. Hameyer,
\textit{Transient System Model for the Analysis of Structural Dynamic Interactions of Electric Drivetrains},
Energies,
14 (4) (2021) 1108.

\bibitem{Shi2022}
P. Shi, K. Liu, K. Shi,
\textit{Integrated design of the torsion beam electric driving axle},
Journal of Vibroengineering,
 (2022) 537--549.

\bibitem{Shetty2014}
P. Shetty, S. Dawnee,
\textit{Modeling and simulation of the complete electric power train of a hybrid electric vehicle},
in: 2014 Annual International Conference on Emerging Research Areas: Magnetics, Machines and Drives,
IEEE, 2014, pp. 1--5.

\bibitem{McDonald2012}
D. McDonald,
\textit{Electric Vehicle Drive Simulation with {Matlab}/{Simulink}},
in: Proceedings of the 2012 North-Central Section Conference,
2012, pp. 1-24.

\bibitem{Butler1999}
K. L. Butler, M. Ehsani, P. Kamath,
\textit{A {Matlab}-Based Modeling and Simulation Package for Electric and Hybrid Electric Vehicle Design},
IEEE Transactions on Vehicular Technology,
48 (6) (1999) 1770-1778.

\bibitem{Zhao2023}
R. Zhao, W. Deng, Y. Wang, K. Huang, B. Zheng, J. Ding,
\textit{An Improved Data-Driven Method for Steering Feedback Torque of Driving Simulator},
IEEE/ASME Transactions on Mechatronics,
28 (5) (2023) 2953-2963.

\bibitem{Li2022SOHEstimation}
G. Li, M. Huang, R. Wang, W. Zheng,
\textit{{SOH} Estimation of Power Battery Based on the Real Vehicle Multi-feature and {LightGBM} Algorithm},
in: 2022 4th International Conference on Frontiers Technology of Information and Computer (ICFTIC),
2022, pp. 511-514.

\bibitem{Zhou2022State-of-Health}
L. Zhou, Y. Zhao, D. Li, Z. Wang,
\textit{State-of-Health Estimation for LiFePO4 Battery System on Real-World Electric Vehicles Considering Aging Stage},
IEEE Transactions on Transportation Electrification,
8 (2) (2022) 1724-1733.

\bibitem{Sun2024DataAugExtrapolation}
S. Sun, L. Feng, P. Benner,
\textit{Data-Augmented Predictive Deep Neural Network: Enhancing the extrapolation capabilities of non-intrusive surrogate models},
arXiv preprint arXiv:2410.13376,
2024.

\bibitem{Raissi2019PINNs}
M. Raissi, P. Perdikaris, G.E. Karniadakis,
\textit{Physics-informed neural networks: A deep learning framework for solving forward and inverse problems involving nonlinear partial differential equations},
Journal of Computational Physics,
378 (2019) 686-707.

\bibitem{Kontolati2024OpLearningDySys}
K. Kontolati, S. Goswami, G. Em Karniadakis, M. D. Shields,
\textit{Learning nonlinear operators in latent spaces for real-time predictions of complex dynamics in physical systems},
Nature Communications,
15 (1) (2024) 5101.

\bibitem{Hu2024StateSpaceModelsOpLearning}
Z. Hu, N. A. Daryakenari, Q. Shen, K. Kawaguchi, G. E. Karniadakis,
\textit{State-space models are accurate and efficient neural operators for dynamical systems},
arXiv preprint arXiv:2409.03231,
2024.

\bibitem{Wang2022hybrid}
J. Wang, Y. Li, R. X. Gao, F. Zhang,
\textit{Hybrid physics-based and data-driven models for smart manufacturing: Modelling, simulation, and explainability},
Journal of Manufacturing Systems,
63 (2022) 381--391.

\bibitem{Kasilingam2024Hybrid}
S. Kasilingam, R. Yang, S. K. Singh, M. A. Farahani, R. Rai, T. Wuest,
\textit{Physics-based and data-driven hybrid modeling in manufacturing: a review},
Production \& Manufacturing Research,
12 (1) (2024) 2305358.

\bibitem{Shah2024hybrid}
P. Shah, S. Pahari, R. Bhavsar, J. S.-I. Kwon,
\textit{Hybrid modeling of first-principles and machine learning: A step-by-step tutorial review for practical implementation},
Computers \& Chemical Engineering,
 (2024) 108926.

\bibitem{Skull2024}
M. Skull, M. Bach, T. Roulet, M. Hauss,
\textit{{AI}-Based Methodology for the Calibration of Anti-Jerk Control on Electric Drives},
ATZ - Automobiltechnische Zeitschrift,
(2024).

\bibitem{Griffin1984}
D. Griffin, J. Lim,
\textit{Signal Estimation from Modified Short-Time {Fourier} Transform},
IEEE Transactions on Acoustics, Speech, and Signal Processing,
32 (2) (1984) 236-243.

\bibitem{Oppenheim1999STFT}
A. V. Oppenheim,
\textit{Discrete-time signal processing},
Pearson Education India,
1999.

\bibitem{Paszke2019pytorch}
A. Paszke, S. Gross, F. Massa, A. Lerer, J. Bradbury, G. Chanan, T. Killeen, Z. Lin, N. Gimelshein, L. Antiga, others,
\textit{Pytorch: An imperative style, high-performance deep learning library},
Advances in neural information processing systems,
32 (2019).

\bibitem{Harshvardhan2020GenModels}
G. Harshvardhan, M. K. Gourisaria, M. Pandey, S. S. Rautaray,
\textit{A comprehensive survey and analysis of generative models in machine learning},
Computer Science Review,
38 (2020) 100285.

\bibitem{Doersch2016tutorialVAEs}
C. Doersch,
\textit{Tutorial on variational autoencoders},
arXiv preprint arXiv:1606.05908,
2016.

\bibitem{Kingma_2019}
D. P. Kingma, M. Welling,
\textit{An Introduction to Variational Autoencoders},
Foundations and Trends® in Machine Learning,
12 (4) (2019) 307–392.

\bibitem{Mushtaq2011}
R. Mushtaq,
\textit{Augmented {Dickey} {Fuller} Test},
SSRN,
2011.

\bibitem{Kokoszka2016}
P. Kokoszka, G. Young,
\textit{{KPSS} Test for Functional Time Series},
Statistics,
50 (5) (2016) 957-973.

\bibitem{Seabold2010statsmodels}
S. Seabold, J. Perktold,
\textit{Statsmodels: econometric and statistical modeling with python.},
SciPy,
7 (1) (2010).

\bibitem{Cai2022theoreticaltSNE}
T. T. Cai, R. Ma,
\textit{Theoretical foundations of {t-SNE} for visualizing high-dimensional clustered data},
Journal of Machine Learning Research,
23 (301) (2022) 1--54.

\bibitem{Van2008visualizingtSNE}
L. Van der Maaten, G. Hinton,
\textit{Visualizing data using {t-SNE}},
Journal of machine learning research,
9 (11) (2008).

\bibitem{roweis2004neighbourhood}
S. Roweis, G. Hinton, R. Salakhutdinov,
\textit{Neighbourhood component analysis},
Adv. Neural Inf. Process. Syst.(NIPS),
17 (4) (2004).

\bibitem{Pedregosa2011ScikitLearn}
F. Pedregosa, G. Varoquaux, A. Gramfort, V. Michel, B. Thirion, O. Grisel, M. Blondel, P. Prettenhofer, R. Weiss, V. Dubourg, J. Vanderplas, A. Passos, D. Cournapeau, M. Brucher, M. Perrot, E. Duchesnay,
\textit{Scikit-learn: Machine Learning in {P}ython},
Journal of Machine Learning Research,
12 (2011) 2825--2830.

\bibitem{WINKLER2021101072}
S. Winkler, A. Körner, F. Breitenecker,
\textit{Modelling framework for artificial hybrid dynamical systems},
Nonlinear Analysis: Hybrid Systems,
42 (2021) 101072.

\bibitem{HARLIM2021109922}
J. Harlim, S. W. Jiang, S. Liang, H. Yang,
\textit{Machine learning for prediction with missing dynamics},
Journal of Computational Physics,
428 (2021) 109922.

\bibitem{Kotter2018X_in_theLoop}
M. K{\"o}tter, B. Lindemann, D. Bergmann, M. Ehrly, T. Jung, M. Nijs, S. Thewes, T. K{\"o}rfer, S. Trampert, T. Drecq, P. Gautier,
\textit{Powertrain calibration based on X-in-theLoop: Virtualization in the vehicle development process},
in: 18. Internationales Stuttgarter Symposium,
Springer Fachmedien Wiesbaden, Wiesbaden,
2018,
pp. 1187--1201.

\bibitem{ronneberger2015unetconv}
O. Ronneberger, P. Fischer, T. Brox,
\textit{U-Net: Convolutional Networks for Biomedical Image Segmentation},
Medical Image Computing and Computer-Assisted Intervention (MICCAI),
2015.

\bibitem{stoller2018waveunetnet}
D. Stoller, S. Ewert, S. Dixon,
\textit{{Wave-U-Net}: A Multi-Scale Neural Network for End-to-End Audio Source Separation},
in: Proceedings of the 19th ISMIR Conference,
2018.

\bibitem{liu2022convnet2020s}
Z. Liu, H. Mao, C.-Y. Wu, C. Feichtenhofer, T. Darrell, S. Xie,
\textit{A ConvNet for the 2020s},
in: 2022 IEEE/CVF Conference on Computer Vision and Pattern Recognition (CVPR),
2022.

\bibitem{Sandler2018}
M. Sandler, A. Howard, M. Zhu, A. Zhmoginov, L.-C. Chen,
\textit{Mobilenetv2: Inverted residuals and linear bottlenecks},
in: Proceedings of the IEEE conference on computer vision and pattern recognition,
2018, pp. 4510--4520.

\bibitem{Dilokthanakul2016}
N. Dilokthanakul, P. A. Mediano, M. Garnelo, M. C. Lee, H. Salimbeni, K. Arulkumaran, M. Shanahan,
\textit{Deep Unsupervised Clustering with {Gaussian} Mixture Variational Autoencoders},
arXiv preprint,
2016.

\bibitem{Dillon2021}
B. M. Dillon, T. Plehn, C. Sauer, P. Sorrenson,
\textit{Better Latent Spaces for Better Autoencoders},
SciPost Physics,
11 (3) (2021) 061.

\bibitem{Kramer1992}
M. A. Kramer,
\textit{Autoassociative Neural Networks},
Computers and Chemical Engineering,
16 (1992) 313--328.

\bibitem{Pinaya2020}
W. H. L. Pinaya, S. Vieira, R. Garcia-Dias, A. Mechelli,
\textit{Autoencoders},
in: Machine Learning,
Academic Press, 2020, pp. 193-208.

\bibitem{Aqeel2021}
A. Anwar,
\textit{Difference between AutoEncoder (AE) and Variational AutoEncoder (VAE)},
Towards Data Science,
2021.

\bibitem{Vo-Duy2016}
T. Vo-Duy, C. M. Ta,
\textit{A Signal Hardware-in-the-Loop Model for Electric Vehicles},
Robomech Journal,
3 (2016) 29.

\bibitem{Vedal2019}
A. Hansen Vedal,
\textit{Unsupervised Audio Spectrogram Compression using Vector Quantized Autoencoders},
2019.

\bibitem{Wei2020}
R. Wei, C. Garcia, A. El-Sayed, V. Peterson, A. Mahmood,
\textit{Variations in Variational Autoencoders-A Comparative Evaluation},
IEEE Access,
8 (2020) 153651-153670.

\bibitem{inbook_hybrid_model}
A. Daw, A. Karpatne, W. Watkins, J. Read, V. Kumar,
\textit{Physics-Guided Neural Networks (PGNN): An Application in Lake Temperature Modeling},
2022, pp. 353-372.

\end{thebibliography}

\end{document}